\documentclass[10pt,letterpaper]{article}
\usepackage[top=0.85in,left=2.75in,footskip=0.75in]{geometry}

\usepackage{amsmath,amssymb}

\usepackage{changepage}

\usepackage[utf8x]{inputenc}
\usepackage{textcomp,marvosym}

\usepackage{cite}

\usepackage{nameref,hyperref}

\usepackage[right]{lineno}

\usepackage{microtype}
\DisableLigatures[f]{encoding = *, family = * }

\usepackage[table]{xcolor}

\usepackage{array}

\usepackage{booktabs}       %

\usepackage{pgfplots, amsthm, graphicx, subcaption, dsfont, tikz, tikz-qtree, tabularx, enumerate, verbatim, enumitem,makecell,multirow, bm, arydshln, soul, float}%
\usepackage[colorinlistoftodos, textwidth=30mm,color=blue!10]{todonotes}%
\usepackage[noabbrev, capitalise]{cleveref} %

\newcolumntype{+}{! {\vrule width 2pt}}

\newlength\savedwidth

\raggedright
\usepackage[aboveskip=1pt,labelfont=bf,labelsep=period,justification=raggedright,singlelinecheck=off]{caption}

\makeatletter
\renewcommand{\@biblabel}[1]{\quad#1.}
\makeatother

\usepackage{lastpage,fancyhdr,graphicx}
\usepackage{epstopdf}
\fancyheadoffset[L]{2.25in}
\fancyfootoffset[L]{2.25in}
\DeclareMathOperator\arctanh{arctanh}

\begin{document}
\vspace*{0.2in}

\begin{flushleft}
{\Large
\textbf\newline{Inspecting class hierarchies in classification-based metric learning models} %
}%
\newline
\\
Hyeongji Kim%
\textsuperscript{1,2*},
Pekka Parviainen\textsuperscript{2},
Terje Berge\textsuperscript{1},
Ketil Malde\textsuperscript{1,2}
\\
\bigskip
\textbf{1} %
Institute of Marine Research, Bergen, Norway
\\
\textbf{2} %
Department of Informatics, University of Bergen, Norway
\\
\bigskip

* hjk92g@gmail.com

\end{flushleft}
\section*{Abstract}
Most classification models treat all misclassifications equally. However, different classes may be related, and these hierarchical relationships must be considered in some classification problems. %
These problems can be addressed by using hierarchical information during training. %
Unfortunately, this information is not available for all datasets. %
Many classification-based metric learning methods use class representatives in embedding space to represent different classes. %
The relationships among the learned class representatives can then be used to
estimate class hierarchical structures. %
If we have a predefined class hierarchy, the learned class representatives can be assessed to determine whether the metric learning model learned semantic distances that match %
our prior knowledge. %
In this work, we train a softmax classifier and three metric learning models with several training %
options on benchmark and real-world datasets. In addition to the standard classification accuracy, we evaluate the hierarchical inference performance by inspecting learned class representatives and the hierarchy-informed performance, i.e., the classification performance, and the metric learning performance by considering predefined hierarchical structures. %
Furthermore, we investigate how the considered measures are affected by various models and training options. %
When our proposed ProxyDR model is trained without using predefined hierarchical structures, the hierarchical inference performance is significantly better than that of the popular NormFace model. Additionally, our model enhances some hierarchy-informed performance measures under the same training options. %
We also found that convolutional neural networks (CNNs) with random weights correspond to the predefined hierarchies better than random chance.

\section*{Introduction}

Neural network-based classifiers have shown impressive classification %
accuracy. For instance, a convolutional neural network (CNN) classifier \cite{he2015delving} surpassed human-level top-5 classification accuracy (94.9\%) \cite{russakovsky2015imagenet} on the 1000-class classification challenge on the ImageNet dataset \cite{deng2009imagenet}. Most training loss functions in neural network classifiers treat all misclassifications %
equally. %
However, in practice, the severity of various misclassifications may differ considerably. For instance, in an autonomous vehicle system, mistaking a person as a tree can result in a more %
catastrophic consequence than mistaking a streetlight as a tree \cite{bertinetto2020making}.
In addition, some classification tasks include a large number of classes, such as the 1000-class ImageNet classification challenge, and hierarchical relationships may exist among these classes. %
When %
several classification models achieved similar accuracy, 
one would prefer to choose models in which the wrongly predicted classes are “hierarchically” close to the ground-truth classes. 
To address the severity %
of misclassifications and relationships among classes, hierarchical information can be used.
For instance, predefined hierarchical structures can be incorporated into training by replacing standard labels with soft labels based on this hierarchical information \cite{bertinetto2020making}. %
Some metric learning approaches also use hierarchical information \cite{barz2019hierarchy, garnot2021leveraging, jayathilaka2021ontology}.\\
In general, metric learning methods learn embedding functions in which similar data points are close and dissimilar data points are far apart according to the distance metric in the learned embedding space. For class-labeled datasets, data points in the same class %
are regarded as similar, %
while %
data points in different classes %
are regarded as dissimilar. %
Metric learning can be applied in image retrieval tasks \cite{musgrave2020metric} to identify relevant images or few-shot classification tasks \cite{snell2017prototypical, chen2019closer}, which are classification tasks with only a few examples per class. %
Usually, metric learning methods assume that there are no special relations among classes, i.e., a flat hierarchy is assumed. However, a predefined class hierarchy can be incorporated into the training process of metric learning models to improve the hierarchy-informed performance %
\cite{barz2019hierarchy, garnot2021leveraging, jayathilaka2021ontology}. \\%Particularly, a known class hierarchy can be incorporated in the training process for hierarchical classification. \\%hierarchy-informed metric learning methods... %
Class hierarchical structures can be defined in several ways. Hierarchical structures can be defined by domain experts \cite{garnot2021leveraging} or extracted from WordNet \cite{miller1998wordnet}, which is a database that contains semantic relations among English words.
For instance, ImageNet classes \cite{deng2009imagenet} are %
organized according to WordNet. However, hierarchy determination from WordNet requires that a class name or its higher class is in the WordNet database. %
When these %
approaches are not applicable, hierarchy can be inferred by estimating the class distance matrix based on learned classifiers. For instance, the confusion matrix of a classifier can be used to estimate relations among class pairs \cite{godbole2002exploiting}. Each row in the confusion matrix can be treated as a vector, and the distance between these vectors can be calculated to estimate the class hierarchical structure. %
However, this approach can be cumbersome for %
hierarchy-informed classification tasks, as they require separate training and evaluation (validation) processes to determine the hierarchical structure. Moreover, this approach %
becomes challenging when some classes contain a very small number of data points, as some elements in the confusion matrix may be uninformative. %
One type of metric learning model uses a unique position in embedding space (class representative) to represent each class \cite{wang2017normface, kim2020proxy, wang2018cosface, deng2019arcface}. %
Class representatives can also be used to infer hierarchical structures by considering their distances \cite{garnot2021leveraging, wan2021nbdt}. This approach does not require a separate training process, as class representatives can be learned automatically. \\
On the other hand, when we have a predefined hierarchy, the learned class representatives can improve our understanding of the trained metric learning model. %
For example, we can determine if the semantic distance learned by a metric learning model matches our prior knowledge (that is, the predefined hierarchy). For instance, when a model has been trained to classify species, we can determine if the model regards a dog as closer to a cat than to a rose.
Furthermore, these inspections can be used to evaluate the trustworthiness of a model. However, previous works have paid little attention to the relationships among class representatives. 
In this work, we assess several metric learning methods and training options by focusing on their learned class representatives and hierarchy-informed performance. %
Moreover, we attempt to determine conditions that improve the hierarchy inference performance and evaluate  whether such models enhance the hierarchy-informed performance. %
We also investigate different training options with predefined hierarchies for comparison. \\

\subsection*{Problem settings}\label{sec:setting}

\begin{figure}[H]
    \centering
    \includegraphics[width=1.0\linewidth,height=0.375\linewidth]{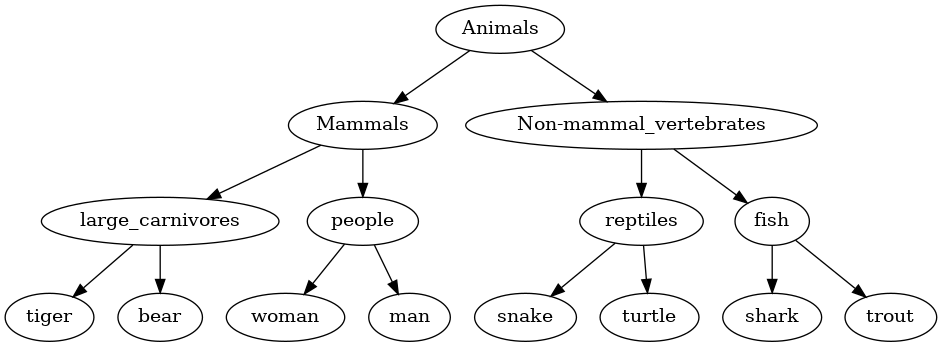}
\caption{{\bf Pruned CIFAR100 tree structure for visualization.} The whole hierarchical structure is shown in %
a table in \nameref{S2_appendix}. %
}\label{fig:Fig1}%
\end{figure}

The predefined hierarchical distance, i.e., the distance between two classes, often needs to be considered when training models with hierarchical information and evaluating model performance. Barz and Denzler \cite{barz2019hierarchy} and Bertinetto et al. \cite{bertinetto2020making} used bounded (\(\left [ 0,1 \right ]\)) dissimilarity based on the height of the lowest common ancestor (LCA) between two classes.
In this work, %
similar to Garnot and Landrieu \cite{garnot2021leveraging}, we define the hierarchical distance as the shortest path distance between two classes. %
For instance, in the predefined hierarchical tree shown in Fig. \ref{fig:Fig1}, %
the hierarchical distance between the classes ``tiger'' and ``woman'' is \(4\) (\(=2+2\)), as we need to move two steps upward and two steps downward to move from one class to the other. Similarly, the hierarchical distance between the classes ``tiger'' and ``shark'' is \(6\). %
Hierarchical structures can be expressed as trees or directed acyclic graph (DAG) structures \cite{silla2011survey}. In this work, we consider tree-structured hierarchies. In other words, each node cannot have more than one parent node. %

Let \(\mathcal{X} \subseteq \mathbb{R}^{d_{I}}\) be %
an input space and \(\mathcal{Y}\) be a set of classes. The data %
points \(x\in\mathcal{X}\) and corresponding classes \(c\in\mathcal{Y}\) %
are sampled from the joint distribution \(\mathcal{D}\). %
The feature mapping \(f: \mathcal{X}\rightarrow \mathcal{Z}\) %
extracts feature vectors according to the inputs, where \(\mathcal{Z}=\mathbb{R}^{d_{F}}\) is a raw %
feature %
space. In this work, we assume that this mapping is %
modeled by a neural network. %

\subsection*{Softmax classification}
The softmax classifier is commonly used in neural network-based classification tasks. %
The softmax classifier estimates the class probability \(p(c|x)\), %
which is the probability that data point \(x\) belongs to class \(c\), as follows:
\begin{align}\label{eq:softmax}
    p(c|x)=\frac{\exp(l_c (x))}{\sum\limits_{y\in \mathcal{Y}}{\exp(l_y (x))}},
\end{align}
where logit \(l_y (x)\) is the neural network output according to input \(x\in\mathcal{X}\) and class \(y\in \mathcal{Y}\). %
Usually, logit \(l_y (x)\) is calculated as:
\begin{align}\label{eq:logit}
    l_y (x)=W_y^T f(x)+b_y,
\end{align}
where the feature vector \(f(x)\) is an output of a penultimate layer, %
\(W_y\) is a weight vector, and \(b_y\) is a bias term for class \(y\).
The probability \(p(c|x)\) estimated by the model is often called the confidence value (score). Based on the estimated class probabilities, we can use the cross-entropy loss to train the model. This loss measures the difference between the predicted and target probability distributions. The cross-entropy (CE) loss of a mini-batch %
\(B\subseteq \mathcal{D}\) is defined as:
\begin{align}\label{eq:CE_loss}
    L_{CE}=-\frac{1}{|B|}\sum\limits_{(x_i,c_i)\in B}{\log (p(c_i|x_i))}. %
\end{align}

\subsection*{Metric learning}

While softmax classifiers are commonly used in deep learning classification, metric learning approaches can be beneficial, as they better control data points in embedding space. %
Hence, metric learning can provide %
information on the similarity between different data points. %
Metric learning approaches can be divided into two categories: %
(direct) embedding-based methods and classification-based methods. %
Embedding-based methods \cite{weinberger2005distance, hadsell2006dimensionality} directly compare data points in embedding space to train an embedding function (feature map) \(f(\cdot)\). Because embedding-based methods compare data points directly, they must be trained with pairs (using the contrastive loss) or triplets (using the triplet loss) of data points. Thus, embedding-based methods have 
high training complexity and require special mining algorithms \cite{wang2017normface} to prevent slow convergence speeds \cite{kim2020proxy}. %
On the other hand, classification-based methods \cite{wang2017normface, wang2018cosface, deng2019arcface, zhe2019improve, zhang2019adacos} use %
class representatives to represent classes. According to the classification loss (cross-entropy loss),
class representatives guide data points to converge to %
class-specific positions. %
Classification-based methods converge faster than embedding-based methods due to their reduced sampling complexity (training with single data points). 
In this paper, %
we focus on classification-based metric learning methods. %
\subsubsection*{NormFace}
\begin{figure}[H]
\centering
   \includegraphics[width=0.3\linewidth,height=0.25\linewidth]{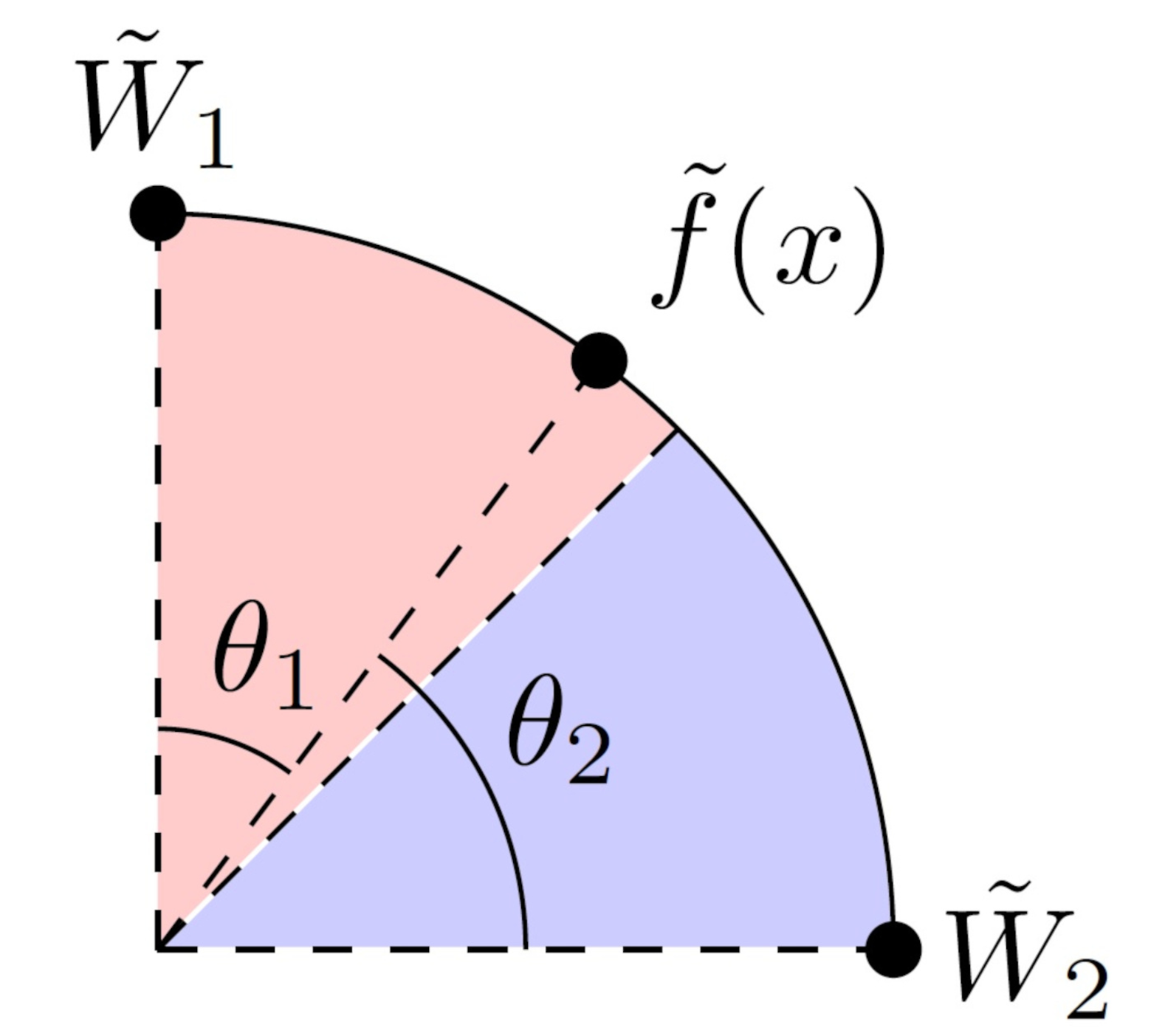}
\caption{{\bf Visualization of NormFace classification results for two classes with \(\Tilde{W}_1\)=(0,1) and \(\Tilde{W}_2\)=(1,0).} As \(\Tilde{f}(x)\) is closer to \(\Tilde{W}_1\) than \(\Tilde{W}_2\), data point \(x\) is classified as belonging to class 1.}\label{fig:Fig2}%
\end{figure}

NormFace \cite{wang2017normface}, which is also known as normalized softmax \cite{zhai2019classification}, modifies Eq. \ref{eq:logit} in the softmax classifier. %
NormFace was motivated by the normalization of features during feature comparisons to improve face verification during the testing phase. 
To apply normalization during both the testing and training phases, %
NormFace normalizes the feature (embedding) vectors and weight vectors and uses a zero bias term.  %
Previous experiments on metric learning models \cite{musgrave2020metric} have shown that NormFace and its variants \cite{liu2017sphereface, wang2018cosface, deng2019arcface} achieved competitive performance on metric learning tasks. \\%
We denote \(\Tilde{v}=\frac{v}{\left\| v\right\|}\) for any nonzero vector \(v\) %
and the angle between vectors \(\Tilde{W}_y\) and \(\Tilde{f}(x)\) as \(\theta_y\). %
Then, as \(\left\|\Tilde{W}_y \right\|=1= \left\|\Tilde{f}(x) \right\|\), we obtain the following equation: 
\begin{align}\label{eq:cos_theta}
    \Tilde{W}_y^T \Tilde{f}(x)=\left\|\Tilde{W}_y \right\| \left\|\Tilde{f}(x) \right\|\cos{\theta_y}=\cos{\theta_y}.
\end{align}
According to Eq. \ref{eq:cos_theta}, the class probability \(p(c|x)\) estimated by NormFace can be expressed as:
\begin{align}\label{eq:normface}
    p(c|x)=\frac{\exp(s \Tilde{W}_c^T \Tilde{f}(x))}{\sum\limits_{y\in \mathcal{Y}}{\exp(s \Tilde{W}_y^T \Tilde{f}(x))}}=\frac{\exp(s\cos{\theta_c})}{\sum\limits_{y\in \mathcal{Y}}{\exp(s\cos{\theta_y})}},
\end{align}
where \(s>0\) is a scaling factor. NormFace learns embeddings according to this estimation and the cross-entropy loss in Eq. \ref{eq:CE_loss}.\\%, \(\Tilde{v}=\frac{v}{\left\| v\right\|}\) for any nonzero vector \(v\), and \(\theta_y\) is the angle between vectors \(\Tilde{W}_y\) and \(\Tilde{f}(x)\). \\%As \(\Tilde{W}_y^T \Tilde{f}(x)=\cos{\theta_y}\) where \(\theta_y\) is the angle between \(\Tilde{W}_y\) and \(\Tilde{f}(x)\),
We next investigate the geometrical meaning of the NormFace classification results. If a data point \(x\) is classified as belonging to class \(c\) by NormFace, according to Eq. \ref{eq:normface}, %
we obtain \(\exp(s\cos{\theta_{c}}) \ge \exp(s \cos{\theta_y})\), i.e., \(\cos{\theta_{c}} \ge \cos{\theta_y}\). As the cosine function is a monotonically decreasing function in the interval %
\(\left ( 0, \pi \right)\), we obtain \(\theta_{c}\le \theta_{y}\). %
In terms of angles, the normalized embedding vector \(\Tilde{f}(x)\) is closer (or equally close) to \(\Tilde{W}_{c}\) %
than \(\Tilde{W}_y\).
As we can classify data points using normalized weight vectors according to this geometrical interpretation, we can consider \(\Tilde{W}_y\) as the class representative for a class \(y\in\mathcal{Y}\). %
Fig. \ref{fig:Fig2} %
visualizes the above explanation.

\subsubsection*{Proxies and prototypes}
Although NormFace \cite{wang2017normface} uses a learnable weight vector as a class representative for each class, the average point for a class can also be used as a class representative. To prevent confusion, %
we define two kinds of class representatives for each class: a proxy representative and a prototype representative.
We refer to the learnable weight vectors as proxy representatives.
In NormFace, the weight vectors \(\Tilde{W}_{y}\) are proxies. In this work, %
the proxy representative are used to define the training loss and directly update the network. Note that we can predefine the proxy representatives, and the proxies can be fixed. %
On the other hand, we refer to the (normalized) average embedding for each class as a prototype representative. %
We denote the average embedding for class \(y\) as vector \(\mu_y\). In normalized space, this vector is defined as:
\begin{align*}
    \mu_y=\frac{1}{|X_{y}|} \sum_{x\in X_{y}} \tilde{f}(x),
\end{align*}
where \(X_{y}\subseteq\mathcal{X}\) is a set of data points in class \(y\).
Then, the prototype representatives are defined as \(\tilde{\mu}_y=\frac{\mu_y}{\left\| \mu_y\right\|}\). %
A metric learning method known as the prototypical network %
\cite{snell2017prototypical} was devised for few-shot learning tasks. During the training process, this network uses local prototypes based on special mini-batches known as episodes. In their work, the authors used unnormalized average embedding as prototypes because they considered Euclidean space. %
In this paper, we do not use prototypes for training, %
and %
(global) prototypes using training data are used only for evaluation. \\
Note that the proxy and prototype representatives %
are not necessarily the same. For instance, when we train the model with fixed proxies, the positions of the prototypes change during training, while the positions of the proxies remain fixed. Moreover, the confidence values are not necessarily maximized at the proxies \cite{kim2022distance}. In this case, the data points may not converge to their proxies. This concept is explained further in the next subsubsection. \\%}\\%in most cases.

\subsubsection*{SD softmax and DR formulation}
In our previous work \cite{kim2022distance}, we analyzed %
a softmax formulation for metric learning in which negative squared distances were used as logit values. %
We called this formulation the ``softmax-based formulation'' in our previous paper. In this work, we refer to this formulation as ``squared distance (SD) softmax'' to prevent confusion with the standard softmax formulation shown in Eq. \ref{eq:softmax}. %

We denote the class representative for class \(y\) as \(R_y\) and the distance to class \(y\) %
as \(d_{x,y}:=d(f(x),R_y)\) according to the distance function \(d(\cdot,\cdot)\). For the normalized embeddings, we define \(d_{x,y}:=d(\tilde{f}(x), R_y)\) by assuming that point \(R_y\) is normalized. 
Then, the class probability \(p(c|x)\) estimated using the SD softmax formulation is:%
\begin{align}\label{eq:SD_softmax}
    p(c|x)=\frac{\exp(- d_{x,c}^2)}{\sum\limits_{y\in \mathcal{Y}}{\exp(-d_{x,y}^2)}}.
\end{align}
The SD softmax formulation uses the difference in the squared distance for training. \\
Consider the following equation:
\begin{align*}%
\left\| \Tilde{W}_y-\Tilde{f}(x)\right\|^2=\left\| \Tilde{W}_y\right\|^2+\left\| \Tilde{f}(x)\right\|^2-2\Tilde{W}_y^T \Tilde{f}(x)=1+1-2\cos \theta_y.
\end{align*}
Then, we obtain:
\begin{align*}
s\cos \theta_y -s=-\frac{s}{2}\left\| \Tilde{W}_y-\Tilde{f}(x)\right\|^2.%
\end{align*}
Thus, we obtain the equation:
\begin{align}\label{eq:normface_SD}
\frac{\exp(s\cos{\theta_c})}{\sum\limits_{y\in \mathcal{Y}}{\exp(s\cos{\theta_y})}}=\frac{\exp(s\cos{\theta_c}-s)}{\sum\limits_{y\in \mathcal{Y}}{\exp(s\cos{\theta_y}-s)}} =\frac{\exp(-\frac{s}{2}\left\| \Tilde{W}_c-\Tilde{f}(x)\right\|^2)}{\sum\limits_{y\in \mathcal{Y}}{\exp(-\frac{s}{2}\left\| \Tilde{W}_y-\Tilde{f}(x)\right\|^2)}},
\end{align}
(This equation is taken from \cite{wang2017normface}).
The above equation shows that the NormFace formulation in Eq. \ref{eq:normface} can be considered %
an SD softmax formulation \ref{eq:SD_softmax} that uses the Euclidean distance on a hypersphere with radius \(\sqrt{\frac{s}{2}}\).\\
In \cite{kim2022distance}, we found that the SD softmax formulation had two main limitations. First, the estimated probability and corresponding loss may be affected by scaling changes. However, while this limitation must be considered when a Euclidean space is used, this limitation can be addressed by using normalized embeddings, as in the NormFace model \cite{wang2017normface}. The second limitation is that the estimated class probabilities are not optimized at the class representatives. %
For instance, the maximum estimated class probability \(p(c|x)\) is not found at the class representative of class \(c\). The NormFace model also encounters this issue. %
In the example shown in Fig. \ref{fig:Fig2} %
with scaling factor \(s=2\), point \((-\frac{1}{\sqrt{2}},\frac{1}{\sqrt{2}})\) is the point that maximizes the confidence value of class \(1\).\\
To address the above limitations, we proposed the distance ratio (DR)-based formulation \cite{kim2022distance} for metric learning models. %
Mathematically, the DR formulation estimates the class probability \(p(c|x)\) as:
\begin{align}\label{eq:DR_form}
    p(c|x)=\frac{\frac{1}{d_{x,c}^s}}{\sum\limits_{y\in \mathcal{Y}}{\frac{1}{d_{x,y}^s}}}=\frac{d_{x,c}^{-s}}{\sum\limits_{y\in \mathcal{Y}}{d_{x,y}^{-s}}}.
\end{align}
The DR formulation uses ratios of distances for training. \\
Moreover, the DR formulation \cite{kim2022distance} resolves the above two limitations of the SD softmax %
formulation. %
In the example shown in Fig. \ref{fig:Fig2}, %
for any scaling factor \(s\), point \((0,1)=\Tilde{W}_1\) is the point that maximizes the confidence value of class \(1\). %
Thus, our %
experiments showed that the DR formulation has faster or comparable training speed in Euclidean (unnormalized) embedding spaces.\\%\\%}
\subsubsection*{Exponential moving average (EMA) approach} %
Zhe et al. \cite{zhe2019improve} theoretically showed that, while training NormFace \cite{wang2017normface}, the commonly used gradient descent update method for the proxies based on the training loss cannot guarantee that the updated proxies approach the corresponding prototypes. %
To address this issue, they proposed using the normalized exponential
moving average (EMA) to update the proxies. Mathematically, when %
updating a proxy for a data point \(x\) in class \(c\), the proxy \(\Tilde{W}_c\) is updated as: %
\begin{align}\label{eq:EMA}
    \Tilde{W}_c :=\frac{\alpha \Tilde{f}(x)+(1-\alpha) \Tilde{W}_c}{\left\|\alpha \Tilde{f}(x)+(1-\alpha) \Tilde{W}_c \right\|},
\end{align}
where \(0<\alpha<1\) is a parameter that controls the speed and stability of the updates. Their experimental results showed that the EMA approach achieved better performance than the standard NormFace model on multiple datasets.\\%}
\subsubsection*{Adaptive scaling factor approach} %
Based on previous observations that the convergence and performance of NormFace models \cite{wang2017normface} depend on the scale parameter \(s\),
Zhang et al. \cite{zhang2019adacos} proposed AdaCos, which is a NormFace model trained by using the adaptive scale factor \(s\) in Eq. (\ref{eq:normface}). %
Moreover, they suggested to use parameter \(s\), which significantly changes the probability \(p(c|x)\) estimated by Eq. (\ref{eq:normface}), where \(c\) is the class of data point \(x\). 
In other words, they attempted to find a parameter \(s\) that maximizes %
\(\left\| \frac{\partial p(c|x) (\theta_c)}{\partial \theta_c}\right\|\) by approximating an \(s\) value that satisfies the equation: %
\begin{align}\label{eq:adacos}
\left\| \frac{\partial^2 p(c|x) (\theta_{c}')}{\partial \theta_{c}'^2}\right\|=0,
\end{align}
where \(\theta_{c}':=\text{clip}(\theta_{c},0,\frac{\pi}{2}) %
\) and \(\text{clip} (\cdot, \cdot, \cdot)\) is a function that limits a value within a specified range. They proposed two AdaCos models: static (fixed) and dynamic versions. The static model determines a good scale parameter \(s\) before training the NormFace model based on observations of the angles between the data points and proxies. In the static model, the scale parameter \(s\) is not updated. %
The dynamic model updates the scale parameter \(s\) during each iteration based on the current angles between the data points and proxies.\\%}

\subsubsection*{CORR loss} %
In contrast to previous metric learning approaches that ignored hierarchical relationships among classes, Barz and Denzler \cite{barz2019hierarchy} used normalized embeddings to achieve hierarchy-informed classification. %
First, they predefined the positions of the proxies using a given hierarchical structure. 
These proxies are fixed, i.e., they are not updated during training.
Then, they used the predefined proxies to train the models according to the CORR loss. For a data point \(x\) in class \(c\), %
the CORR loss ensures that the embedding vector \(\Tilde{f}(x)\) is close to the corresponding proxy \(\Tilde{W}_{c}\). %
The CORR loss of a mini-batch %
\(B\subseteq \mathcal{D}\) is defined as:
\begin{align}\label{eq:CORR_loss}
    L_{CORR}=\frac{1}{|B|}\sum\limits_{(x_i,c_i)\in B}{\left (1-\Tilde{W}_{c_i}^T \Tilde{f}(x_i)\right )}=\frac{1}{|B|}\sum\limits_{(x_i,c_i) \in B}{\left (1-\cos{\theta_{c_i}}\right )},
\end{align}
where \(\theta_{c_i}\) is the angle between \(\Tilde{W}_{c_i}\) and \(\Tilde{f}(x_i)\).

\section*{Methods}

We investigated several methods and training options. The details of the settings are described in the following sections. The codes for our experiments %
will be available in \url{https://github.com/hjk92g/Inspecting_Hierarchies_ML}.

\paragraph{Dataset}
We conducted experiments using three plankton datasets (small microplankton, large microplankton, and mesozooplankton) %
and two benchmark datasets (CIFAR100 \cite{Krizhevsky09learningmultiple} and NABirds %
\cite{van2015building}). Table \ref{tab:datasets} summarizes %
the number of classes and images in each dataset. %
The three plankton datasets were obtained from the Institute of Marine Research (IMR), where flow imaging microscopy is used during routine monitoring. %
The plankton samples were imaged using three FlowCams, \copyright 2022 Yokogawa Fluid Imaging Technologies, Inc., with different magnification settings. %
The three plankton datasets contain nonliving classes of artifacts and debris. %
Moreover, these datasets contain class names that are not in the WordNet database \cite{miller1998wordnet}. %
In addition, the three plankton datasets have severe class imbalances. For example, each class in the small microplankton (MicroS) dataset contains 1 to 456 images. Further details on the plankton datasets %
can be found in \nameref{S2_appendix}. %
We randomly divided each plankton dataset into 70\(\%\) training data, 10\(\%\) validation data, and 20\(\%\) test data. For the plankton datasets, we use images without any augmentation. %
As the input images have diverse image sizes, we resized the input images to \(128 \times 128\). The CIFAR100 dataset contains 100 classes and 600 images per class. The CIFAR100 contains 50000 training and 10000 test images, and we randomly divided the original training data into 45000 training and 5000 validation data. Furthermore, we applied random transformations (15 degree range rotations, 10\% range translations, 10\% range scaling, 10 degree range shearing, and horizontal flips) on the CIFAR100 training data. We did not resize images from their original size of \(32 \times 32\). %
The NABirds dataset contains 23929 training and 24633 test images. Similar to the plankton datasets, we randomly divided  this dataset into 70\(\%\) training data, 10\(\%\) validation data, and 20\(\%\) test data. We resized the images to \(128 \times 128\) pixels. %
We also applied the same augmentations applied to the CIFAR100 dataset on the NABirds dataset. %
The predefined hierarchical structures are available in \nameref{S1_appendix}. %

\newcolumntype{M}[1]{>{\arraybackslash}m{#1}}
\newcolumntype{\Mc}[1]{>{\centering\arraybackslash}m{#1}}

\begin{table}[H]
\begin{adjustwidth}{-2.25in}{0in} %
\centering
\caption{
{\bf Summary of the studied datasets.} }
\begin{tabular}{\Mc{3.0cm}|\Mc{2.5cm}|\Mc{2.5cm}|\Mc{2.5cm}|\Mc{2.5cm}}
 \small \makecell{Dataset} & \footnotesize \makecell{Target particle\\(plankton datasets)} & \footnotesize %
 Images per class & \footnotesize Images & \footnotesize Classes \\
          \toprule
\small Small microplankton (MicroS) & \footnotesize \(5\) to \(50 \,\mu m\) &  \footnotesize 1 to 456 & \footnotesize 6738 & \footnotesize 109\\ \midrule 
\small Large microplankton (MicroL) & \footnotesize \(35\) to \(500 \,\mu m\)  & \footnotesize 2 to 613 & \footnotesize 8348 & \footnotesize 102\\ \midrule 
\small Mesozooplankton (MesoZ) & \footnotesize \(180\) to \(2000 \,\mu m\) & \footnotesize 3 to 486 & \footnotesize 6738 & \footnotesize 52\\ \midrule 
\small CIFAR100 \cite{Krizhevsky09learningmultiple}& \footnotesize --- & \footnotesize 600 & \footnotesize 60000 & \footnotesize 100\\ \midrule
\small NABirds \cite{van2015building}& \footnotesize --- & \footnotesize 13 to 120 & \footnotesize 48562 & \footnotesize 555 \\ \bottomrule
\end{tabular}
\label{tab:datasets}
\end{adjustwidth}
\end{table}

\paragraph{Models} %
We consider four types of models: the softmax classifier, NormFace, ProxyDR (explained below), and a CORR loss-based model. We focus on metric learning models with normalized embeddings, as normalization is commonly used in metric learning models to improve performance \cite{wang2017normface, movshovitz2017no, barz2019hierarchy}. ProxyDR is a model that uses proxies for classification and DR formulations (\ref{eq:DR_form}) to estimate class probabilities \(p(c|x)\). Similar to the NormFace model, ProxyDR uses the Euclidean distance on a hypersphere. %
The difference %
between the two models is that ProxyDR uses DR formulations, while NormFace uses SD softmax formulations (as shown in Eq. \ref{eq:normface_SD}). %
For the plankton datasets, we used a pretrained Inception version 3 architecture %
\cite{szegedy2016rethinking} as the backbone. %
For the CIFAR100 and NABirds datasets, we used a pretrained ResNet50 \cite{he2016deep} as the backbone. In addition to the backbones, we applied a learnable linear transformation %
to obtain \(128\)-dimensional embeddings \(f(x)\). %
For the plankton datasets, we incorporate the size information. Specifically, when a
data point \(x\) has size \(v_{size}=\left [width, height\right ]^T\), we take the elementwise logarithm \(v'_{size}=\left [ \log (width), \log (height)\right ]^T\). %
By applying a linear transformation, %
we obtain an embedding vector \(f_{size}(x)\) according to the size information, i.e., \(f_{size}(x)=W_{size}^T v'_{size}+b_{size}\), where \(W_{size}\) is a learnable matrix with shape \(2\times128\) and \(b_{size}\in \mathbb{R}^2\) is a learnable vector. Then, we add this vector to the original embedding vector, namely, \(f(x):=f(x)+f_{size}(x)\). %

\paragraph{Training settings}

We trained the models according to the backbone weights and other weights (linear transformations, proxies). We used the Adam optimizer \cite{kingma2014adam} with a learning rate of \(10^{-4}\). %
Except in the case of a dynamic approach (explained below), we use \(10.0\) as a scaling factor in both NormFace and ProxyDR. We set the training batch size in all experiments to \(32\). For the plankton datasets, we trained the models for \(50\) epochs. For the CIFAR100 and NABirds datasets, we trained the models for \(100\) epochs. During each epoch, we assessed the model accuracy. We chose the model with the highest validation accuracy for testing. For each setting, we trained the models five times with different seeds for the random split.\\%}

\paragraph{Training options}
The different training %
options are described as follows. %
The dynamic approach affects the scale factors in Eqs. \ref{eq:normface} and \ref{eq:DR_form}. The EMA and MDS approaches both affect the proxy calculations.

\begin{itemize}
\item Standard: standard training with a fixed scale factor \(s=10\), with proxies updated %
using the cross-entropy loss (\ref{eq:CE_loss}).
\item EMA (exponential moving average): training using the normalized exponential moving average \cite{zhe2019improve}, as shown in Eq. \ref{eq:EMA}, to update the proxies. %
When multiple data points have the same class \(c\) in a mini-batch, we apply a modified expression.
Specifically, instead of applying the EMA using a single data point, as in Eq. \ref{eq:EMA}, %
we use the normalized average embedding (local prototype) of the mini-batch. %
Mathematically, when %
updating a proxy for \(m\) data points \(x_{B;1}, \cdots, x_{B;m}\) with class \(c\) in mini-batch \(B\), the normalized average embedding is defined as:
\begin{align*}
    \Tilde{\mu}_{B;c}=\frac{\sum\limits_{i=1}^{m}{\Tilde{f}(x_{B;i})}}{\left\| \sum\limits_{i=1}^{m}{\Tilde{f}(x_{B;i})}\right\|}.
\end{align*}
Then, proxy \(\Tilde{W}_c\) is updated as:
\begin{align}\label{eq:EMA_modified}
    \Tilde{W}_c :=\frac{\alpha \Tilde{\mu}_{B;c}+(1-\alpha) \Tilde{W}_c}{\left\|\alpha \Tilde{\mu}_{B;c}+(1-\alpha) \Tilde{W}_c \right\|},
\end{align}
where  %
\(\alpha\) is the same parameter as in Eq. \ref{eq:EMA}. %
We set the parameter \(\alpha\) to \(0.001\).\\
\item Dynamic: training with a dynamic scale factor, similar to AdaCos \cite{zhang2019adacos}. In contrast to the original paper, which chooses a scale factor using an approximate expression, %
we use the Adam \cite{kingma2014adam} optimizer to determine a scale factor that satisfies Eq. \ref{eq:adacos}. More details are included in \nameref{S1_appendix}. %
\item MDS (multidimensional scaling): training according to predefined hierarchical information. We use the hierarchical information to set (fixed) proxies. First, as in the distance calculation method in the problem setting subsection, %
we use a predefined hierarchy to generate a distance matrix. %
We denote the (hierarchical) distance between the \(i\)th and \(j\)th classes as \(d_H (i,j)\). %
For each hierarchical distance \(d_H\), we apply a transformation \(d_T=\frac{\sqrt{2} d_H}{\beta+d_H}\) for a scalar \(\beta>0\) to address the limited Euclidean distance (\(\le\sqrt{2}\)) on unit spherical spaces. We set \(\beta=1.0\) in all of our experiments. (Note that if \(d\) is a metric, the transformed distance \(\frac{\sqrt{2} d}{\beta+d}\) is also a metric.) %
Then, according to the transformed distance matrix \(D_T\), we use multidimensional scaling (MDS) to set the proxies. Mathematically, %
we minimize a value known as the normalized stress, which can be expressed as:\\
\begin{align}\label{eq:stress}
    \text{Stress}(D_{\Tilde{W}}):=\frac{\left\|D_{\Tilde{W}} -D_T \right\|_F}{\left\| D_T\right\|_F},
\end{align}
where \(D_{\Tilde{W}}\) is a pairwise Euclidean distance matrix according to proxies \(\Tilde{W}_y\) and \(\left\| \cdot \right\|_F\) is the Frobenius norm. We use the Adam optimizer with a learning rate of \(10^{-3}\) for 1000 iterations to obtain the proxies. While we used stochastic gradient descent for the MDS option, different methods, such as those applied by Barz and Denzler \cite{barz2019hierarchy}, can also be used for MDS. During training, we fix the obtained proxies and update only the embedding function \(f(\cdot)\). %
\end{itemize}

\paragraph{Performance measures}
In our experiments, we consider three types of performance measures: standard classification measures, hierarchical inference performance measures, and hierarchy-informed performance measures. %
The hierarchical inference performance measures are used to estimate how well the learned class representatives match the predefined hierarchies. %
The hierarchy-informed performance measures %
are used to estimate how well a model performs on classification or similarity measures according to the predefined hierarchies.\\%... To analyze the class representatives (proxies and prototypes), we consider the following measures: mean correlations (MC), %
We used the top-k accuracy as a standard classification measure, the mean correlation as the hierarchical inference performance measure, and the average hierarchical distance (AHD), hierarchical precision at \(k\) (HP@k), hierarchical similarity at \(k\) (HS@k), and average hierarchical similarity at \(k\) (AHS@k) as hierarchy-informed performance measures. %
The utilized measures are defined as follows.
\begin{itemize}
\item Top-\(k\) accuracy: The classification accuracy was calculated, with correct classification defined as whether the labeled class is in the top-\(k\) predictions (most likely classes), %
i. e., the \(k\) classes with the highest confidence values. %
We report results for \(k=1, 5\).
\item Mean correlations: We introduce this measure to evaluate how well the learned class representatives match the predefined hierarchical structure. We obtain the class representatives (either proxies or prototypes) from the learned model using training data points. Then, we obtain the pairwise distance matrix \(D_{L}\) based on the class representatives. %
We compare matrix \(D_{L}\) with the distance matrix \(D_{H}\) based on the predefined hierarchical structure.
Specifically, for each class (row), we calculate Spearman's rank correlation coefficient according to the two matrices. Then, we determine the mean correlation using Fisher transformations. More precisely, we apply a Fisher transformation \(\arctanh (\cdot)\) on each correlation coefficient, take the average of the transformed values, and apply \(\tanh (\cdot)\) on the average value. \\
For the plankton datasets, the predefined hierarchical structures of the living classes are based on biological taxonomies. However, these datasets also contain some nonliving classes, such as ``large bubbles'' and ``dark debris''. %
Considering that the predefined hierarchical structures of the living classes are scientifically defined, %
for the plankton datasets, we report mean correlations among whole classes or only among living classes. %
\item AHD: The average hierarchical distance of the top-\(k\) predictions \cite{bertinetto2020making} was calculated as the average hierarchical distance \(d_H\) between the labeled classes and each of the top-\(k\) most likely classes. 
In contrast to Bertinetto et al. \cite{bertinetto2020making}, who considered only misclassified cases, we consider all cases. Hence, in our calculations, the final denominators differ.
When \(k=1\), the AHD measure %
is the same as %
the average hierarchical cost (AHC) measure %
defined by Garnot and Landrieu \cite{garnot2021leveraging}. We report results for \(k=1, 5\).
\item HP@\(k\): The hierarchical precision at \(k\) \cite{frome2013devise} was taken as a performance measure. Specifically, let us denote a (hierarchical) neighborhood set of a class \(c\) with the distance threshold \(\epsilon\) as \(N(c,\epsilon)\), i.e., \(n\in N(c,\epsilon) \Longleftrightarrow   d_{H}(c,n)\le \epsilon\). We define \(\text{hCorrectSet}(c,k)\) as the neighborhood set \(N(c,\epsilon)\) with the smallest \(\epsilon\) such that \(|\text{hCorrectSet}(c,k)|\ge k\). Then, the hierarchical precision at \(k\) is calculated as the fraction of the top-\(k\) predictions in \(\text{hCorrectSet}(c,k)\). %
We report the results for \(k= 5\).
\item HS@\(k\): 
The hierarchical similarity at \(k\) is a measure that was introduced by Barz and Denzler \cite{barz2019hierarchy} with the name “hierarchical precision at \(k\)”, although this metric does not evaluate precision and instead assesses similarity. Here, we use a different measure with the same name that was defined by Frome et al. \cite{frome2013devise}. 
Hence, we renamed the measure ``hierarchical similarity at \(k\)''. When \(c\) is a label of a query data point \(x\), let \(R=\left ( \left ( x_1,c_1 \right ), \cdots, \left ( x_m, c_m \right ) \right )\) be the ordered list of image-label pairs based on the distance (sorted by ascending distance) to point \(x\) in the normalized embedding space. Considering \(\cos(\theta)=1-\frac{\left\| u_1-u_2\right\|^{2}}{2}\), where \(u_1\) and \(u_2\) are unit vectors and \(\theta\) is the angle between \(u_1\) and \(u_2\), we defined the similarity between the \(i\)th and \(j\)th classes \(s_H (i,j)\) as: 
\begin{align}\label{eq:similarity}
    s_H (i,j) =1-\frac{d_T (i,j)^{2}}{2},
\end{align}
where \(i\) and \(j\) are the indices for classes (\(1\le i,j\le|\mathcal{Y}|\)).\\
The hierarchical similarity at \(k\) is then defined as:
\begin{align}\label{eq:HS@k}
    HS@k := \frac{\sum\limits_{i=1}^{k}{s_H (I(c),I(c_i))}}{\max_{\pi} \sum\limits_{i=1}^{k}{s_H (I(c),I(c_{\pi_i}))}},
\end{align}
where \(I(\cdot)\) is an index function that outputs the corresponding index (between \(1\) and \(|\mathcal{Y}|\)) for a class and \(\pi\) is an index permutation that ranges from \(1\) to \(m\). We report results for \(k=50, 250\).
\item AHS@\(K\): The average hierarchical similarity at \(K\) was introduced by Barz and Denzler \cite{barz2019hierarchy} as the ``average hierarchical precision at \(K\)''. Due to similar reasons as for HS@\(k\), we renamed the measure. The average hierarchical similarity at \(K\) is defined as the area under the curve of HS@\(k\) from \(k=1\) to \(k=K\). We report results for \(K=250\).
\end{itemize}%

\section*{Results}

\subsection*{Main results}
\begin{figure}[H]
    \centering
\vspace{.05in}
\includegraphics[width=1.0\linewidth,height=0.4\linewidth]{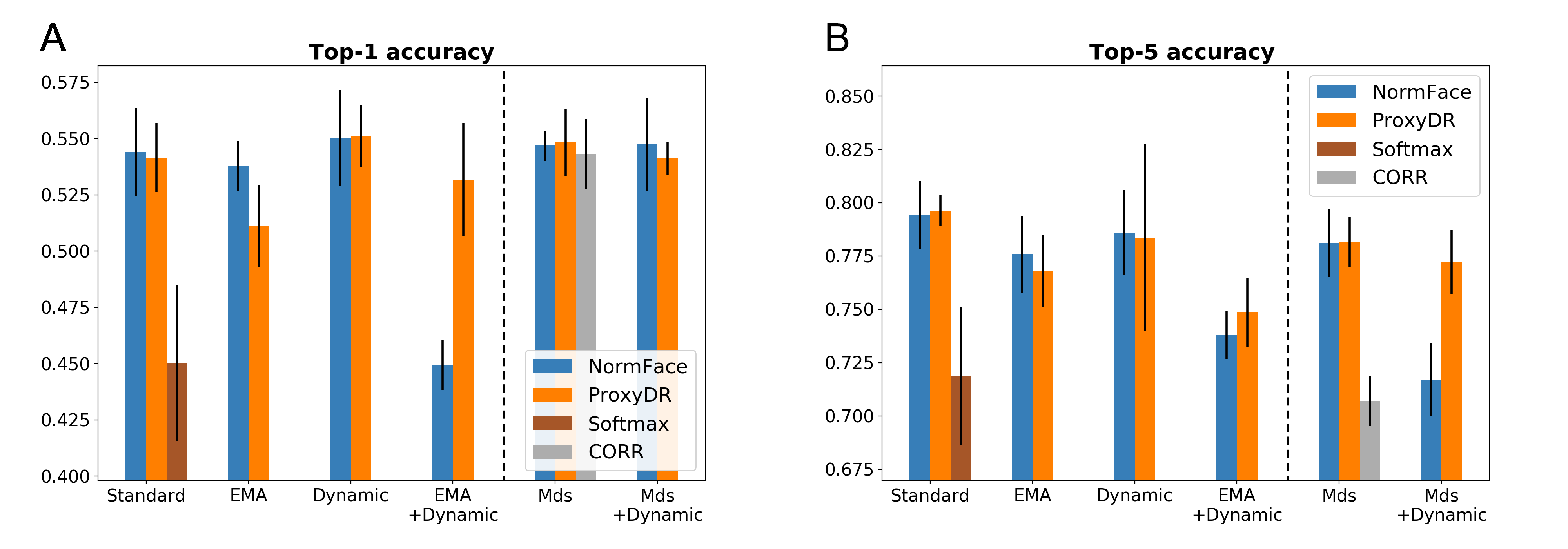}
\caption{{\bf Top-\(k\) accuracy results (A: \(k=1\), B: \(k=5\)) on the CIFAR100 dataset.} }
    \label{fig:Fig3}%
\end{figure}

\begin{figure}[H]
    \centering
\vspace{.05in}
\includegraphics[width=1.0\linewidth,height=0.4\linewidth]{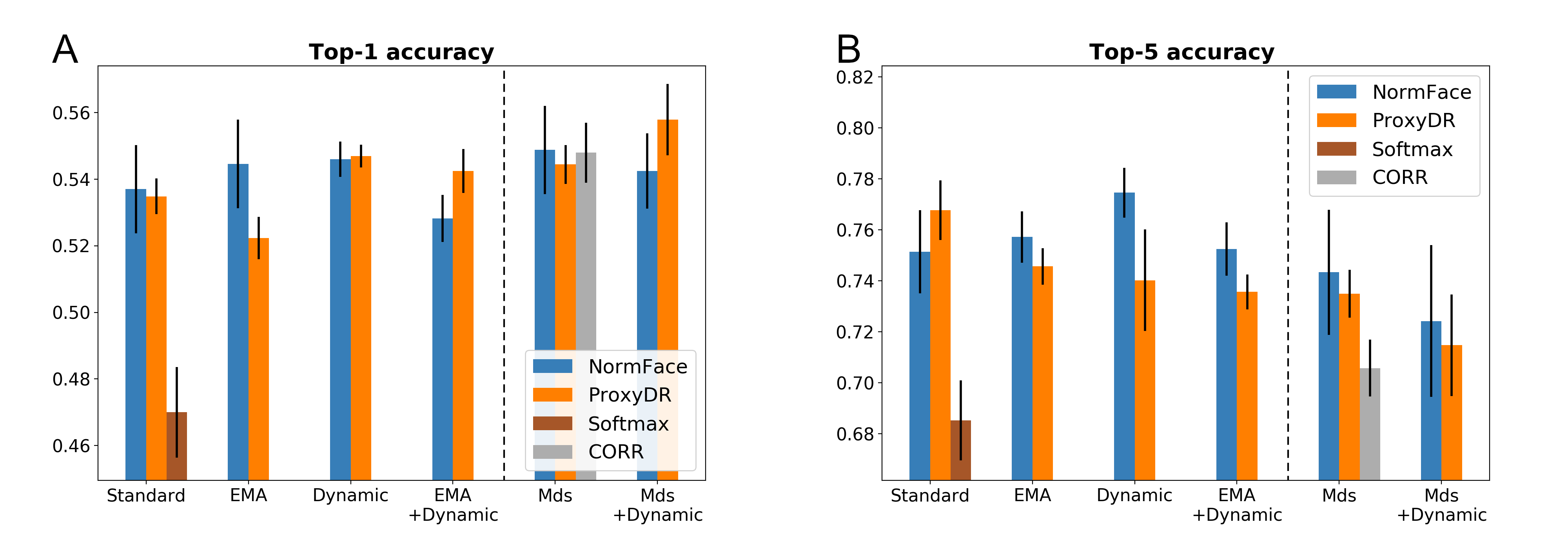} 
\caption{{\bf Top-\(k\) accuracy results (A: \(k=1\), B: \(k=5\)) on the NABirds dataset.} }
    \label{fig:Fig4}%
\end{figure}

The performance measure evaluation results are shown in bar plots with 95\% confidence intervals. Dashed lines separate models trained with or without predefined hierarchical knowledge. We show only the results on the CIFAR100 and NABirds datasets in the main text. %
All results are included in \nameref{S3_appendix}. %
To reduce spurious findings, %
we focus on consistent trends across %
the five datasets. \\%from the results.
Figs. \ref{fig:Fig3} %
and \ref{fig:Fig4} %
and the figures in \nameref{S3_appendix} %
show the top-\(k\) accuracy for various training settings. The NormFace and ProxyDR models achieved comparable top-\(k\) accuracy for both \(k\) values on most datasets. The softmax loss model obtained low top-\(k\) accuracy. %
While the result was not significant, using the dynamic option %
achieved higher top-1 accuracy than standard training. When the EMA option was added to the standard and dynamic options, the top-5 accuracy decreased, except for standard NormFace on the NABirds dataset. %
While the CORR loss achieved top-1 accuracy that was comparable to that achieved by other training options with predefined hierarchical information, this model obtained low top-5 accuracy on all datasets. The top-5 accuracy with the CORR loss was even lower than that with the softmax loss, except on the NABirds dataset. %
Although the results were better than those of the CORR loss model, the use of predefined hierarchical information during training for NormFace and ProxyDR also reduced the top-5 accuracy. %
These results show the opposite trend to the changes in the top-1 accuracy, which showed comparable or enhanced performance. 
Moreover, the dynamic MDS approach obtained lower top-5 accuracy than the MDS approach without the dynamic option.

\begin{figure}[H]
    \centering
\vspace{.05in}
\includegraphics[width=1.0\linewidth,height=0.4\linewidth]{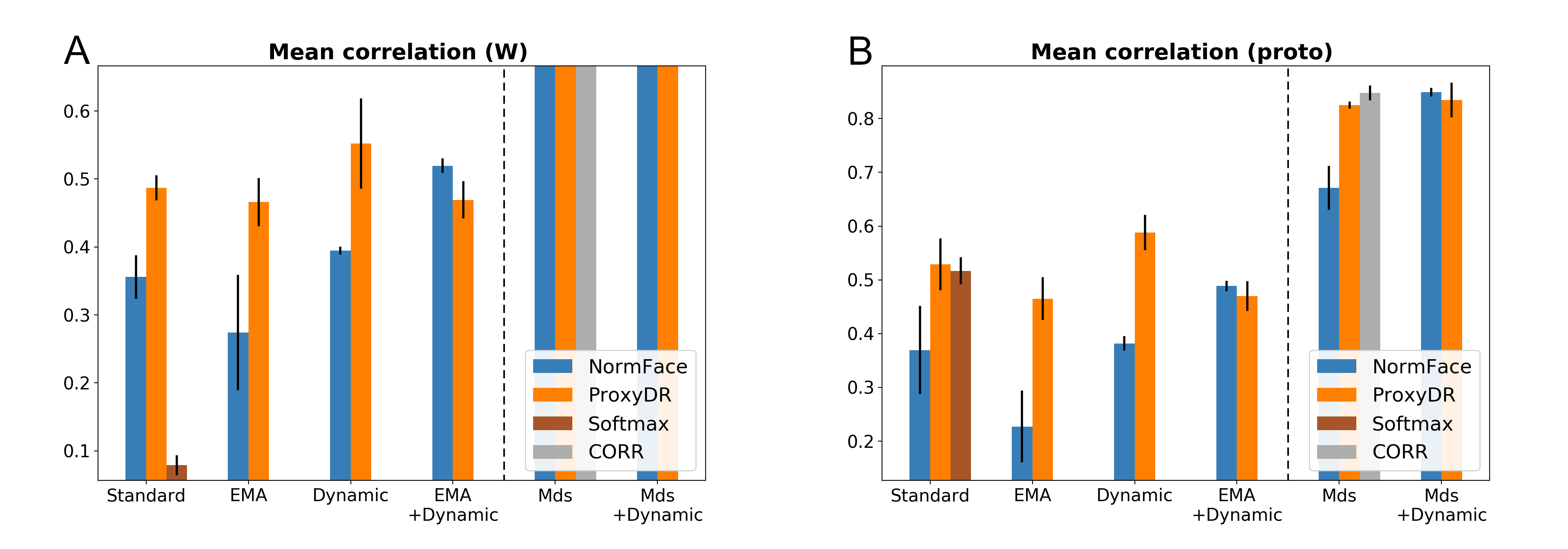} 
\caption{{\bf Correlation measures on the CIFAR100 dataset.} (A) Values using proxies. (B) Values using prototypes. The mean correlation value based on proxies with the MDS option %
was \(0.8580\). }
    \label{fig:Fig5}%
\end{figure}

\begin{figure}[H]
    \centering
\vspace{.05in}
\includegraphics[width=1.0\linewidth,height=0.4\linewidth]{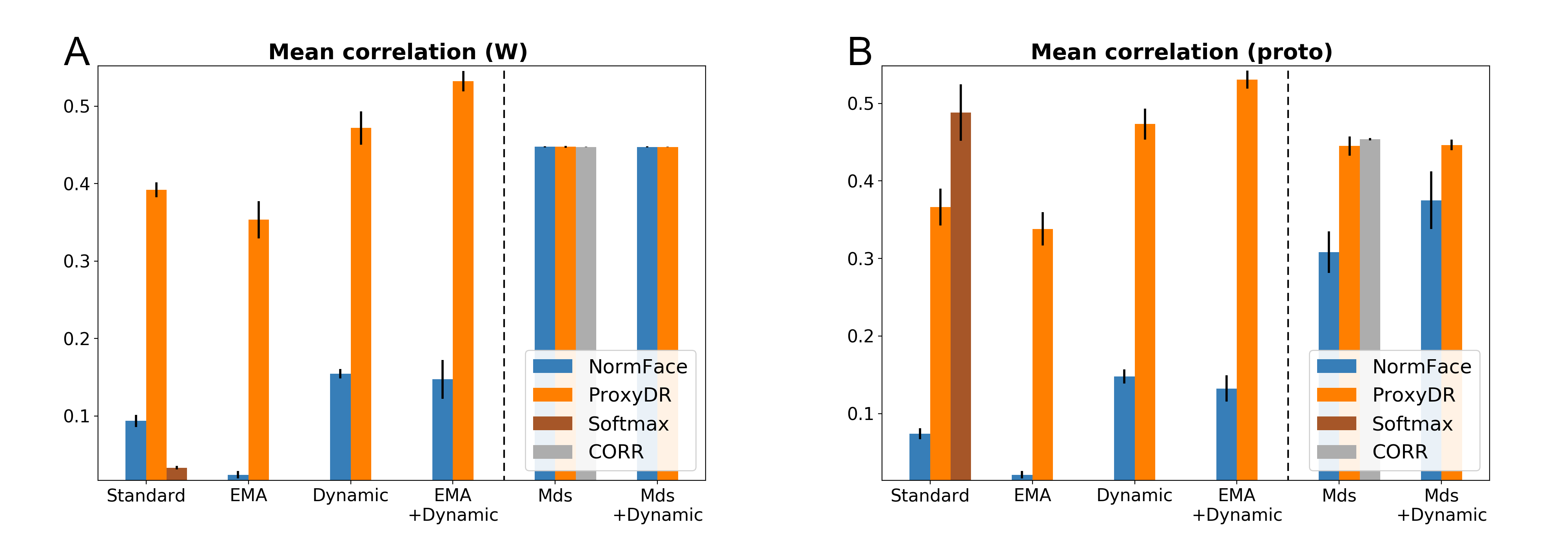} 
\caption{{\bf Correlation measures on the NABirds dataset.} (A) Values using proxies. (B) Values using prototypes. The mean correlation value based on proxies with the MDS option %
was \(0.4476\) (this value is small because the dataset contains 555 classes and the embedding dimension is 128.). }
    \label{fig:Fig6}%
\end{figure}

Figs. \ref{fig:Fig5} %
and \ref{fig:Fig6} %
and the figures in \nameref{S3_appendix} %
show the mean correlation values for various training options. When we consider training options that do not use predefined hierarchical information, ProxyDR obtains higher mean correlations than NormFace, except for the EMA approaches. %
The ProxyDR model with the dynamic option obtained higher mean correlations than the standard ProxyDR model.
As expected, the use of predefined hierarchical information greatly increased the mean correlations based on prototypes in both the NormFace and ProxyDR models, except the ProxyDR model on the NABirds dataset. When we consider training options that utilize predefined hierarchical information, the CORR loss model achieved the highest mean correlations based on prototypes in most cases. The ProxyDR model typically achieved the second-best mean correlations. %

\begin{figure}[H]
    \centering
\vspace{.05in}
\includegraphics[width=1.0\linewidth,height=1.0\linewidth]{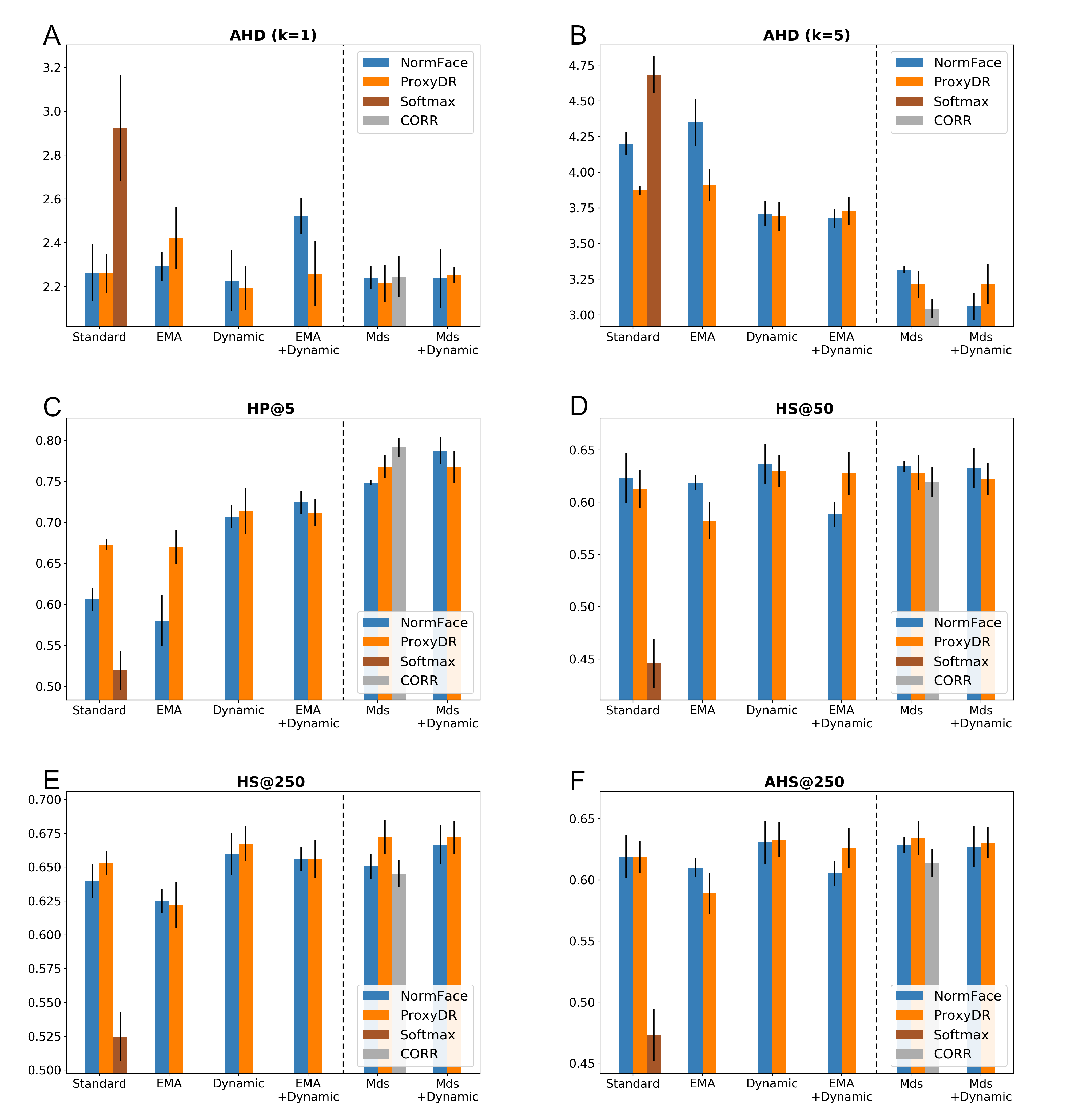}
\caption{{\bf Hierarchical performance measures on the CIFAR100 dataset.} The symbol \(\downarrow\) denotes that lower values indicate better performance. %
The symbol \(\uparrow\) denotes that higher values indicate better performance. %
(A) AHD (k=1): \(\downarrow\). (B) AHD (k=5): \(\downarrow\). (C) HP@5: \(\uparrow\). (D) HS@50: \(\uparrow\). (E) HS@250: \(\uparrow\). (F) AHS@250: \(\uparrow\). %
}
    \label{fig:Fig7}%
\end{figure}

\begin{figure}[H]
    \centering
\vspace{.05in}
\includegraphics[width=1.0\linewidth,height=1.0\linewidth]{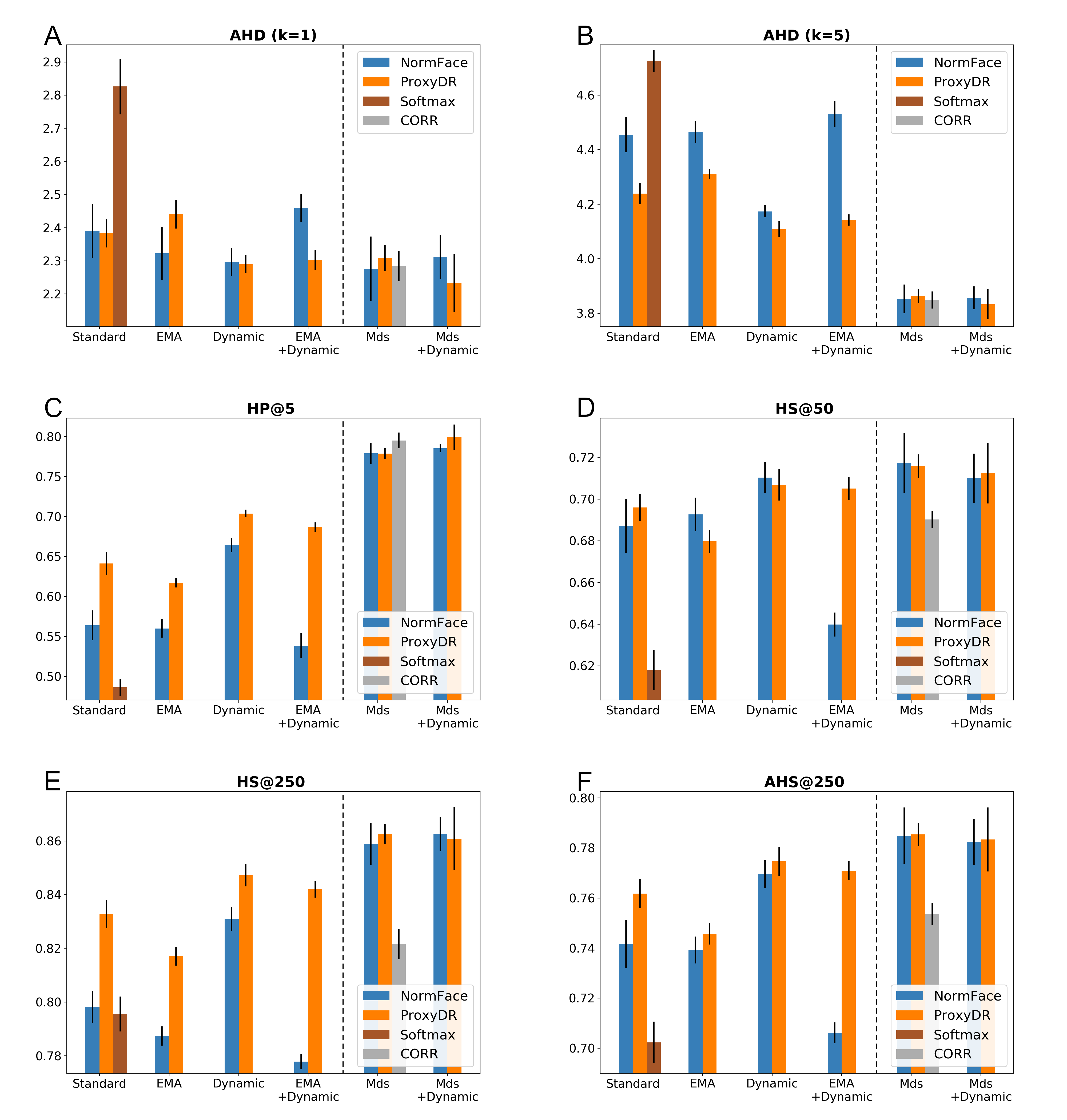}
\caption{{\bf Hierarchical performance measures on the NABirds dataset.} The symbol \(\downarrow\) denotes that lower values indicate better performance. %
The symbol \(\uparrow\) denotes that higher values indicate better performance. %
(A) AHD (k=1): \(\downarrow\). (B) AHD (k=5): \(\downarrow\). (C) HP@5: \(\uparrow\). (D) HS@50: \(\uparrow\). (E) HS@250: \(\uparrow\). (F) AHS@250: \(\uparrow\). %
}
    \label{fig:Fig8}%
\end{figure}

Figs. \ref{fig:Fig7} %
and \ref{fig:Fig8} %
and the figures in \nameref{S3_appendix} %
show the hierarchical performance measures obtained with various training options. %
The softmax loss option achieved the worst hierarchical performance, %
except for AHS@\(50\) on the NABirds dataset. Although these %
measures are used to assess the hierarchy-informed performance, %
some of the measures are not substantially affected by the use of predefined %
hierarchical information during training. For instance, the use of predefined hierarchical information in the CIFAR100 dataset (Fig. \ref{fig:Fig7}) %
did not show noticeable improvements in terms of the AHD (k=1), HS@\(50\), %
and AHS@\(250\) measures. %
Moreover, while the use of predefined hierarchical information significantly improved the HS@\(250\) results on most datasets, only marginal improvements were observed for the CIFAR100 dataset. %
On the other hand, the use of predefined hierarchical information significantly improved the AHD (k=5) and HP@\(5\) results on all datasets. The CORR loss achieved the best results on these two measures, except on the NABirds dataset. Adding the dynamic option to the standard and MDS options improved performance on these two measures, except for the ProxyDR model on the CIFAR100 dataset. %
When we consider training options that do not use predefined hierarchical information, the ProxyDR model shows better performance than NormFace in terms of these two measures, except for the EMA and dynamic options on the CIFAR100 dataset. %
Under the same settings, ProxyDR performed better than NormFace in terms of the HS@\(250\) measure, except for the EMA option on the CIFAR100 dataset. %
Moreover, under the same settings, the ProxyDR model with %
the dynamic option achieved the highest HS@\(250\) and AHS@\(250\) values among the compared %
models. %

\subsection*{Additional mean correlation results}
\begin{figure}[H]
    \centering
\vspace{.05in}
\includegraphics[width=1.0\linewidth,height=0.7\linewidth]{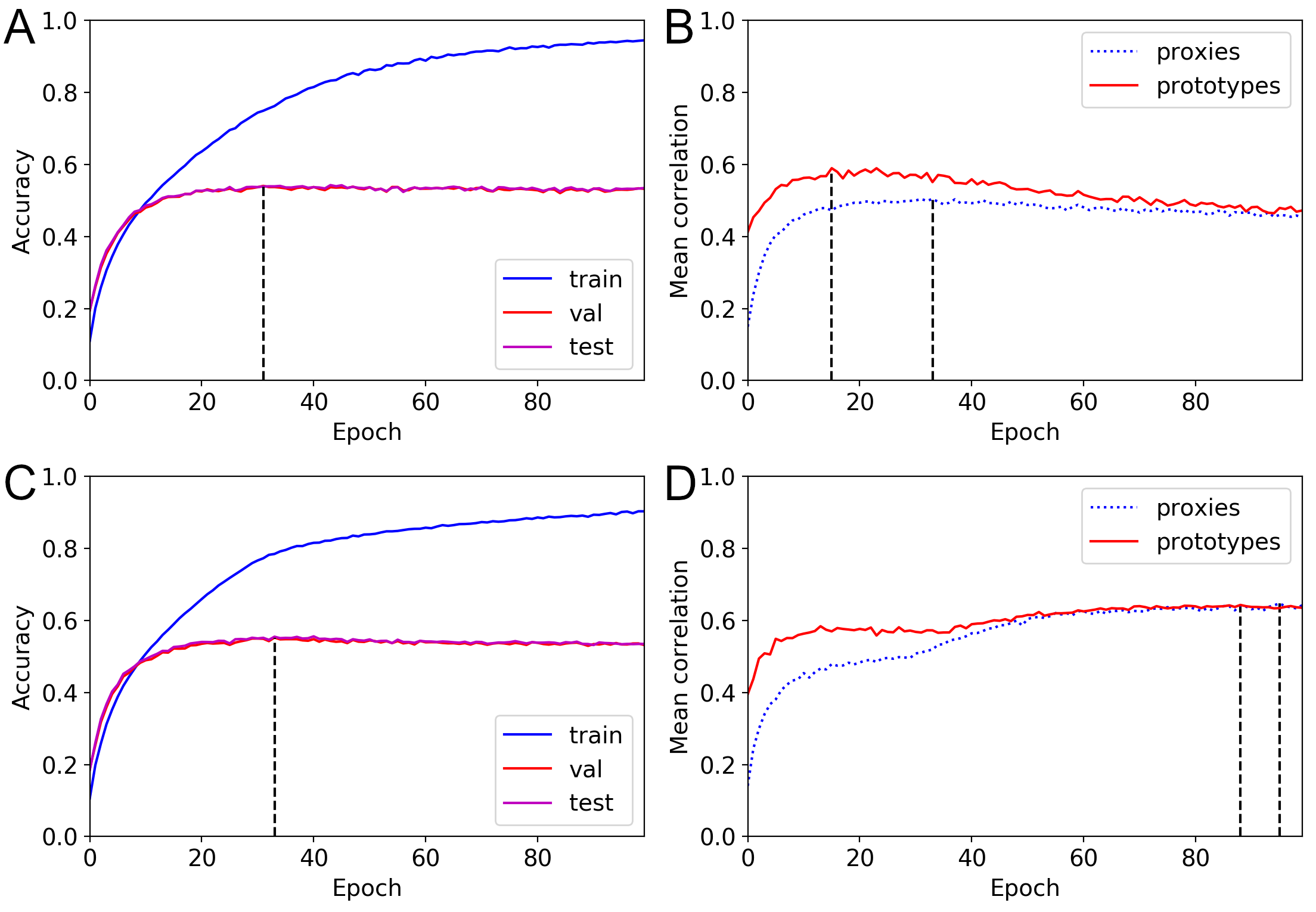} %
\caption{{\bf Changes in accuracy and mean correlations %
for the CIFAR100 dataset (ProxyDR).} %
(A) Accuracy curve with standard training. (B) Mean correlation curve with standard training. (C) Accuracy curve with the dynamic option. (D) Mean correlation curve with the dynamic option. %
}
    \label{fig:Fig9}%
\end{figure}

\begin{figure}[H]
    \centering
\vspace{.05in}
\includegraphics[width=1.0\linewidth,height=0.7\linewidth]{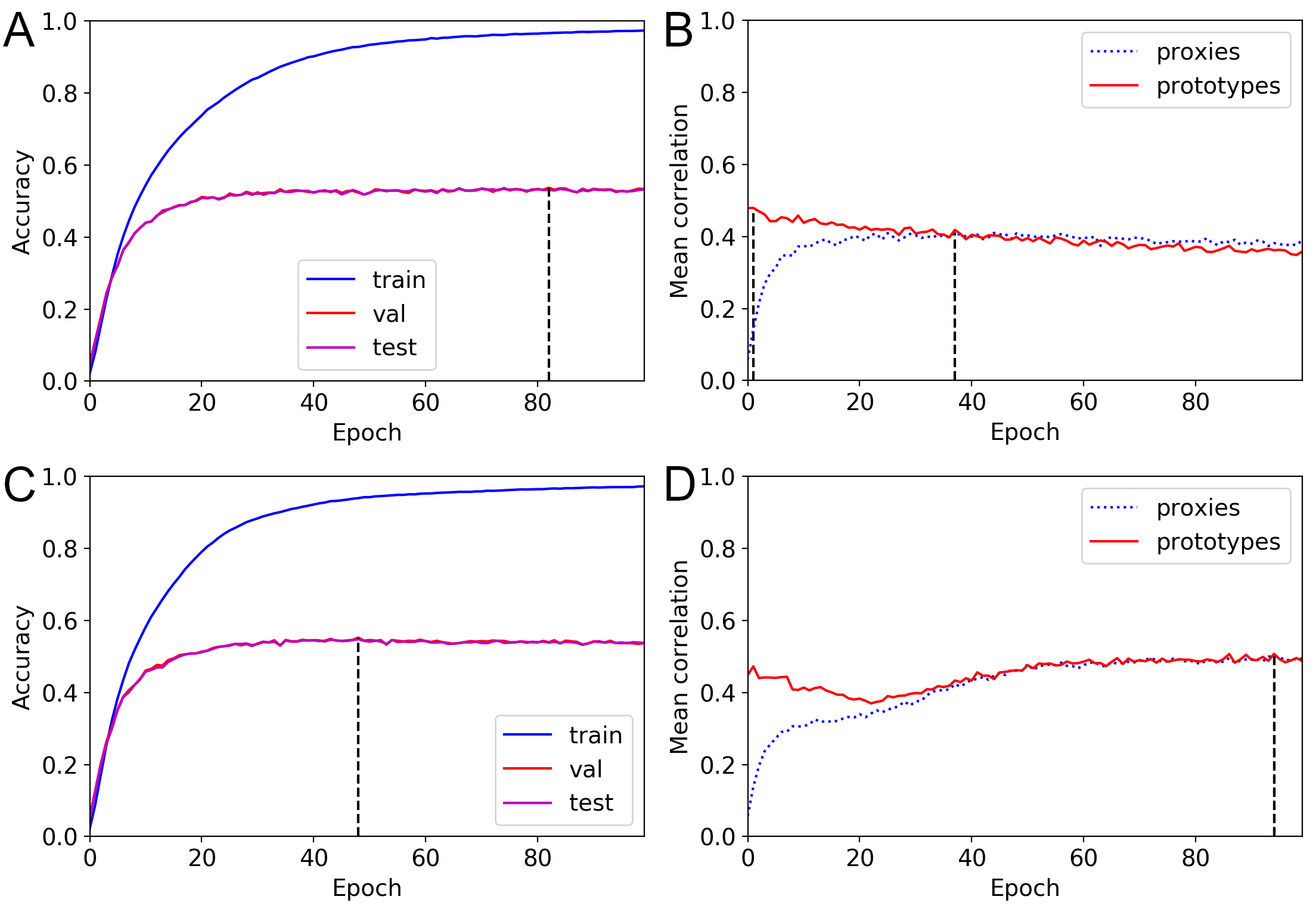} %
\caption{{\bf Changes in accuracy and mean correlations %
for the NABirds dataset (ProxyDR).} %
(A) Accuracy curve with standard training. (B) Mean correlation curve with standard training. (C) Accuracy curve with the dynamic option. (D) Mean correlation curve with the dynamic option.%
}
    \label{fig:Fig10}%
\end{figure}

To investigate the changes in the class representatives, we %
evaluated the mean correlations at the end of each training epoch. 
We report only the results on the CIFAR100 and NABirds datasets. %
Moreover, we report results for ProxyDR models with the standard and dynamic options. %
More results are included in %
\nameref{S3_appendix}. %
Furthermore, we report mean correlations based on random networks, i.e., networks with random weights, pretrained networks, and unnormalized and normalized input spaces. \\ %
Figs. \ref{fig:Fig9} %
and \ref{fig:Fig10} %
visualize the changes in the mean correlation values %
during ProxyDR model training (averaged values from five different seeds). %
Surprisingly, we found that the prototypes of the approximately untrained networks (pretrained on ImageNet \cite{deng2009imagenet} and trained on the target dataset, e.g., CIFAR100 or NABirds, for only one epoch) 
already have relatively high correlations (approximately \(0.4\)), with predefined hierarchical structures. %
While the accuracy curves show no noticeable differences, using the dynamic option modified the %
transition in the mean correlations.
In particular, prototype-based mean correlations increased after 20 to 40 training epochs, and the maximum values were obtained near the end of training (approximately 100 epochs). 
The proxy-based mean correlations started at low values, and the difference with the prototype-based mean correlations was reduced. 
Moreover, the training epoch during which the validation accuracy is maximized often differs from the training epoch during which the correlation measures are maximized. %

Table \ref{tab:corr_proto} shows prototype-based mean correlation values and their 95 percent confidence intervals based on five different seeds. 
The mean correlations with the random networks show that the prototypes and ground-truth hierarchy are correlated. While these values are smaller than the other cases shown in the table, the results show that random networks have some degree of semantic understanding. %

\begin{table}[H]
\begin{adjustwidth}{-2.25in}{0in} %
\centering
\caption{
{\bf Mean correlations based on prototypes.} }
\begin{tabular}{\Mc{2.0cm}|\Mc{3.25cm}|\Mc{3.25cm}|\Mc{3.25cm}|\Mc{3.25cm}}
 \footnotesize Space & \footnotesize Random weights & \footnotesize Pretrained & \footnotesize Input space & \footnotesize Normalized input space \\
          \toprule
\small CIFAR100 & \footnotesize \(0.2841\pm0.0303\) & \footnotesize \(0.3816\pm0.0116\) & \footnotesize \(0.3414\pm0.0353\)& \footnotesize \(0.2609\pm0.0416\)\\ \midrule
\small NABirds & \footnotesize \(0.1088\pm0.0839\) & \footnotesize \(0.2648\pm0.0349\) & \footnotesize \(0.1740\pm0.0056\)& \footnotesize \(0.2369\pm0.0039\)\\ 
          \bottomrule
\end{tabular}
\begin{flushleft} ``Random weights" and ``pretrained" mean embedding space based on the ResNet50 \cite{he2016deep} backbone.
\end{flushleft}
\label{tab:corr_proto}
\end{adjustwidth}
\end{table}

\section*{Discussion}
In this work, we investigate classification and hierarchical %
performance under different models and training options. %
Our experiments reveal several important findings. Under the training options %
that do not consider predefined hierarchical information, the ProxyDR model achieved better hierarchical inference performance %
than NormFace in most cases. %
Furthermore, %
under the same training options,
ProxyDR achieved better %
hierarchy-informed performance in terms of the AHD (k=5) and HP@\(5\) measures. %
Moreover, we observed that the use of a dynamic scaling factor improved %
the hierarchical inference performance. The changes in the mean correlation values (Figs. \ref{fig:Fig9} %
and \ref{fig:Fig10}) %
verified the effect of the dynamic training option. These results reveal the importance of dynamic training approach. %
We also found that some hierarchy-informed performance measures are not significantly improved by the use of known hierarchical structures. This finding indicates that multiple hierarchy-informed performance measures should be considered to compare the hierarchy-informed performance %
of different models. %
We also observed a trade-off between the hierarchy-informed performance and top-5 accuracy. %
While the CORR loss model typically achieved the best hierarchical performance, this model obtained the lowest top-5 accuracy among the experimental models. 
Similarly, the use of predefined hierarchical information in NormFace and ProxyDR significantly improved the AHD (k=5) and HP@\(5\) performance but reduced the top-5 accuracy. 
In contrast to previous works that observed a trade-off between hierarchy-informed performance and top-1 accuracy \cite{bertinetto2020making}, we did not observe this trade-off with top-1 accuracy. \\ %
Surprisingly, we found that %
prototypes based on %
CNNs with random weights showed correspondence with predefined hierarchies that was higher than random chance. %
Because CNNs combine convolutional and pooling layers, most CNN architectures have translation invariance properties. %
Because of such priors, %
random networks may know weak perceptual similarity \cite{zhang2018unreasonable}. %
Another possible reason for this result is that prototypes of input spaces show higher correspondence with predefined hierarchies than random chance. The Johnson-%
Lindenstrauss lemma \cite{johnson1984extensions} shows %
that linear projections using random matrices approximately preserve distances. If this property holds for nonlinear projections based on random neural networks, %
network-based prototypes may also correspond to predefined hierarchies. %
This %
phenomenon suggests that when we use metric learning models with proxies, %
the proxies can %
be assigned based on prototypes instead of starting at random positions. %
This may improve training during the initial epochs, as we start from proxies that are more semantically reasonable than random positions. %

Although we observed that the DR formulation improves the hierarchical inference performance and hierarchy-informed performance when training models without predefined hierarchies, we did not study the reasons underlying these phenomena. 
We suggest one possible hypothesis.
While NormFace prevents sudden changes in the scaling factor by using normalized embeddings, its loss function is based on the squared difference of the distance (as it uses the SD softmax formulation). %
As this loss can increase the squared difference of the distance among %
different proxies, the absolute distance between any pairs of proxies may be increased. %
This tendency can result in larger distances, even among semantically similar classes, and proxy positions that are less organized in terms of visual similarity. %
On the other hand, the DR formulation-based loss is based on distance ratios. Thus, %
there is no tendency to increase absolute distances among proxies, and proxies can be structured %
according to visual similarity. %
However, further investigations are needed to verify this hypothesis and reveal the underlying cause.\\%}

\section*{Conclusion}
The hierarchy-informed performance must be improved to more broadly adopt classification models. We explored this concept with classification-based metric learning models in situations in which hierarchical information is and is not available during training. %
Our results show that when the class hierarchical relations are unknown, the ProxyDR model achieves the best hierarchical inference and hierarchy-informed performance. In contrast, with hierarchy-informed training, %
the CORR loss model achieves the best hierarchy-informed performance but the lowest top-5 accuracy on most datasets. Since some hierarchy-informed measures may not be improved by the use of hierarchical information during training, multiple hierarchy-informed performance measures should be used to obtain appropriate comparisons. %
Additionally, our experiments reveal that during classification-based metric learning, initializing proxies based on prototypes may be beneficial.

\section*{Acknowledgments}
Hyeongji Kim, Terje Berge, and Ketil Malde acknowledge the Ministry of Trade, Industry and Fisheries for financial support. The authors thank Gayantonia Franze, Hege Lyngv\ae r Mathisen, Magnus Reeve, and Mona Ring Kleiven, from the Plankton Research Group at Institute of Marine Research (IMR), for their contributions to the plankton datasets.

\nolinenumbers

\bibliography{My_library} %

\begin{thebibliography}{10}

\bibitem{he2015delving}
He K, Zhang X, Ren S, Sun J.
\newblock Delving deep into rectifiers: Surpassing human-level performance on
  imagenet classification.
\newblock In: Proceedings of the IEEE international conference on computer
  vision; 2015. p. 1026--1034.

\bibitem{russakovsky2015imagenet}
Russakovsky O, Deng J, Su H, Krause J, Satheesh S, Ma S, et~al.
\newblock Imagenet large scale visual recognition challenge.
\newblock International journal of computer vision. 2015;115(3):211--252.

\bibitem{deng2009imagenet}
Deng J, Dong W, Socher R, Li LJ, Li K, Fei-Fei L.
\newblock Imagenet: A large-scale hierarchical image database.
\newblock In: 2009 IEEE conference on computer vision and pattern recognition.
  Ieee; 2009. p. 248--255.

\bibitem{bertinetto2020making}
Bertinetto L, Mueller R, Tertikas K, Samangooei S, Lord NA.
\newblock Making better mistakes: Leveraging class hierarchies with deep
  networks.
\newblock In: Proceedings of the IEEE/CVF Conference on Computer Vision and
  Pattern Recognition; 2020. p. 12506--12515.

\bibitem{barz2019hierarchy}
Barz B, Denzler J.
\newblock Hierarchy-based image embeddings for semantic image retrieval.
\newblock In: 2019 IEEE Winter Conference on Applications of Computer Vision
  (WACV). IEEE; 2019. p. 638--647.

\bibitem{garnot2021leveraging}
Garnot VSF, Landrieu L.
\newblock Leveraging Class Hierarchies with Metric-Guided Prototype Learning.
\newblock In: British Machine Vision Conference (BMVC); 2021.

\bibitem{jayathilaka2021ontology}
Jayathilaka M, Mu T, Sattler U.
\newblock Ontology-based n-ball concept embeddings informing few-shot image
  classification.
\newblock arXiv preprint arXiv:210909063. 2021;.

\bibitem{musgrave2020metric}
Musgrave K, Belongie S, Lim SN.
\newblock A metric learning reality check.
\newblock In: European Conference on Computer Vision. Springer; 2020. p.
  681--699.

\bibitem{snell2017prototypical}
Snell J, Swersky K, Zemel R.
\newblock Prototypical Networks for Few-shot Learning.
\newblock Advances in Neural Information Processing Systems.
  2017;30:4077--4087.

\bibitem{chen2019closer}
Chen WY, Liu YC, Kira Z, Wang YCF, Huang JB.
\newblock A closer look at few-shot classification.
\newblock arXiv preprint arXiv:190404232. 2019;.

\bibitem{miller1998wordnet}
Miller GA.
\newblock WordNet: An electronic lexical database.
\newblock MIT press; 1998.

\bibitem{godbole2002exploiting}
Godbole S.
\newblock Exploiting confusion matrices for automatic generation of topic
  hierarchies and scaling up multi-way classifiers.
\newblock Annual Progress Report, Indian Institute of Technology--Bombay,
  India. 2002;.

\bibitem{wang2017normface}
Wang F, Xiang X, Cheng J, Yuille AL.
\newblock Normface: L2 hypersphere embedding for face verification.
\newblock In: Proceedings of the 25th ACM international conference on
  Multimedia; 2017. p. 1041--1049.

\bibitem{kim2020proxy}
Kim S, Kim D, Cho M, Kwak S.
\newblock Proxy anchor loss for deep metric learning.
\newblock In: Proceedings of the IEEE/CVF Conference on Computer Vision and
  Pattern Recognition; 2020. p. 3238--3247.

\bibitem{wang2018cosface}
Wang H, Wang Y, Zhou Z, Ji X, Gong D, Zhou J, et~al.
\newblock Cosface: Large margin cosine loss for deep face recognition.
\newblock In: Proceedings of the IEEE conference on computer vision and pattern
  recognition; 2018. p. 5265--5274.

\bibitem{deng2019arcface}
Deng J, Guo J, Xue N, Zafeiriou S.
\newblock Arcface: Additive angular margin loss for deep face recognition.
\newblock In: Proceedings of the IEEE/CVF Conference on Computer Vision and
  Pattern Recognition; 2019. p. 4690--4699.

\bibitem{wan2021nbdt}
Wan A, Dunlap L, Ho D, Yin J, Lee S, Petryk S, et~al.
\newblock {\{}NBDT{\}}: Neural-Backed Decision Tree.
\newblock In: International Conference on Learning Representations;
  2021.Available from: \url{https://openreview.net/forum?id=mCLVeEpplNE}.

\bibitem{silla2011survey}
Silla CN, Freitas AA.
\newblock A survey of hierarchical classification across different application
  domains.
\newblock Data Mining and Knowledge Discovery. 2011;22(1):31--72.

\bibitem{weinberger2005distance}
Weinberger KQ, Blitzer J, Saul L.
\newblock Distance metric learning for large margin nearest neighbor
  classification.
\newblock Advances in neural information processing systems. 2005;18.

\bibitem{hadsell2006dimensionality}
Hadsell R, Chopra S, LeCun Y.
\newblock Dimensionality reduction by learning an invariant mapping.
\newblock In: 2006 IEEE Computer Society Conference on Computer Vision and
  Pattern Recognition (CVPR'06). vol.~2. IEEE; 2006. p. 1735--1742.

\bibitem{zhe2019improve}
Zhe X, Ou-Yang L, Yan H.
\newblock Improve L2-normalized Softmax with Exponential Moving Average.
\newblock In: 2019 International Joint Conference on Neural Networks (IJCNN).
  IEEE; 2019. p. 1--7.

\bibitem{zhang2019adacos}
Zhang X, Zhao R, Qiao Y, Wang X, Li H.
\newblock Adacos: Adaptively scaling cosine logits for effectively learning
  deep face representations.
\newblock In: Proceedings of the IEEE/CVF Conference on Computer Vision and
  Pattern Recognition; 2019. p. 10823--10832.

\bibitem{zhai2019classification}
Zhai A, Wu HY.
\newblock Classification is a strong baseline for deep metric learning.
\newblock British Machine Vision Conference (BMVC). 2019;.

\bibitem{liu2017sphereface}
Liu W, Wen Y, Yu Z, Li M, Raj B, Song L.
\newblock Sphereface: Deep hypersphere embedding for face recognition.
\newblock In: Proceedings of the IEEE conference on computer vision and pattern
  recognition; 2017. p. 212--220.

\bibitem{kim2022distance}
Kim H, Parviainen P, Malde K.
\newblock Distance-Ratio-Based Formulation for Metric Learning.
\newblock arXiv preprint arXiv:220108676. 2022;.

\bibitem{Krizhevsky09learningmultiple}
Krizhevsky A.
\newblock Learning multiple layers of features from tiny images; 2009.

\bibitem{van2015building}
Van~Horn G, Branson S, Farrell R, Haber S, Barry J, Ipeirotis P, et~al.
\newblock Building a bird recognition app and large scale dataset with citizen
  scientists: The fine print in fine-grained dataset collection.
\newblock In: Proceedings of the IEEE Conference on Computer Vision and Pattern
  Recognition; 2015. p. 595--604.

\bibitem{movshovitz2017no}
Movshovitz-Attias Y, Toshev A, Leung TK, Ioffe S, Singh S.
\newblock No fuss distance metric learning using proxies.
\newblock In: Proceedings of the IEEE International Conference on Computer
  Vision; 2017. p. 360--368.

\bibitem{szegedy2016rethinking}
Szegedy C, Vanhoucke V, Ioffe S, Shlens J, Wojna Z.
\newblock Rethinking the inception architecture for computer vision.
\newblock In: Proceedings of the IEEE conference on computer vision and pattern
  recognition; 2016. p. 2818--2826.

\bibitem{he2016deep}
He K, Zhang X, Ren S, Sun J.
\newblock Deep residual learning for image recognition.
\newblock In: Proceedings of the IEEE conference on computer vision and pattern
  recognition; 2016. p. 770--778.

\bibitem{kingma2014adam}
Kingma DP, Ba J.
\newblock Adam: A method for stochastic optimization.
\newblock arXiv preprint arXiv:14126980. 2014;.

\bibitem{frome2013devise}
Frome A, Corrado GS, Shlens J, Bengio S, Dean J, Ranzato M, et~al.
\newblock Devise: A deep visual-semantic embedding model.
\newblock Advances in neural information processing systems. 2013;26.

\bibitem{zhang2018unreasonable}
Zhang R, Isola P, Efros AA, Shechtman E, Wang O.
\newblock The unreasonable effectiveness of deep features as a perceptual
  metric.
\newblock In: Proceedings of the IEEE conference on computer vision and pattern
  recognition; 2018. p. 586--595.

\bibitem{johnson1984extensions}
Johnson W, Lindenstrauss J.
\newblock Extensions of Lipschitz mappings into a Hilbert space.
\newblock Contemporary Mathematics. 1984;26:189--206.

\end{thebibliography}

\setcounter{figure}{0}
\setcounter{table}{0}
\renewcommand{\thefigure}{S\arabic{figure}}
\renewcommand{\thetable}{S\arabic{table}}

\newpage
\section*{Supporting information}

\section*{S1 Appendix}\label{S1_appendix} %

\subsection*{{\bf Detailed %
explanation of the dynamic (adaptive) scaling factors in the NormFace and ProxyDR models.
}}\label{sec:ada_scale}

Zhang et al. \cite{zhang2019adacos} 
rewrote Eq. \ref{eq:normface} %
as follows:
\begin{align*}
    p(c|x)=\frac{\exp(s\cos{\theta_{c}})}{\exp(s\cos{\theta_{c}})+B_x},
\end{align*}
where \(c\) is the corresponding class of point \(x\) and \(B_x =\sum\limits_{y\neq c, y\in \mathcal{Y}}{\exp(s\cos{\theta_y})}\). %
They found that \(\theta_y\) was close to \(\frac{\pi}{2}\) during the training process for \(y\neq c\), i.e., for different classes. %
Thus, \(B_x\approx \sum\limits_{y\neq c, y\in \mathcal{Y}}{\exp(s\cos{\frac{\pi}{2}})}=\sum\limits_{y\neq c, y\in \mathcal{Y}}{\exp(0)}=\left| \mathcal{Y}\right|-1\). \\
In Eq. \ref{eq:adacos}, %
\(\frac{\partial^2 p(c|x) (\theta_{c})}{\partial \theta_{c}^2}\)
can be written as:
\begin{align*} %
\frac{\partial^2 p(c|x) (\theta_{c})}{\partial \theta_{c}^2}=\frac{-s B_{x}\exp({s\cos{\theta_{c}}}) \psi_{NormFace}(s,\theta_{c}) }{\left( \exp({s\cos{\theta_{c}}}) +B_{x}\right)^3},
\end{align*}
where \(\psi_{NormFace}(s,\theta_{c})=\cos{\theta_{c}}\left( \exp({s\cos{\theta_{c}}}) +B_{x}\right)+s\sin^2{\theta_{c}}\left( \exp({s\cos{\theta_{c}}}) -B_{x}\right) \). %
Moreover, they used \(s=\frac{\log{B_x}}{\cos{\theta_{c}}}\) to approximate %
the solution for Eq. \ref{eq:adacos}. %
For the static version, they used \(\left| \mathcal{Y}\right|-1\) to estimate \(B_x\) and \(\frac{\pi}{4}\) to estimate \(\theta_{c}\). For the dynamic version, they used \(B_{x;avg}\) to estimate \(B_x\) and \(\theta_{c;med}\) to estimate \(\theta_{c}\), where \(B_{x;avg}\) is the average of \(B_x\) in a mini-batch and \(\theta_{c;med}\) is the median of the \(\theta_{c}\) values in a mini-batch. They clipped the \(\theta_{c;med}\) value to be in the range \(\left [ 0,\frac{\pi}{4} \right ]\).
\\
Instead of using \(s=\frac{\log{B_x}}{\cos{\theta_{c}}}\), in our implementation, we use the Adam optimizer \cite{kingma2014adam} %
to update a scale factor \(s\) that minimizes \(\psi_{NormFace}^2 (s,\theta_{c})\), i.e., \(\psi_{NormFace}(s,\theta_{c})\approx 0\). For \(\psi_{NormFace}(s,\theta_{c})\), we use \(\left| \mathcal{Y}\right|-1\) to estimate \(B_x\) and \(\frac{\pi}{4}\) to estimate \(\theta_{c}\) to initialize the value of \(s\). During model training, we use \(B_{x;avg}\) to estimate \(B_x\) and \(\theta_{c;med}\) to estimate \(\theta_{c}\).\\
We can also apply the dynamic scaling factor to the ProxyDR model. First, we define \(d_{x,y}:=\left\| \Tilde{f}(x)-\Tilde{W}_y\right\|\). We rewrite Eq. \ref{eq:DR_form} %
as:
\begin{align*}
    p(c|x)=\frac{d_{x,c}^{-s}}{d_{x,c}^{-s}+B_x},
\end{align*}
where \(B_x =\sum\limits_{y\neq c, y\in \mathcal{Y}}{ d_{x,y}^{-s}}\). %
Assuming \(\theta_y\approx\frac{\pi}{2}\) for \(y\neq c\), we obtain \(B_x\approx \sum\limits_{y\neq c, y\in \mathcal{Y}}{\left ( \frac{\pi}{2} \right )^{-s}}=\left ( \left| \mathcal{Y}\right|-1\right ) \left ( \frac{\pi}{2} \right )^{-s}\). \\%\(B_x\approx \sum\limits_{y\neq c_{x}, y\in \mathcal{Y}}{\left ( \frac{\pi}{2} \right )^{-s}}=\left ( \left| \mathcal{Y}\right|-1\right ) \left ( \frac{\pi}{2} \right )^{-s}\). \\
The expression \(\frac{\partial^2 p(c|x) (\theta_{c})}{\partial \theta_{c}^2}\)
can be written as:
\begin{align*}
    \frac{\partial^2 p(c|x) (\theta_{c})}{\partial \theta_{c}^2}=\frac{B_{x} s \theta_{c}^{(s-2)} \psi_{ProxyDR}(s,\theta_{c})}{\left (B_{x} \theta_{c}^s +1\right )^3}
\end{align*}
where \(\psi_{ProxyDR}(s,\theta_{c})=B_{x} (s+1)\theta_{c}^s-s+1\). %
We then use the Adam optimizer to update a scale factor \(s\) that minimizes \(\psi_{ProxyDR}^2(s,\theta_{c})\).

\nolinenumbers

\newpage

\section*{S2 Appendix}\label{S2_appendix}%
\subsection*{{\bf Dataset details. %
}}\label{sec:detailed_datasets}

All plankton images were obtained using FlowCam (Yokogawa Fluid Imaging Technologies). FlowCam is a flow imaging microscope that captures particles flowing through glass flowcells with well-defined volumes. The three plankton datasets were obtained according to different types of samples (live and Lugol fixed whole seawater or \(180 \,\mu m\) WP2 plankton net samples) using three different FlowCams (FlowCam 8400, FlowCam VS, and FlowCam Macro) with various magnifications. Thus, the datasets include particles ranging from \(5\) to \(2000\,\mu m\) in size, thus representing nano-, micro-, and mesozooplankton. The plankton samples were obtained from three coastal monitoring stations (Institute of Marine Research) along the Norwegian coast, including Holmfjord in the north, Austevoll in the west and Torungen in the south. In addition, for the nano- and microplankton, seawater samples were obtained from a tidal zone at a depth of 1 meter at the research station at Flødevigen in southern Norway, which is approximately 2 nautical miles from the southern monitoring station at Torungen. The sampling period for the three datasets covered all seasons over a period of approximately 2.5 years. %
\paragraph{Small microplankton (MicroS)} %
This dataset contains images of fixed and live seawater samples acquired at a depth of \(5\,m\) at the three monitoring stations and a depth of \(1\,m\) in the tidal zone (see above). The seawater samples were carefully filtered through a \(80\,\mu m\) mesh to ensure that \(100 \,\mu m\) flowcell was not clogged and imaged using a \(10\times\) objective. This FlowCam configuration results in a total magnification of \(100\times\) and images particles ranging from %
\(5\) to \(50 \,\mu m\).
Before resizing, one pixel in an image represented \(0.7330 \,\mu m\).
\paragraph{Large microplankton (MicroL)} %
This dataset contains images of fixed and live seawater samples acquired at a depth of \(5\,m\) at the three monitoring stations and a depth of \(1\,m\) %
in the tidal zone (see above). The seawater samples were not filtered and were imaged using a \(2\times\) objective, targeting \(35\) to \(500 \,\mu m\) particles. Before resizing, one pixel in an image represented \(2.9730 \,\mu m\). Due to instrument repair and adjustments to improve image quality, the camera settings were modified during the 3 years of imaging to acquire this dataset. %
Therefore, the image appearance and quality are slightly variable.
\paragraph{Mesozooplankton (MesoZ)} %
This dataset contains images of mesozooplankton samples acquired at the three coastal monitoring stations (see above) and a transect in the Norwegian Sea (Svinøysnittet). The samples were obtained using an IMR (Institute of Marine Research) standard plankton net (WP2) or a multinet mammoth (both \(180 \,\mu m\) mesh) and fixed with \(4 \%\) %
formaldehyde. The images were acquired by two FlowCam instruments (one in Bergen and one in Flødevigen), and the image appearance differs slightly between the two instruments. %
The FlowCam macro was equipped with a \(0.5\times\) objective, resulting in a total magnification of 12.5 and imaging organisms ranging from \(180\) to \(2000\,\mu m\). Before resizing, one pixel in an image represented \(9.05 \,\mu m\). %
All images in the mesozooplankton dataset are in grayscale.

\paragraph{NABirds} %
Data provided by the Cornell Lab of Ornithology, with thanks to photographers and contributors of crowdsourced data at \url{AllAboutBirds.org/Labs}. This material is based upon work supported by the National Science Foundation under Grant No. 1010818.

\begin{figure}[H]
    \centering
    \includegraphics[width=0.5\linewidth,height=1.45\linewidth]{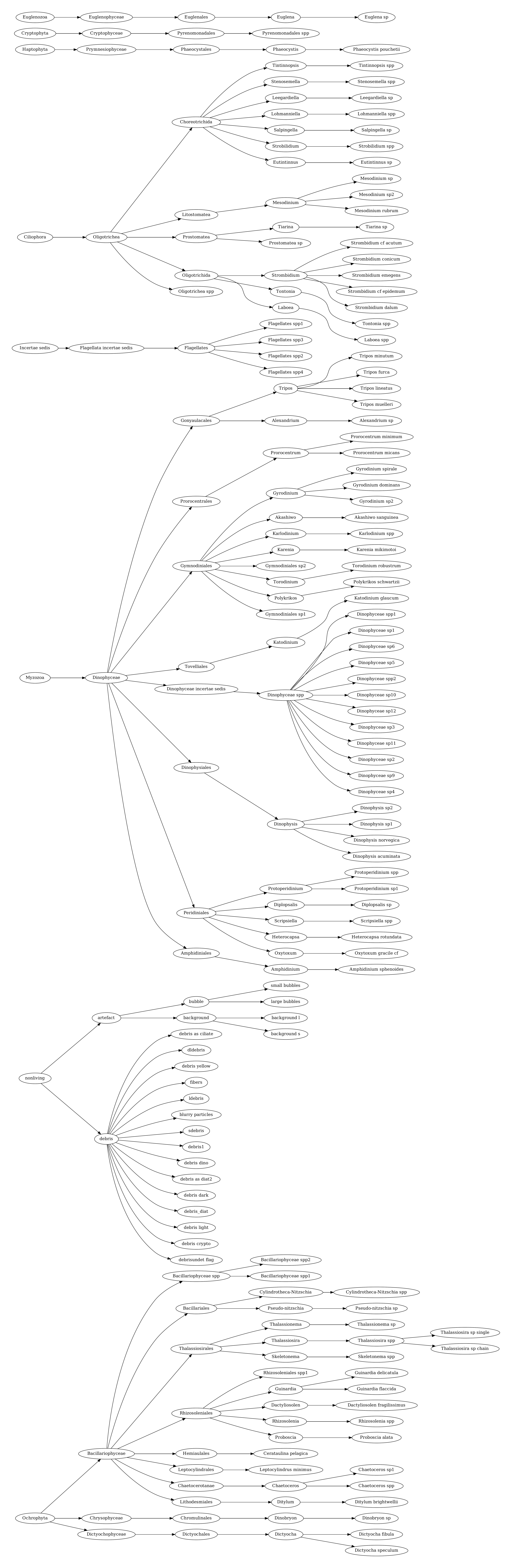}
\caption{{\bf The hierarchical structure of the MicroS %
dataset used in our experiment.} Best viewed by zooming in. }\label{fig:hierarchy_2pri2f}
\end{figure}

\begin{figure}[H]
    \centering
    \includegraphics[width=0.88\linewidth,height=1.45\linewidth]{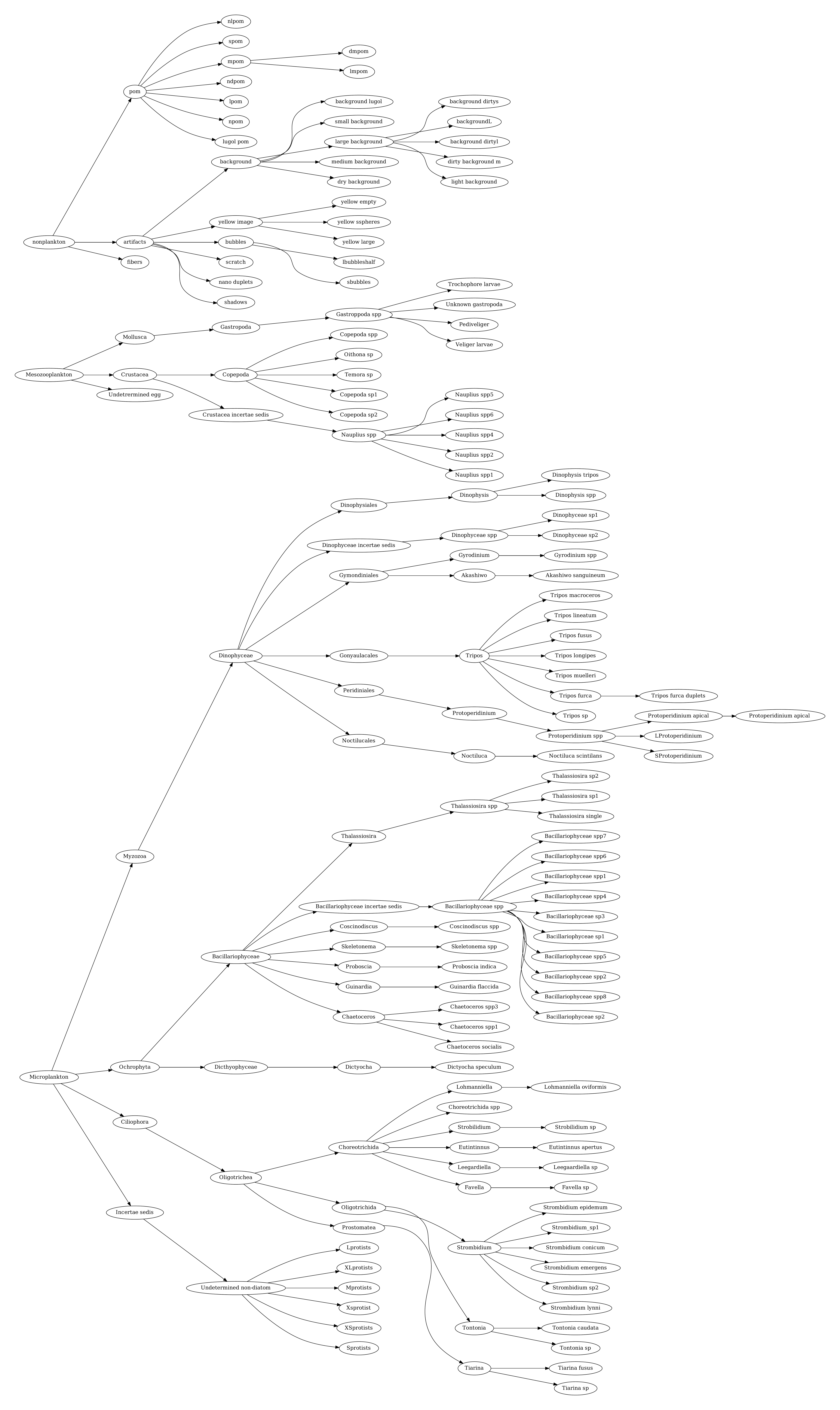}
\caption{{\bf The hierarchical structure of the MicroL %
dataset used in our experiment.} Best viewed by zooming in. }\label{fig:hierarchy_4pri}
\end{figure}

\begin{figure}[H]
    \centering
    \includegraphics[width=1.0\linewidth,height=1.2375\linewidth]{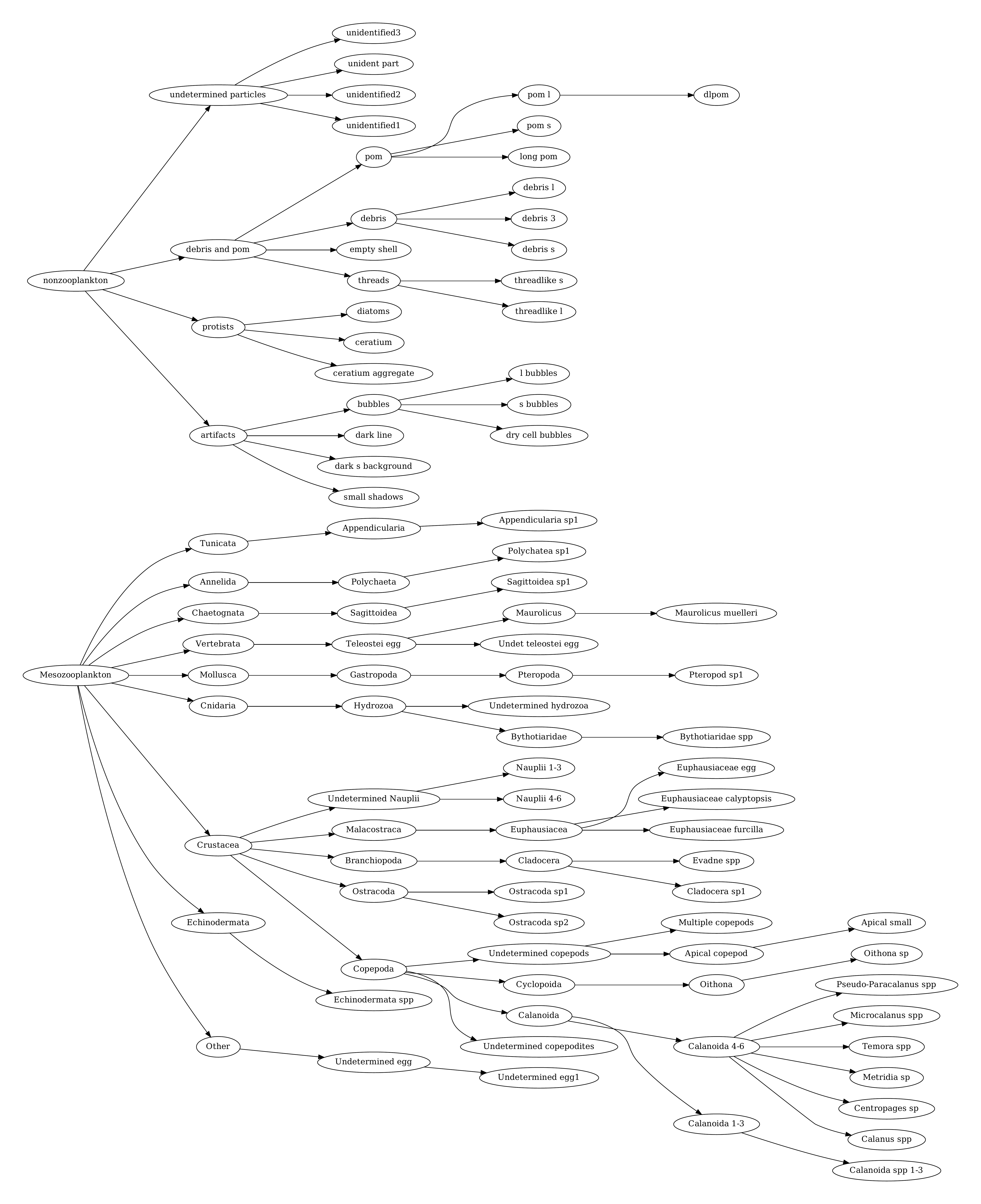}
\caption{{\bf The hierarchical structure of the MesoZ %
dataset used in our experiment.} Best viewed by zooming in. }\label{fig:hierarchy_5pri2}
\end{figure}

\paragraph{Hierarchical structures of the datasets}
Figs. \ref{fig:hierarchy_2pri2f}, \ref{fig:hierarchy_4pri}, and \ref{fig:hierarchy_5pri2} and Table \ref{tab:hierarchy_cifar} show the hierarchical structures of the datasets. As the NABirds dataset contains too many (555) classes to visualize, we do not show the hierarchical structures of the NABirds dataset \cite{van2015building}. %
We used the hierarchy provided by the Cornell Lab of Ornithology.

\newcolumntype{M}[1]{>{\arraybackslash}m{#1}}
\newcolumntype{\Mc}[1]{>{\centering\arraybackslash}m{#1}}

\begin{table}[H]%
    \caption{{\bf The hierarchical structure of the CIFAR100 dataset used in our experiment.}}%
    \label{tab:hierarchy_cifar}
    \centering
    \begin{tabular}{\Mc{2.25cm}|\Mc{2.5cm}|\Mc{2.5cm}|\Mc{4.5cm}}
    \toprule
    \small Level 0 & \small Level 1 & \small Level 2 & \small \makecell{Level 3\\ (class level)} \\
    \midrule
    \multirow{10}{*}{\small Animals}	& \multirow{2}{*}{\small Invertebrates}	& \scriptsize insects	& \scriptsize bee, beetle, butterfly, caterpillar, cockroach\\ \cline{3-4}
    {}	& {} & \scriptsize non-insect invertebrates & \scriptsize crab, lobster, snail, spider, worm\\\cline{2-4}
{}	& \multirow{6}{*}{\small Mammals} & \scriptsize aquatic mammals	& \scriptsize beaver, dolphin, otter, seal, whale\\\cline{3-4}
{}	& {}& \scriptsize large carnivores	& \scriptsize bear, leopard, lion, tiger, wolf\\\cline{3-4}
{}	& {} & \scriptsize large omnivores and herbivores & \scriptsize camel, cattle, chimpanzee, elephant, kangaroo\\\cline{3-4}
{}	& {} & \scriptsize medium-sized mammals & \scriptsize fox, porcupine, possum, raccoon, skunk\\\cline{3-4}
{}	& {} & \scriptsize people & \scriptsize baby, boy, girl, man, woman\\ \cline{3-4}
{}	& {} & \scriptsize small mammals	& \scriptsize hamster, mouse, rabbit, shrew, squirrel\\\cline{2-4}
{}	& \multirow{2}{*}{ \makecell{\small Non-mammal\\ vertebrates}} & \scriptsize fish & \scriptsize aquarium fish, flatfish, ray, shark, trout\\ \cline{3-4}
{}	& {} & \scriptsize reptiles & \scriptsize crocodile, dinosaur, lizard, snake, turtle\\\cline{1-4}

\multirow{6}{*}{\small Artificial} & \multirow{3}{*}{\makecell{Artificial \\(indoor)}}	& \scriptsize food containers & \scriptsize bottles, bowls, cans, cups, plates\\ \cline{3-4}
{} & {} & \scriptsize household electrical devices & \scriptsize clock, computer keyboard, lamp, telephone, television\\ \cline{3-4}
{} & {} & \scriptsize household furniture & \scriptsize bed, chair, couch, table, wardrobe\\ \cline{2-4}
{} & \multirow{3}{*}{\makecell{Artificial \\(outdoor)}} & \scriptsize large man-made outdoor things & \scriptsize bridge, castle, house, road, skyscraper\\ \cline{3-4}
{} & {} & \scriptsize vehicles 1 & \scriptsize bicycle, bus, motorcycle, pickup truck, train\\ \cline{3-4}
{} & {} & \scriptsize vehicles 2 & \scriptsize lawn-mower, rocket, streetcar, tank, tractor\\ \cline{1-4}
\multirow{3}{*}{\makecell{Nature\\(non-animal)}} & \makecell{Nature\\(non-specific\\organism)} & \scriptsize large natural outdoor scenes & \scriptsize cloud, forest, mountain, plain, sea\\ \cline{2-4}
{} & \multirow{3}{*}{Plants} & \scriptsize flowers & \scriptsize orchids, poppies, roses, sunflowers, tulips\\ \cline{3-4}
{} & {} & \scriptsize fruit and vegetables & \scriptsize apples, mushrooms, oranges, pears, sweet peppers\\ \cline{3-4}
{} & {} & \scriptsize trees & \scriptsize maple, oak, palm, pine, willow\\ %
    \bottomrule
    \end{tabular}
\end{table}

\newpage
\section*{S3 Appendix}\label{S3_appendix}%

\begin{figure}[H]
    \centering
\vspace{.05in}
\includegraphics[width=1.0\linewidth,height=0.35\linewidth]{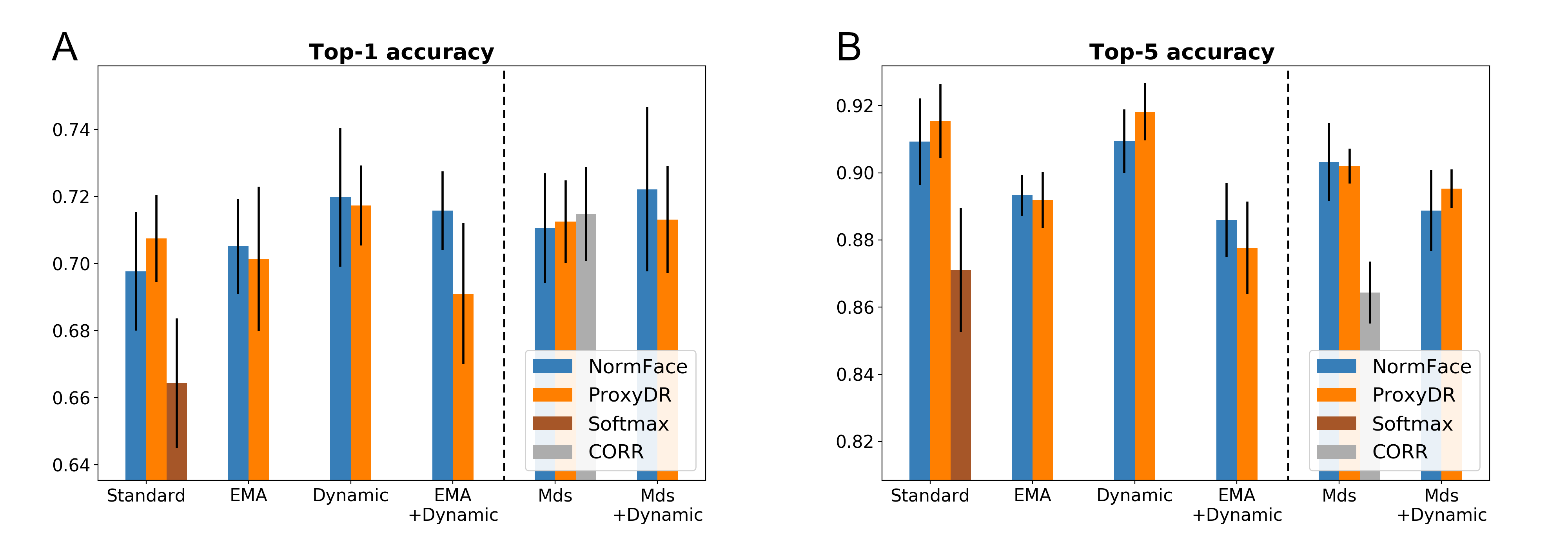}
\caption{{\bf Top-\(k\) accuracy results (A: \(k=1\), B: \(k=5\)) on the MicroS dataset.} %
}
    \label{fig:2pri2f_acc}
\end{figure}

\begin{figure}[H]
    \centering
\vspace{.05in}
\includegraphics[width=1.0\linewidth,height=0.35\linewidth]{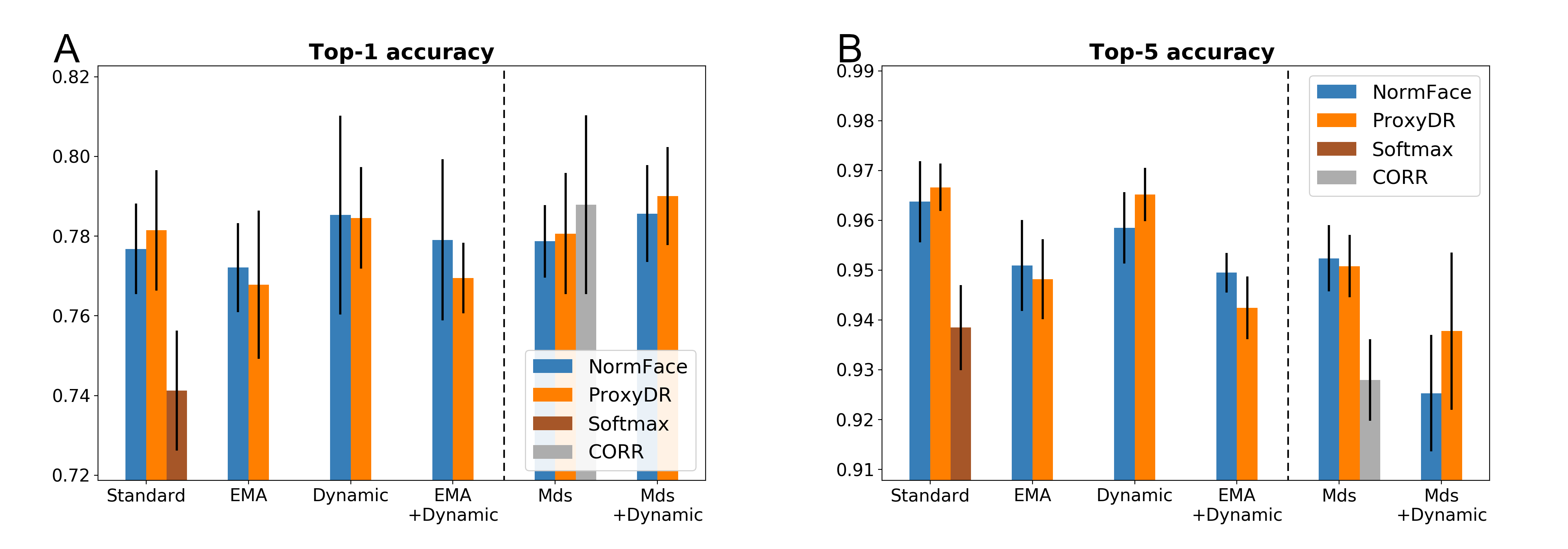}
\caption{{\bf Top-\(k\) accuracy results (A: \(k=1\), B: \(k=5\)) on the MicroL dataset.} %
}
    \label{fig:4pri_acc}
\end{figure}

\begin{figure}[H]
    \centering
\vspace{.05in}
\includegraphics[width=1.0\linewidth,height=0.35\linewidth]{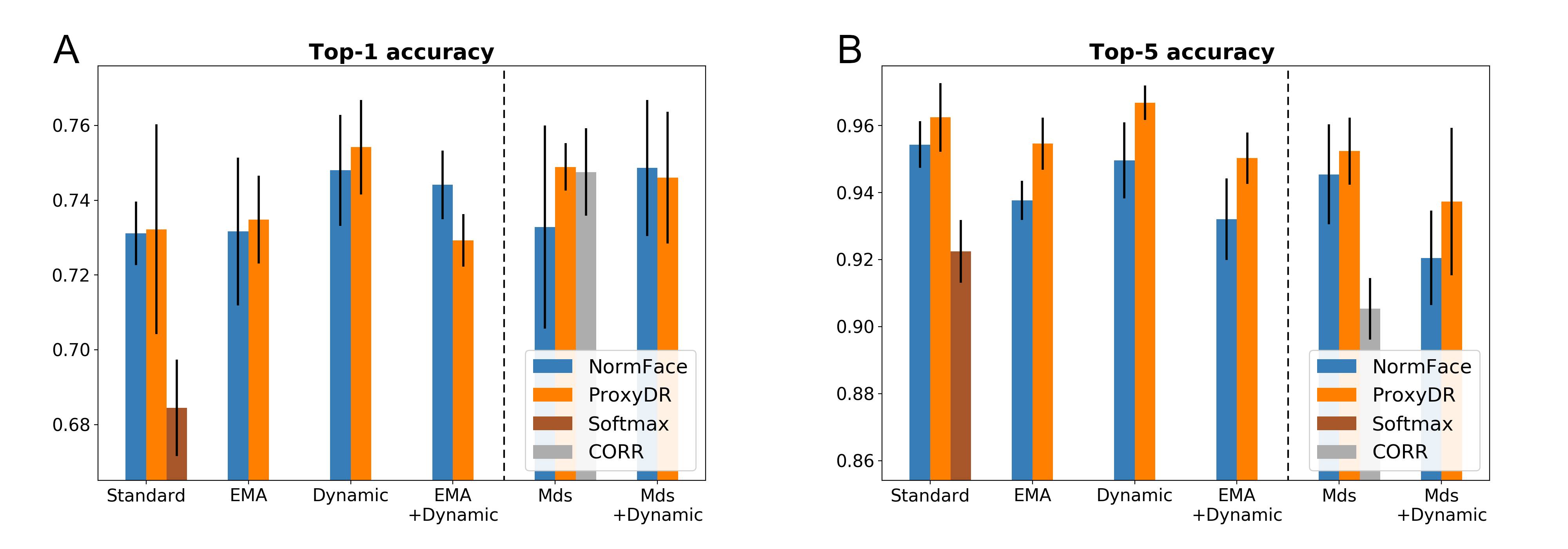}
\caption{{\bf Top-\(k\) accuracy results (A: \(k=1\), B: \(k=5\)) on the MesoZ dataset.} %
}
    \label{fig:5pri2_acc}
\end{figure}

\begin{figure}[H]
    \centering
\vspace{.05in}
\includegraphics[width=1.0\linewidth,height=0.35\linewidth]{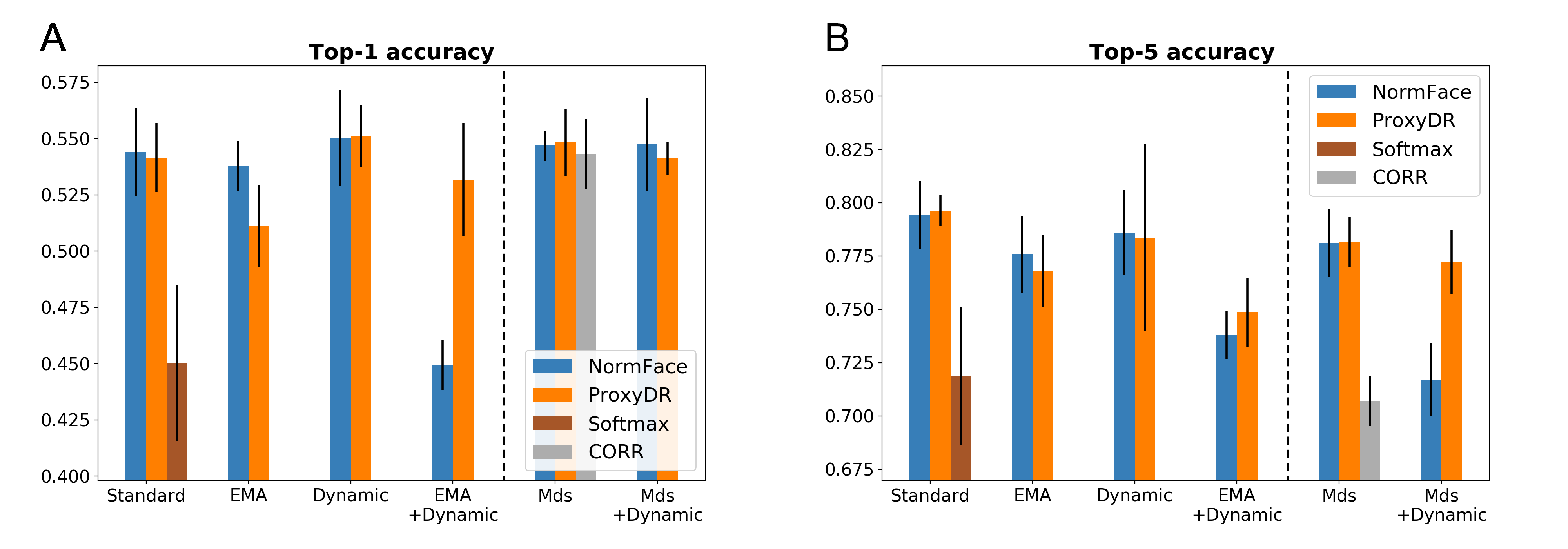}
\caption{{\bf Top-\(k\) accuracy results (A: \(k=1\), B: \(k=5\)) on the CIFAR100 dataset.} }
    \label{fig:cifar_acc}
\end{figure}

\begin{figure}[H]
    \centering
\vspace{.05in}
\includegraphics[width=1.0\linewidth,height=0.35\linewidth]{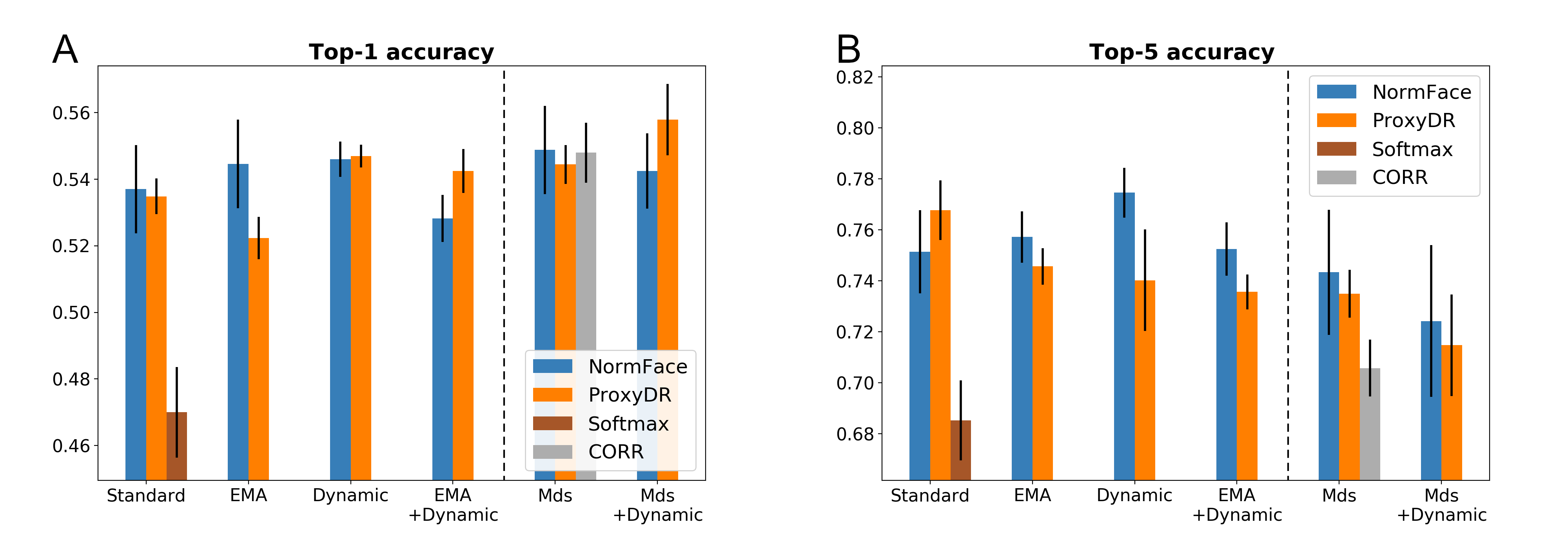}
\caption{{\bf Top-\(k\) accuracy results (A: \(k=1\), B: \(k=5\)) on the NABirds dataset.} }
    \label{fig:nabird_acc}
\end{figure}

{\bf Results for all five datasets.\\}
Figs. \ref{fig:2pri2f_acc}, \ref{fig:4pri_acc}, \ref{fig:5pri2_acc}, \ref{fig:cifar_acc}, and \ref{fig:nabird_acc} show the top-\(k\) accuracies on the different datasets.

\begin{figure}[H]
    \centering
\vspace{.05in}
\includegraphics[width=1.0\linewidth,height=0.7\linewidth]{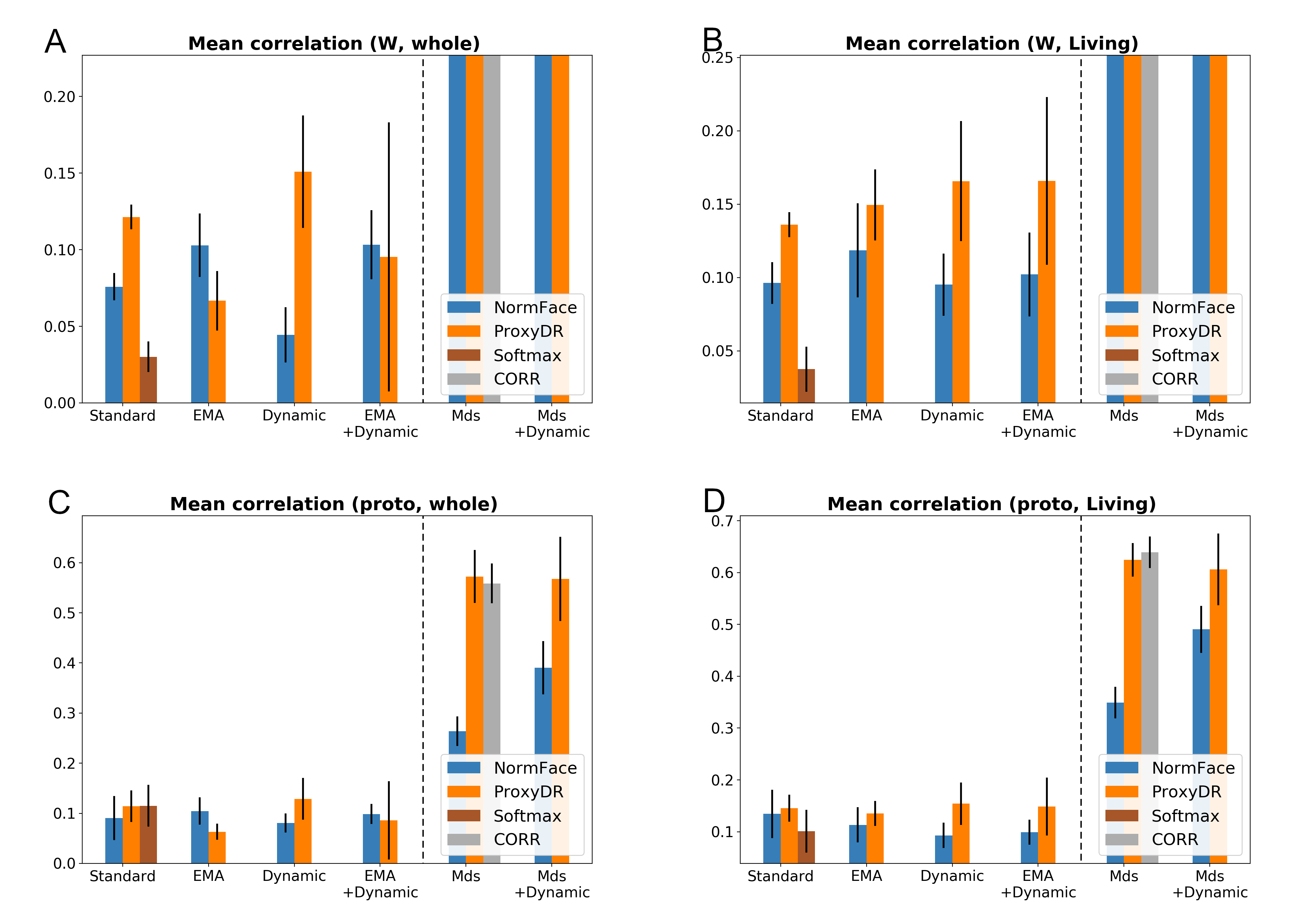}
\caption{{\bf Correlation measures on the MicroS dataset.} %
(Top) Values using proxies (A: whole classes, B: living classes). (Bottom) Values using prototypes (C: whole classes, D: living classes). %
`Living' indicates that only biological classes were used (no nonliving classes). 
The mean correlation values based on proxies with the MDS option were \(0.9306\) (whole) and \(0.9011\) (living). %
}
    \label{fig:2pri2f_corr}
\end{figure}

\begin{figure}[H]
    \centering
\vspace{.05in}
\includegraphics[width=1.0\linewidth,height=0.7\linewidth]{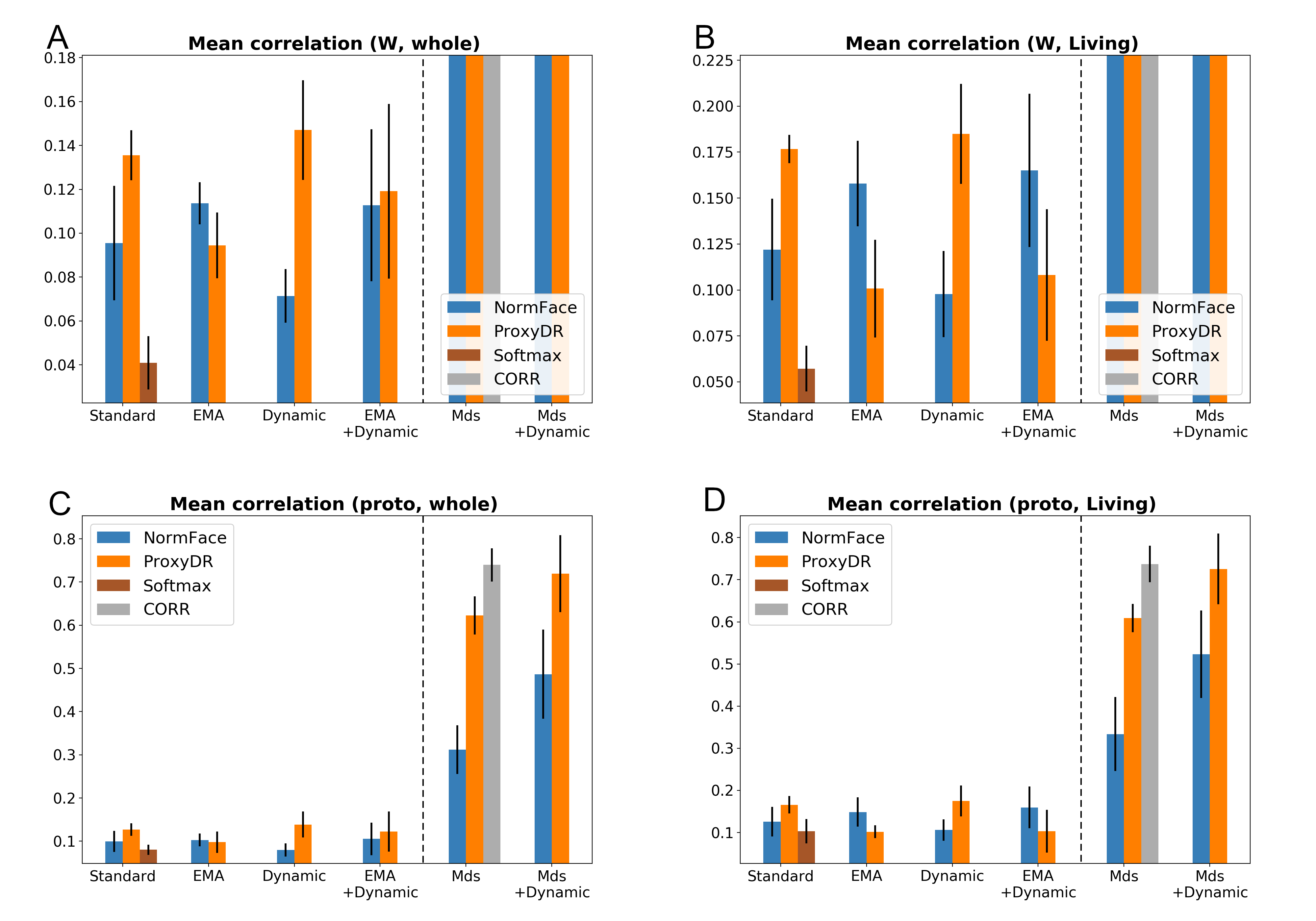}
\caption{{\bf Correlation measures on the MicroL dataset.} %
(Top) Values using proxies (A: whole classes, B: living classes). (Bottom) Values using prototypes (C: whole classes, D: living classes). The mean correlation values based on proxies with %
the MDS option were \(0.9543\) (whole) and \(0.9426\) (living).}
    \label{fig:4pri_corr}
\end{figure}

\begin{figure}[H]
    \centering
\vspace{.05in}
\includegraphics[width=1.0\linewidth,height=0.7\linewidth]{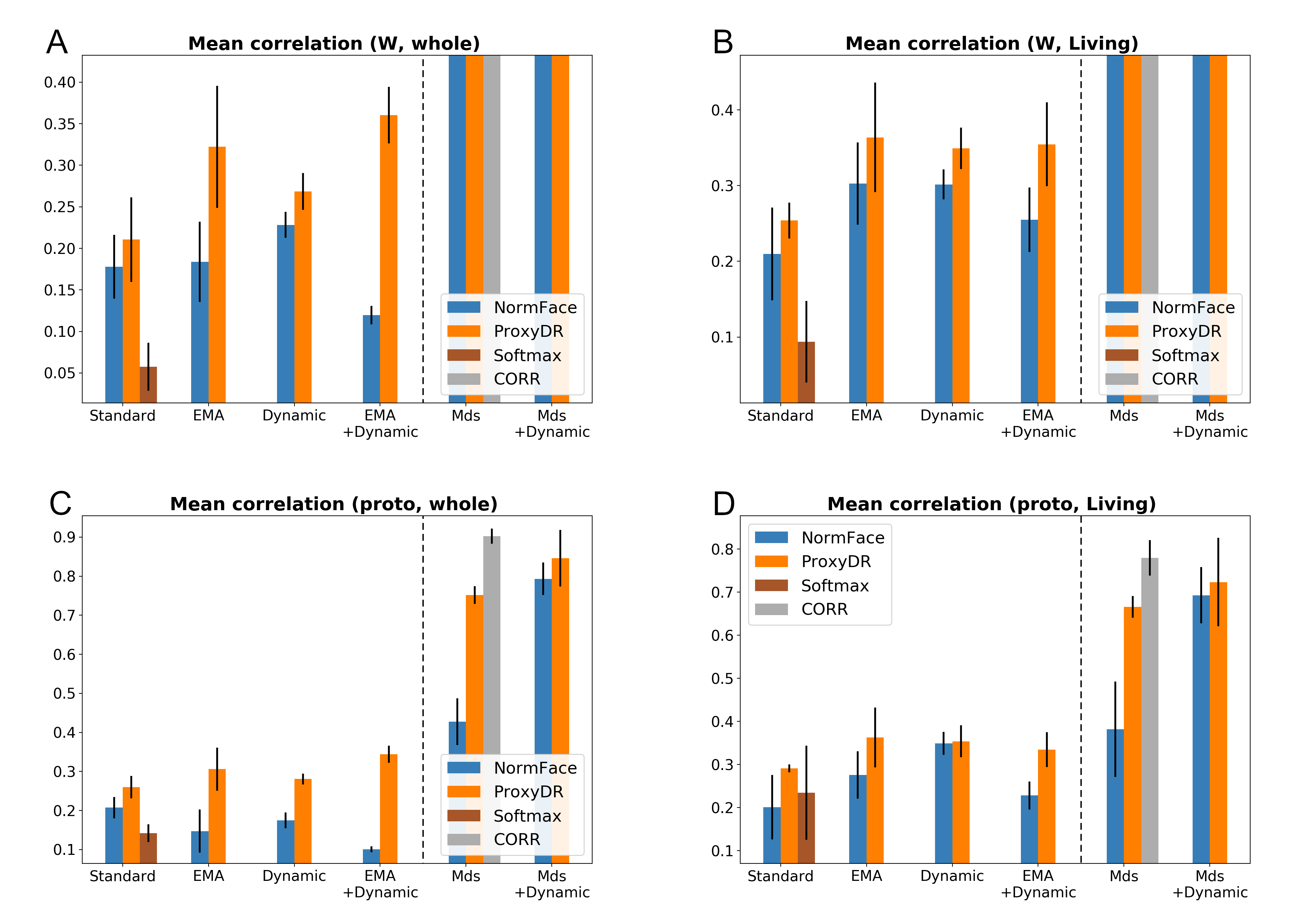}
\caption{{\bf Correlation measures on the MesoZ dataset.} %
(Top) Values using proxies (A: whole classes, B: living classes). (Bottom) Values using prototypes (C: whole classes, D: living classes). The mean correlation values based on proxies with the MDS option %
were \(0.9783\) (whole) and \(0.9602\) (living). }
    \label{fig:5pri2_corr}
\end{figure}

\begin{figure}[H]
    \centering
\vspace{.05in}
\includegraphics[width=1.0\linewidth,height=0.35\linewidth]{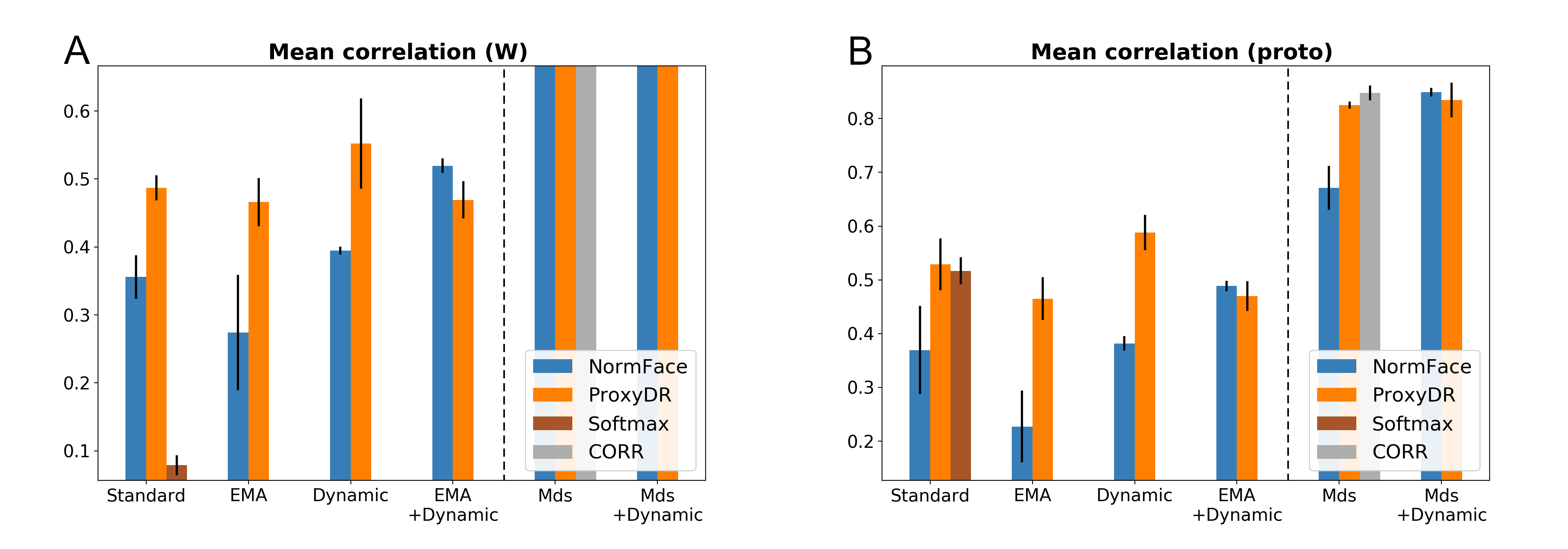}
\caption{{\bf Correlation measures on the CIFAR100 dataset.} (A) Values using proxies. (B) Values using prototypes. The mean correlation value based on proxies with the MDS option %
was \(0.8580\). }
    \label{fig:cifar_corr}
\end{figure}

\begin{figure}[H]
    \centering
\vspace{.05in}
\includegraphics[width=1.0\linewidth,height=0.35\linewidth]{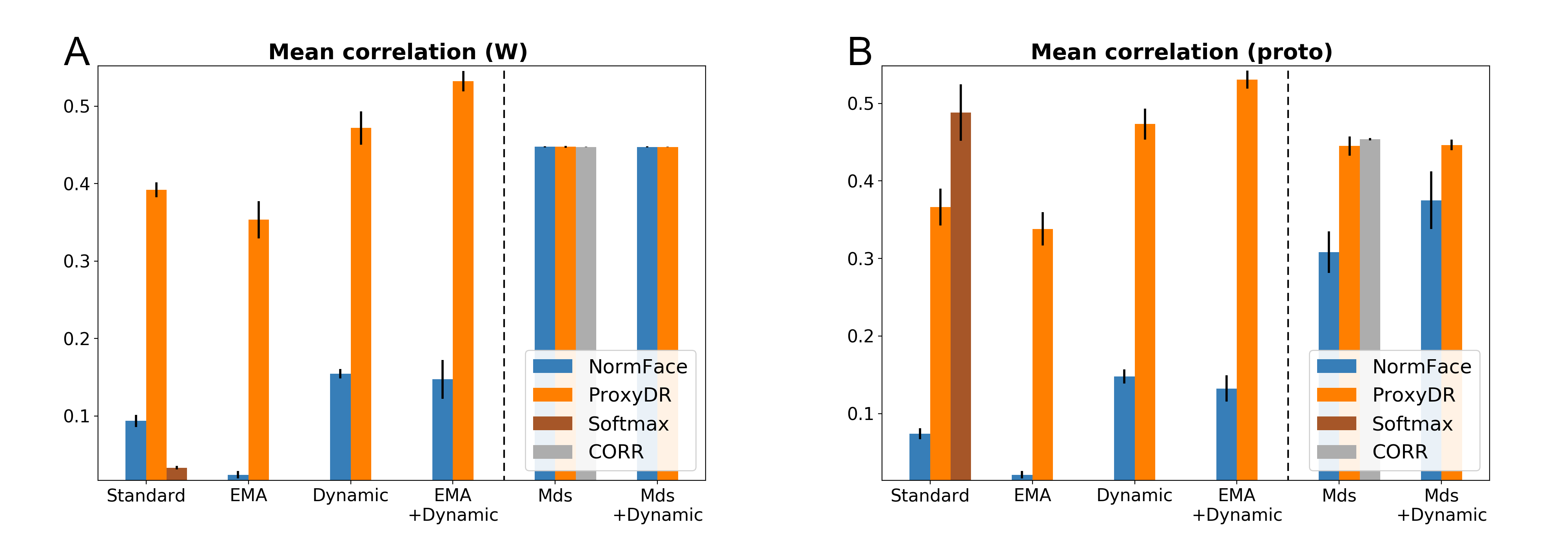}
\caption{{\bf Correlation measures on the NABirds dataset.} (A) Values using proxies. (B) Values using prototypes. The mean correlation value based on proxies with the MDS option %
was \(0.4476\) (this value is small as the dataset contains 555 classes and the embedding dimension is 128.). }
    \label{fig:nabird_corr}
\end{figure}

Figs. \ref{fig:2pri2f_corr}, \ref{fig:4pri_corr}, \ref{fig:5pri2_corr}, \ref{fig:cifar_corr}, and \ref{fig:nabird_corr} show the mean correlation values on the different datasets.

\begin{figure}[H]
    \centering
\vspace{.05in}
\includegraphics[width=1.0\linewidth,height=1.0\linewidth]{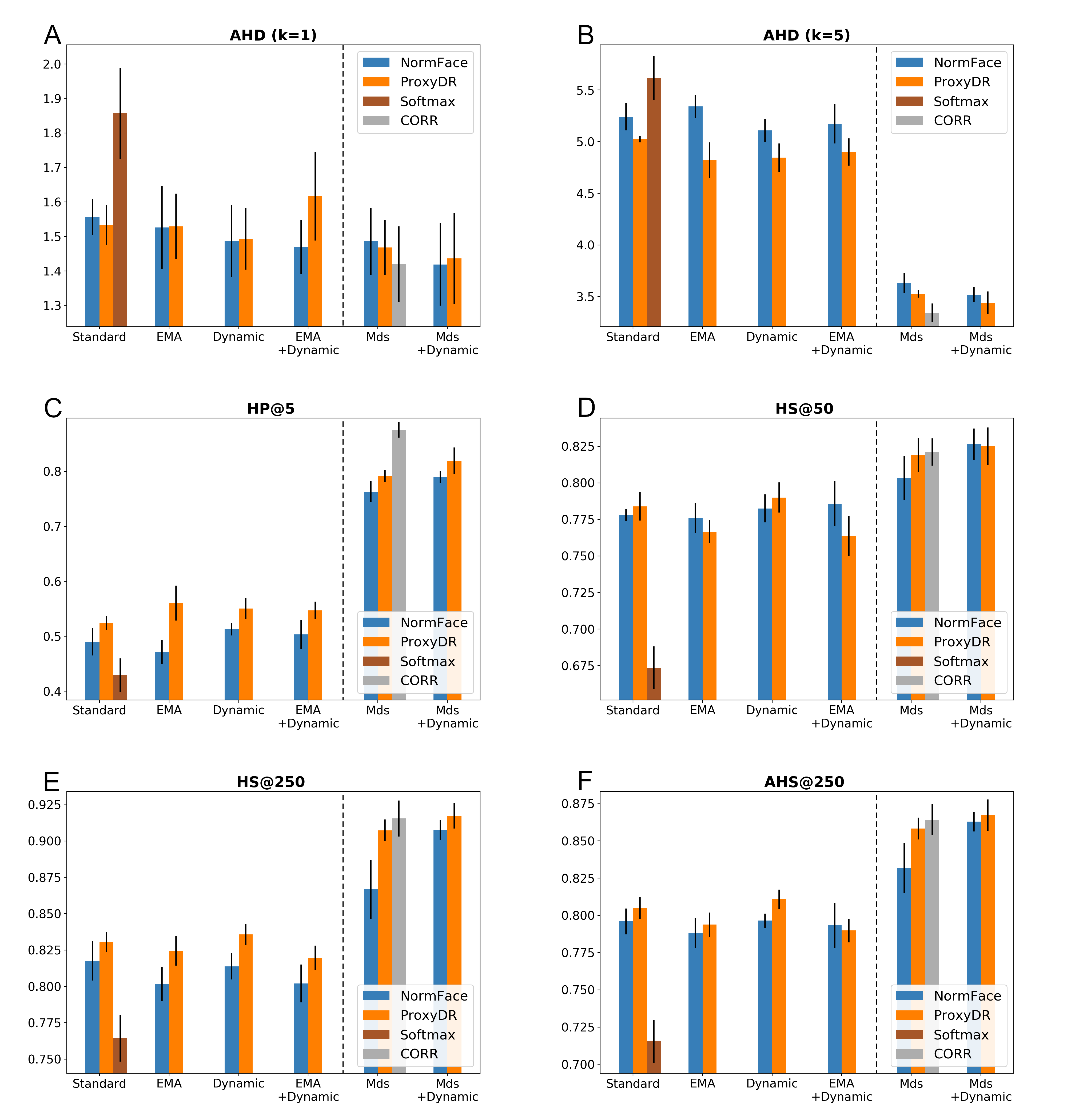}
\caption{{\bf Hierarchical performance measures on the MicroS dataset.} %
The symbol \(\downarrow\) denotes that lower values indicate better performance. %
The symbol \(\uparrow\) denotes that higher values indicate better performance. %
(A) AHD (k=1): \(\downarrow\). (B) AHD (k=5): \(\downarrow\). (C) HP@5: \(\uparrow\). (D) HS@50: \(\uparrow\). (E) HS@250: \(\uparrow\). (F) AHS@250: \(\uparrow\).
}
    \label{fig:2pri2f_H_measures}
\end{figure}

\begin{figure}[H]
    \centering
\vspace{.05in}
\includegraphics[width=1.0\linewidth,height=1.0\linewidth]{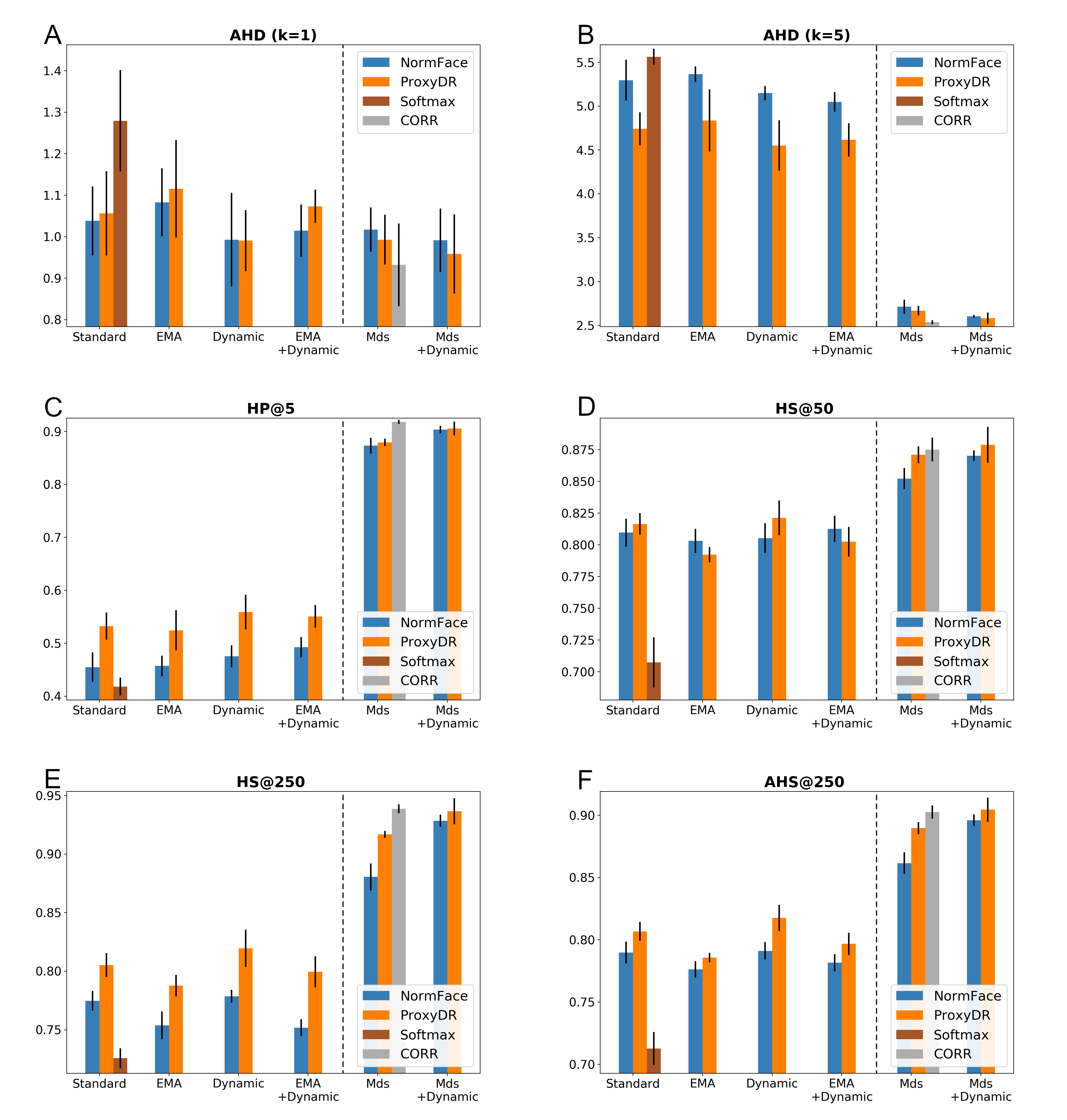}
\caption{{\bf Hierarchical performance measures on the MicroL dataset.} %
The symbol \(\downarrow\) denotes that lower values indicate better performance. %
The symbol \(\uparrow\) denotes that higher values indicate better performance. %
(A) AHD (k=1): \(\downarrow\). (B) AHD (k=5): \(\downarrow\). (C) HP@5: \(\uparrow\). (D) HS@50: \(\uparrow\). (E) HS@250: \(\uparrow\). (F) AHS@250: \(\uparrow\). %
}
    \label{fig:4pri_H_measures}
\end{figure}

\begin{figure}[H]
    \centering
\vspace{.05in}
\includegraphics[width=1.0\linewidth,height=1.0\linewidth]{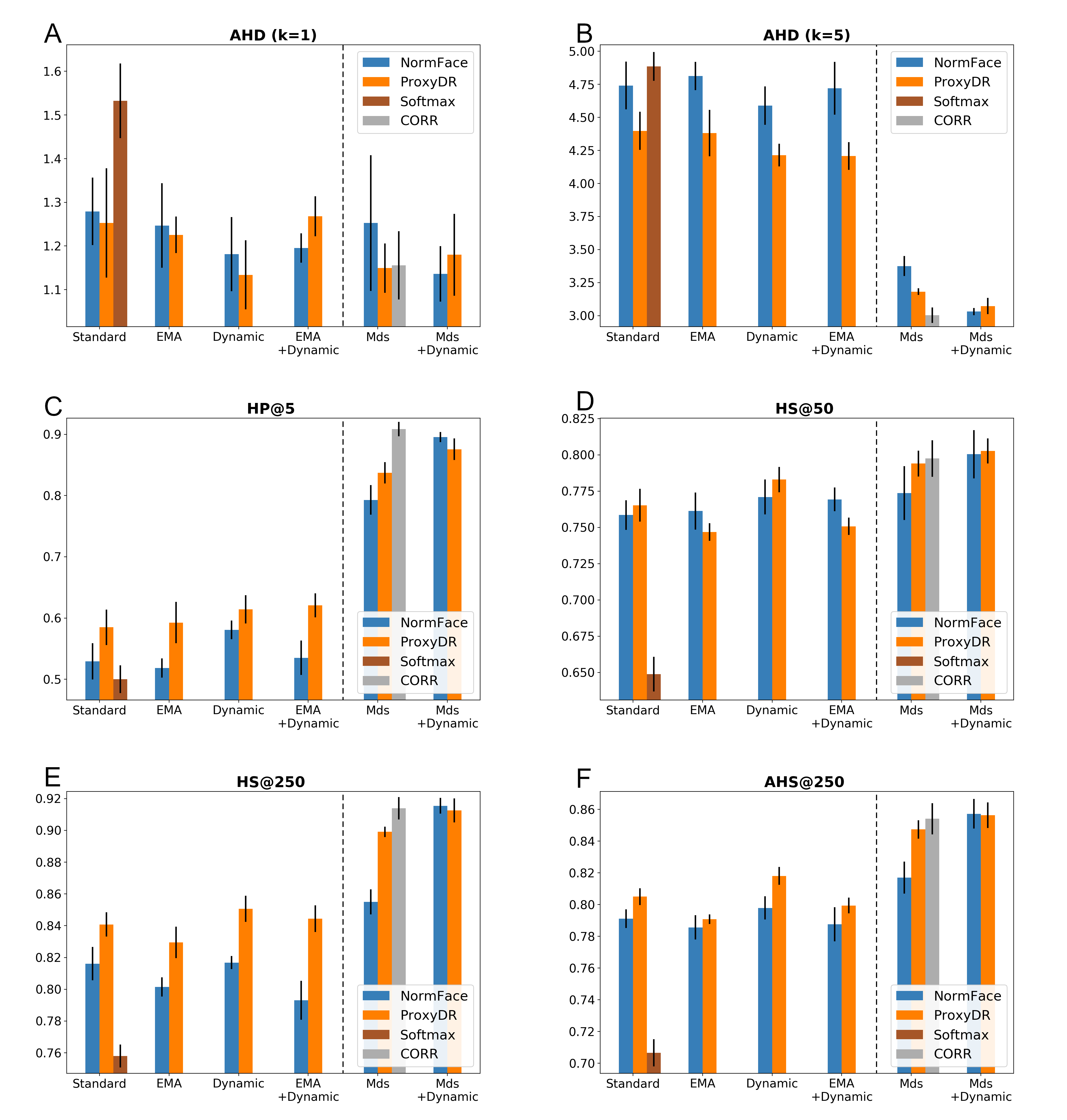}
\caption{{\bf Hierarchical performance measures on the MesoZ dataset.} %
The symbol \(\downarrow\) denotes that lower values indicate better performance. %
The symbol \(\uparrow\) denotes that higher values indicate better performance. %
(A) AHD (k=1): \(\downarrow\). (B) AHD (k=5): \(\downarrow\). (C) HP@5: \(\uparrow\). (D) HS@50: \(\uparrow\). (E) HS@250: \(\uparrow\). (F) AHS@250: \(\uparrow\). %
}
    \label{fig:5pri2_H_measures}
\end{figure}

\begin{figure}[H]
    \centering
\vspace{.05in}
\includegraphics[width=1.0\linewidth,height=1.0\linewidth]{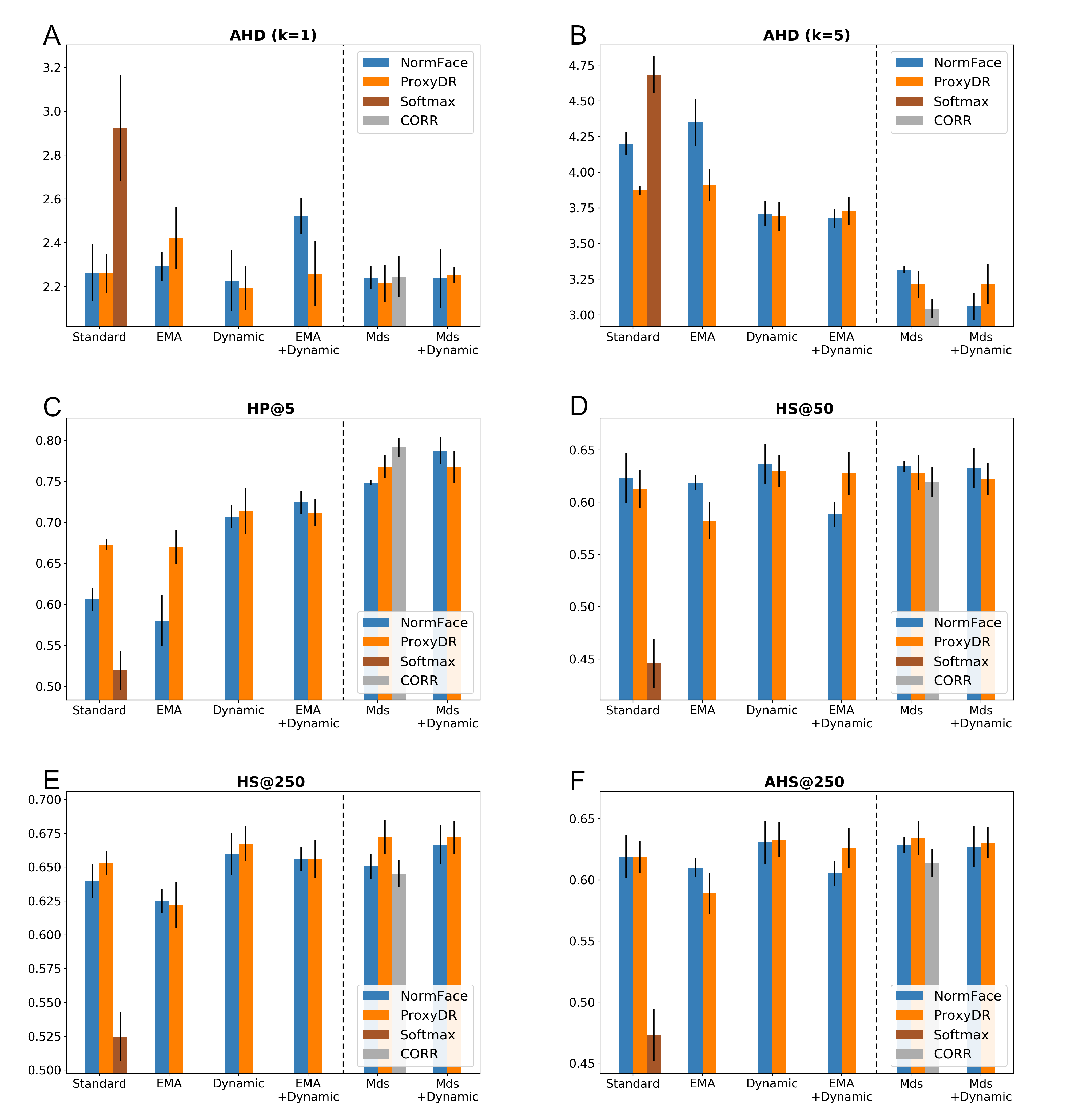}
\caption{{\bf Hierarchical performance measures on the CIFAR100 dataset.} 
The symbol \(\downarrow\) denotes that lower values indicate better performance. %
The symbol \(\uparrow\) denotes that higher values indicate better performance. %
(A) AHD (k=1): \(\downarrow\). (B) AHD (k=5): \(\downarrow\). (C) HP@5: \(\uparrow\). (D) HS@50: \(\uparrow\). (E) HS@250: \(\uparrow\). (F) AHS@250: \(\uparrow\). %
}
    \label{fig:cifar_H_measures}
\end{figure}

\begin{figure}[H]
    \centering
\vspace{.05in}
\includegraphics[width=1.0\linewidth,height=1.0\linewidth]{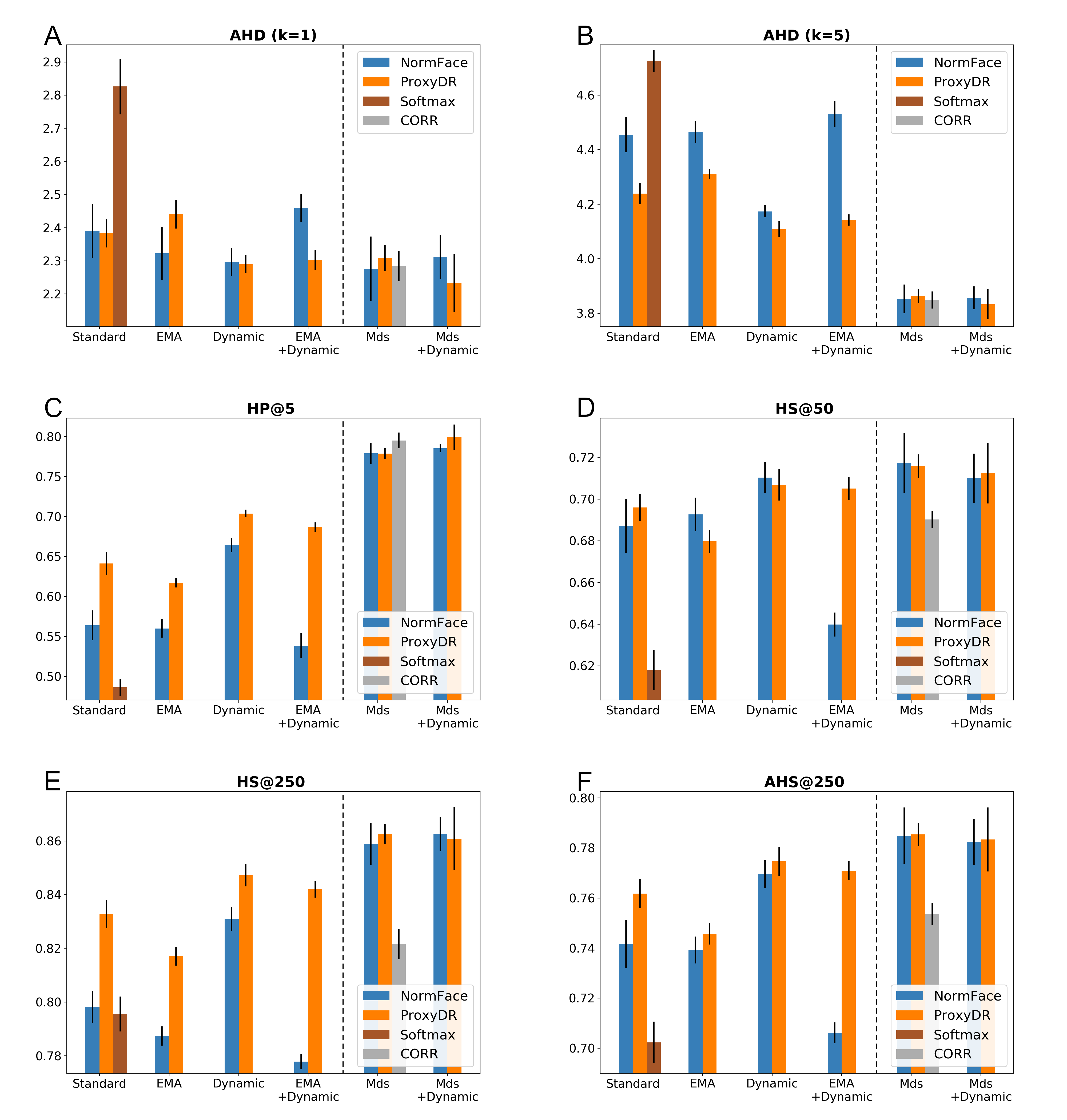}
\caption{{\bf Hierarchical performance measures on the NABirds dataset.}
The symbol \(\downarrow\) denotes that lower values indicate better performance. %
The symbol \(\uparrow\) denotes that higher values indicate better performance. %
(A) AHD (k=1): \(\downarrow\). (B) AHD (k=5): \(\downarrow\). (C) HP@5: \(\uparrow\). (D) HS@50: \(\uparrow\). (E) HS@250: \(\uparrow\). (F) AHS@250: \(\uparrow\). %
}
    \label{fig:nabird_H_measures}
\end{figure}

Figs. \ref{fig:2pri2f_H_measures}, \ref{fig:4pri_H_measures}, \ref{fig:5pri2_H_measures}, \ref{fig:cifar_H_measures}, and \ref{fig:nabird_H_measures} show the hierarchy-informed performance measures on the different datasets.

\begin{figure}[H]
    \centering
\vspace{.05in}
\includegraphics[width=1.0\linewidth,height=1.28\linewidth]{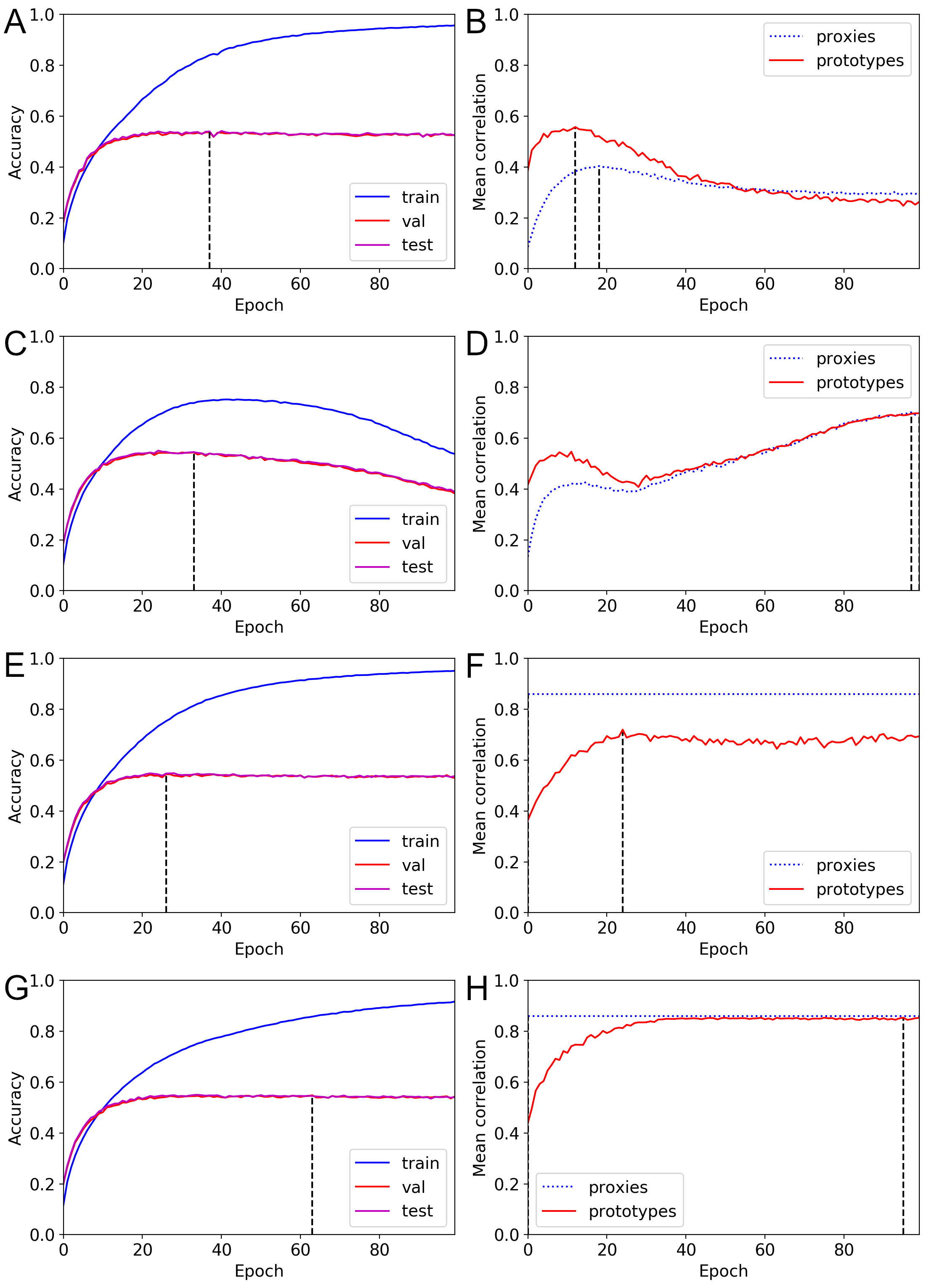} %
\caption{{\bf Changes in accuracy and mean correlations %
for the CIFAR100 dataset (NormFace).} %
(A) Accuracy curve with standard training. (B) Mean correlation curve with standard training. (C) Accuracy curve with the dynamic option. (D) Mean correlation curve with the dynamic option. %
(E) Accuracy curve with the MDS option. (F) Mean correlation curve with the MDS option. (G) Accuracy curve with the MDS \& dynamic options. (H) Mean correlation curve with the MDS \& dynamic options.
}
    \label{fig:Representing_cifar_normface}
\end{figure}

\begin{figure}[H]
    \centering
\vspace{.05in}
\includegraphics[width=1.0\linewidth,height=1.28\linewidth]{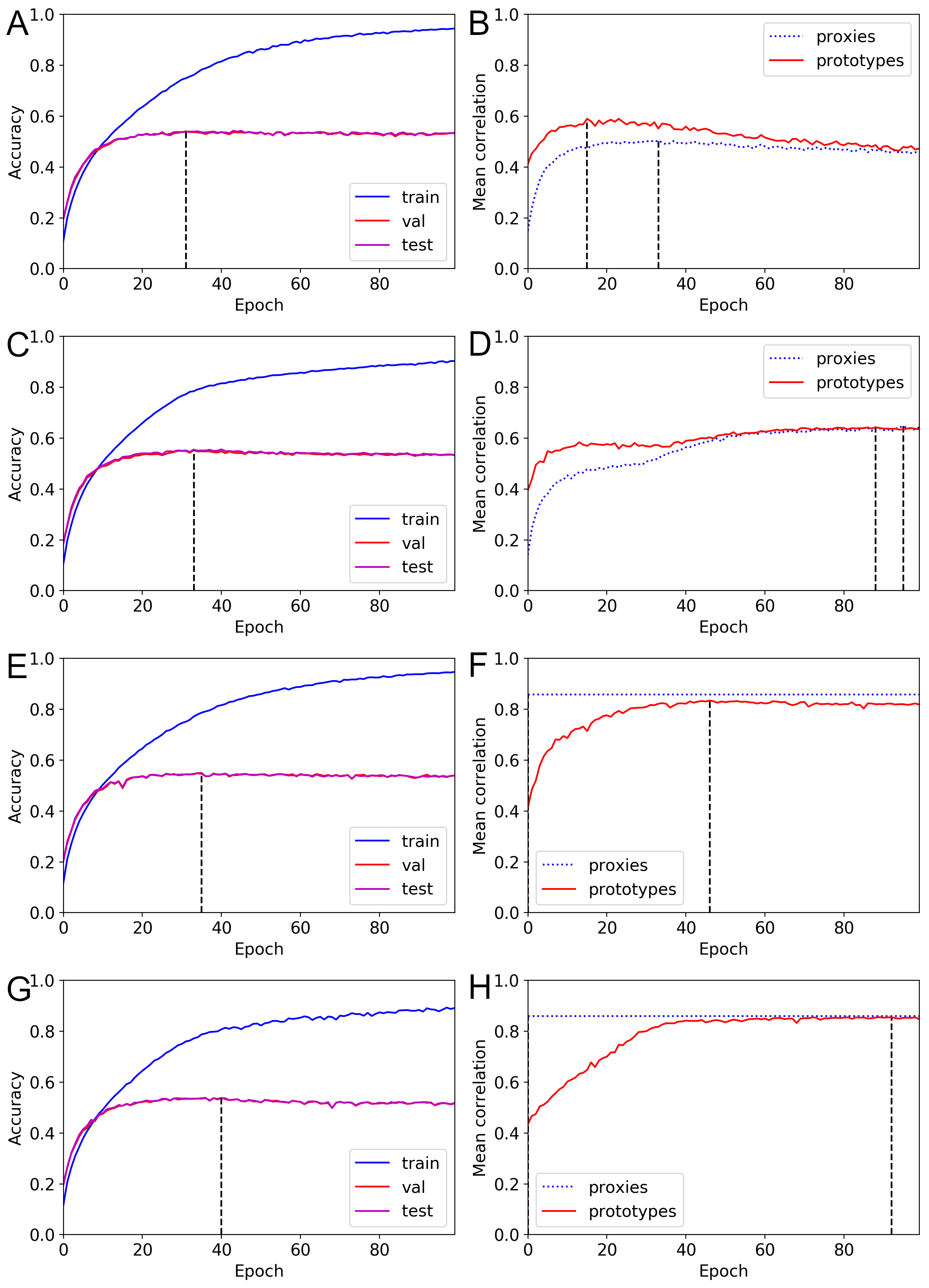} %
\caption{{\bf Changes in accuracy and mean correlations %
for the CIFAR100 dataset (ProxyDR).} %
(A) Accuracy curve with standard training. (B) Mean correlation curve with standard training. (C) Accuracy curve with the dynamic option. (D) Mean correlation curve with the dynamic option. %
(E) Accuracy curve with the MDS option. (F) Mean correlation curve with the MDS option. (G) Accuracy curve with the MDS \& dynamic options. (H) Mean correlation curve with the MDS \& dynamic options.
}
    \label{fig:Representing_cifar_proxyDR}
\end{figure}

\begin{figure}[H]
    \centering
\vspace{.05in}
\includegraphics[width=1.0\linewidth,height=0.32\linewidth]{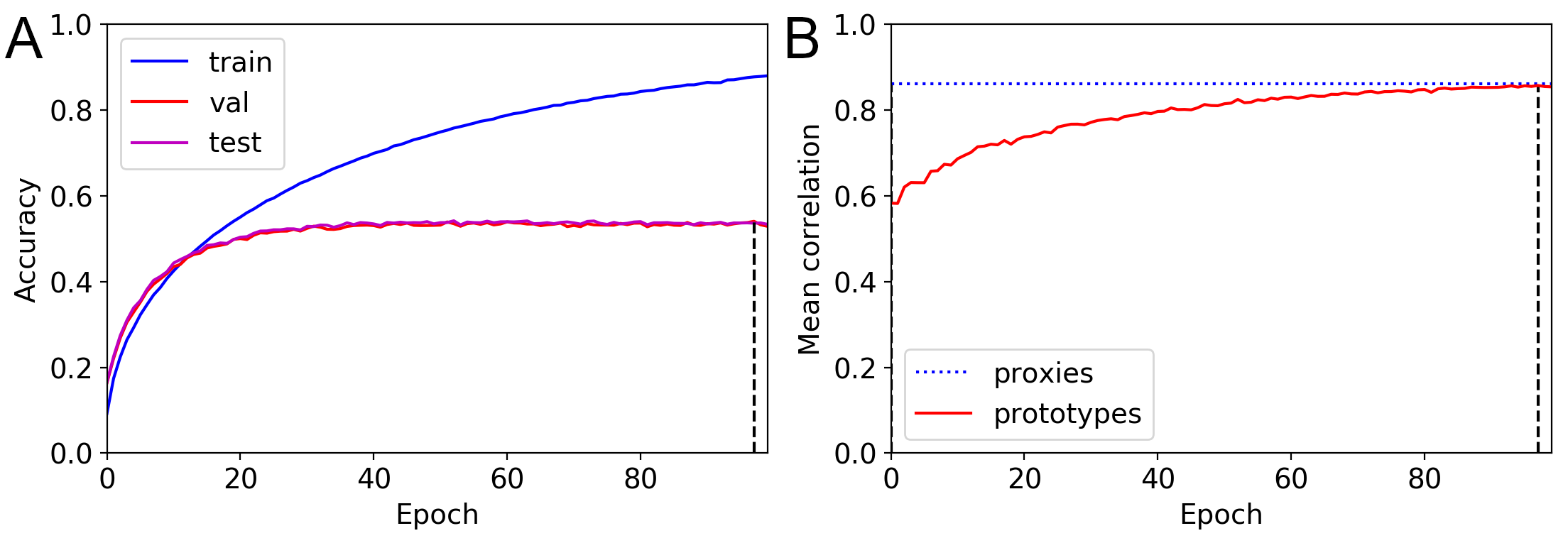}
\caption{{\bf Changes in accuracy and mean correlations %
for the CIFAR100 dataset (CORR).} %
(A) Accuracy curve with CORR loss training. (B) Mean correlation curve with CORR loss training. %
}
    \label{fig:Representing_cifar_CORR}
\end{figure}

\begin{figure}[H]
    \centering
\vspace{.05in}
\includegraphics[width=1.0\linewidth,height=1.28\linewidth]{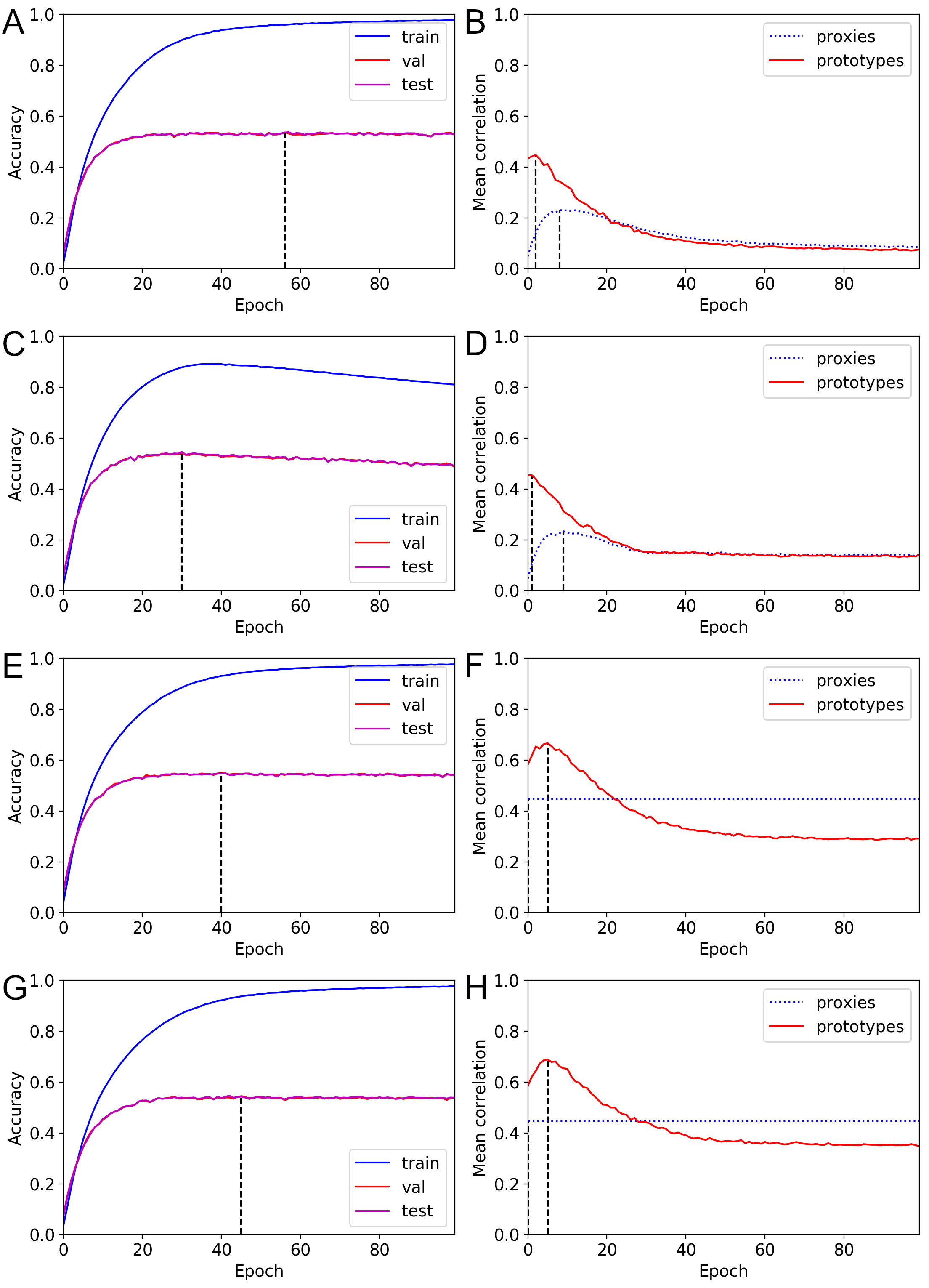} %
\caption{{\bf Changes in accuracy and mean correlations %
for the NABirds dataset (NormFace).} %
(A) Accuracy curve with standard training. (B) Mean correlation curve with standard training. (C) Accuracy curve with the dynamic option. (D) Mean correlation curve with the dynamic option. %
(E) Accuracy curve with the MDS option. (F) Mean correlation curve with the MDS option. (G) Accuracy curve with the MDS \& dynamic options. (H) Mean correlation curve with the MDS \& dynamic options.
}
    \label{fig:Representing_nabird_normface}
\end{figure}

\begin{figure}[H]
    \centering
\vspace{.05in}
\includegraphics[width=1.0\linewidth,height=1.28\linewidth]{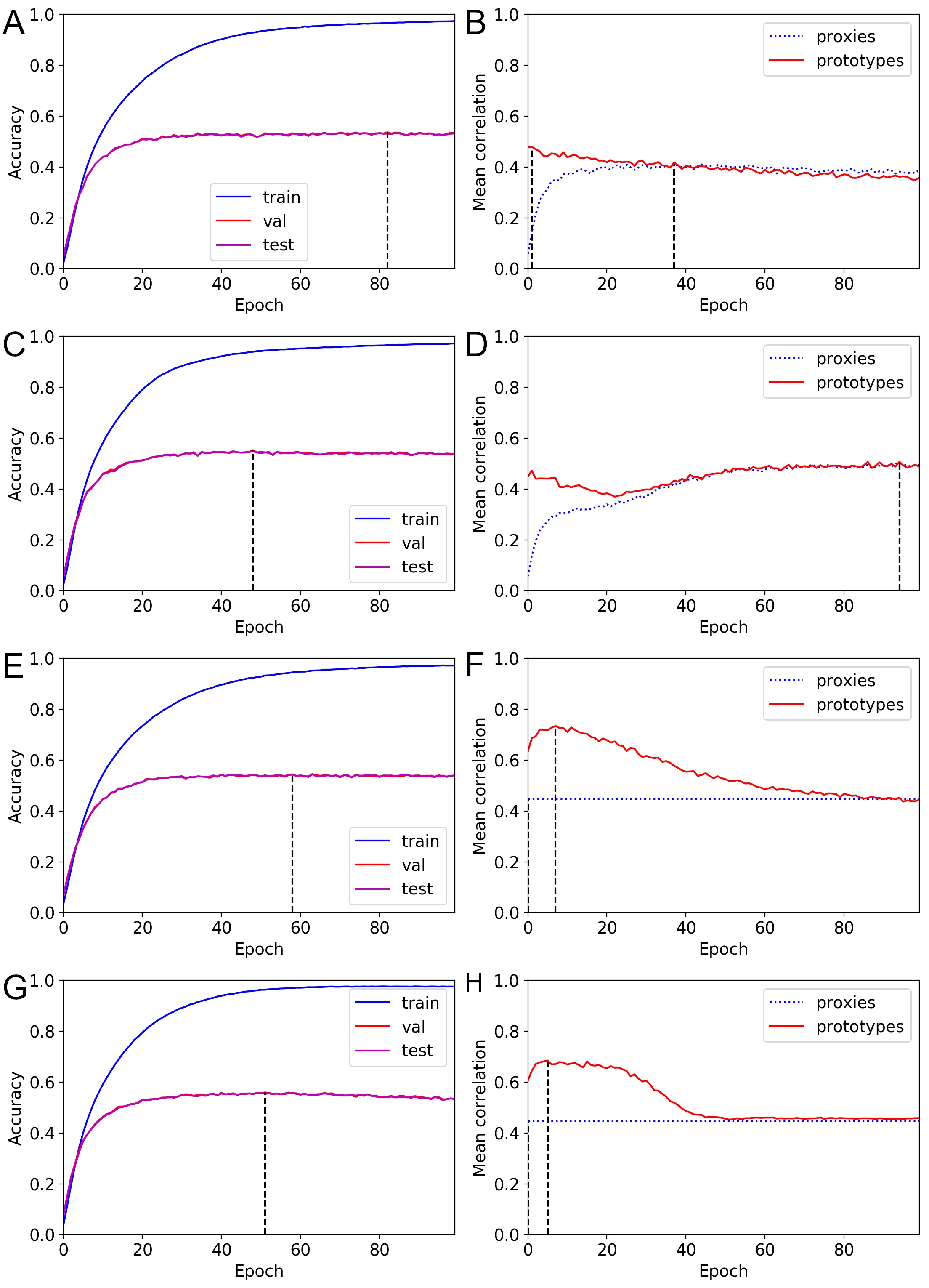} %
\caption{{\bf Changes in accuracy and mean correlations %
for the NABirds dataset (ProxyDR).} %
(A) Accuracy curve with standard training. (B) Mean correlation curve with standard training. (C) Accuracy curve with the dynamic option. (D) Mean correlation curve with the dynamic option. %
(E) Accuracy curve with the MDS option. (F) Mean correlation curve with the MDS option. (G) Accuracy curve with the MDS \& dynamic options. (H) Mean correlation curve with the MDS \& dynamic options.
}
    \label{fig:Representing_nabird_proxyDR}
\end{figure}

\begin{figure}[H]
    \centering
\vspace{.05in}
\includegraphics[width=1.0\linewidth,height=0.32\linewidth]{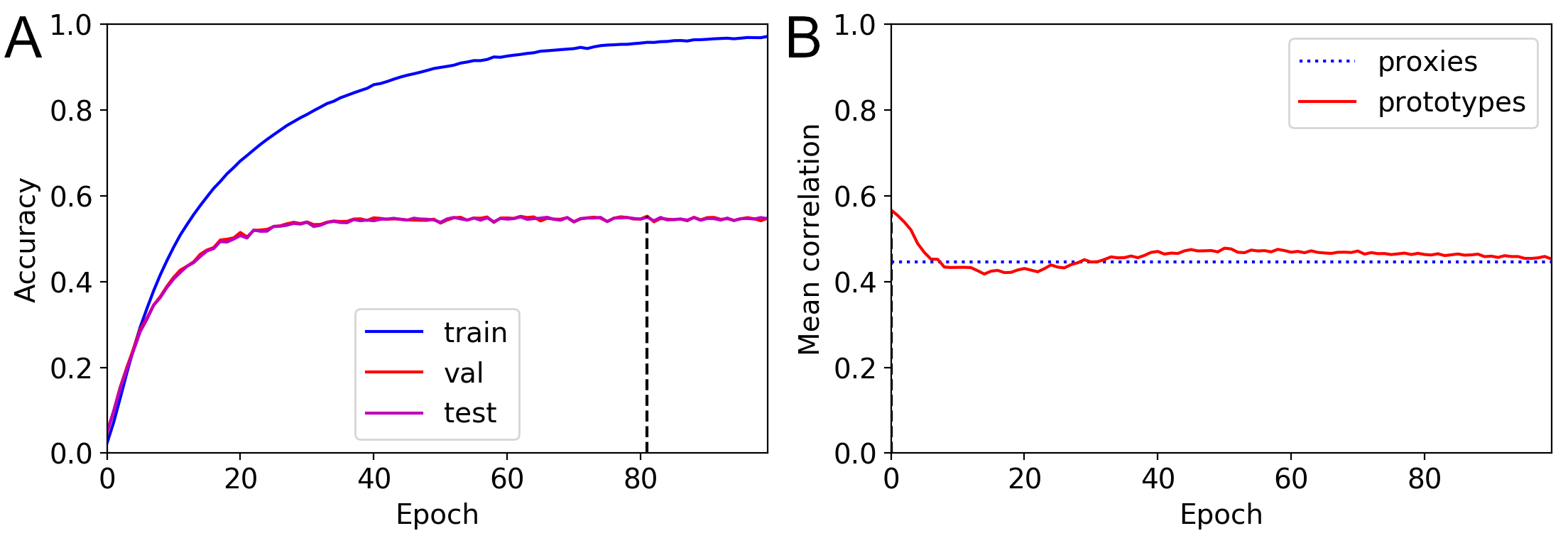} %
\caption{{\bf Changes in accuracy and mean correlations %
for the NABirds dataset (CORR).} %
(A) Accuracy curve with CORR loss training. (B) Mean correlation curve with CORR loss training. %
}
    \label{fig:Representing_nabird_CORR}
\end{figure}

{\bf Additional results on mean correlations\\}
Figs. \ref{fig:Representing_cifar_normface}, \ref{fig:Representing_cifar_proxyDR}, \ref{fig:Representing_cifar_CORR}, \ref{fig:Representing_nabird_normface}, \ref{fig:Representing_nabird_proxyDR} and \ref{fig:Representing_nabird_CORR} visualize the changes in the mean correlation values with different models and training options, showing the averaged values for five different seeds. Figs. \ref{fig:Representing_cifar_normface} and \ref{fig:Representing_nabird_normface} show that the accuracy of the NormFace models decreased after a certain number of epochs when the models were trained with the dynamic option. Interestingly, the mean correlation values increased even when the accuracy decreased. When using predefined hierarchical information and dynamic options (if applicable), the prototype-baesd mean correlation values approached the fixed proxy-based mean correlation values, except the NormFace model on the NABirds dataset.

\paragraph{Rank correlations between performance measures}

We investigate whether different performance measures are correlated. While we cannot derive causal relationships, this investigation is useful for determining if there are trade-offs between various performance measures. We used rank correlation coefficients between the mean correlation values and other measures. We ignored the softmax loss results, as this model is not based on metric learning. \\
In Table \ref{tab:corr_and_measures}, we investigate whether mean correlation values based on proxies (Figs. \ref{fig:2pri2f_corr}, \ref{fig:4pri_corr}, \ref{fig:5pri2_corr}, \ref{fig:cifar_corr}, and \ref{fig:nabird_corr}) are correlated with other performance measures (Figs. \ref{fig:2pri2f_acc}, \ref{fig:4pri_acc}, \ref{fig:5pri2_acc}, \ref{fig:cifar_acc}, \ref{fig:nabird_acc}, \ref{fig:2pri2f_H_measures}, \ref{fig:4pri_H_measures}, \ref{fig:5pri2_H_measures}, \ref{fig:cifar_H_measures}, and \ref{fig:nabird_H_measures}). %
The table shows that the AHD (k=5), HP@5, HS@250, and AHS@250 measures %
are correlated with the proxy-based mean correlation values. Moreover, there does not appear to be any trade-off between the top-1 accuracies and proxy-based mean correlations. However, a trade-off between the top-5 accuracy and proxy-based mean correlations is observed on some datasets.

\begin{table}[H]
\begin{adjustwidth}{-2.25in}{0in} %
\centering
\caption{
{\bf Rank (Spearman) correlation coefficients %
of different measures with mean correlations (based on proxies)}}
\begin{tabular}{\Mc{1.5cm}|\Mc{2.25cm}|\Mc{2.25cm}|\Mc{2.25cm}|\Mc{2.25cm}|\Mc{2.25cm}}
 Measure & MicroS & MicroL & MesoZ & CIFAR100 & NABirds \\
          \toprule
\scriptsize \multirow{2}{*}{Top-1}     & \scriptsize 0.2039 %
& \scriptsize  0.2260   &  \scriptsize  0.2171    &    \scriptsize 0.1675     & \scriptsize 0.3112      \\
          &  \scriptsize\((1.0336\times 10^{-1})\) %
          &  \scriptsize\((7.0284\times 10^{-2})\)  &  \scriptsize \((8.2377\times 10^{-2})\)     &  \scriptsize\((1.8239\times 10^{-1})\)     &   \scriptsize\((1.1627\times 10^{-2})\)   \\
          \midrule
\scriptsize \multirow{2}{*}{Top-5}     & \scriptsize -0.1350 %
& \scriptsize {\bf -0.3716}    &  \scriptsize -0.2120     & \scriptsize {\bf -0.3364 }       & \scriptsize {\bf -0.5669 }     \\
          &  \scriptsize\((2.8361\times 10^{-1})\) %
          &  \scriptsize\(( \bm{2.3043\times 10^{-3}})\)    &  \scriptsize \((9.0045\times 10^{-2})\)     &  \scriptsize\((\bm{6.1476\times 10^{-3}})\)     &   \scriptsize\((\bm{8.4964\times 10^{-7}})\)    \\
          \midrule
\scriptsize \multirow{2}{*}{AHD (k=1)} & \scriptsize {\bf -0.3423} %
& \scriptsize {\bf -0.3855 }   &  \scriptsize -0.2959     & \scriptsize -0.1821        &  \scriptsize{\bf -0.4267 }     \\
          & \scriptsize\((\bm{5.2583\times 10^{-3}})\) %
          &  \scriptsize\((\bm{1.5179\times 10^{-3}})\)   &  \scriptsize\((1.6684\times 10^{-2})\)     &  \scriptsize\((1.4658\times 10^{-1})\)   &  \scriptsize\((\bm{3.9270\times 10^{-4}})\)    \\
          \midrule
\scriptsize \multirow{2}{*}{AHD (k=5)} & \scriptsize {\bf -0.7407} %
& \scriptsize{\bf -0.8258 }    & \scriptsize {\bf -0.9026}     &   \scriptsize{\bf -0.8856 }      &   \scriptsize{\bf -0.7023 }    \\
          & \scriptsize\((\bm{1.7522\times 10^{-12}})\) %
          &  \scriptsize\((\bm{2.5540\times 10^{-17}})\)   & \scriptsize\((\bm{9.6705\times 10^{-25}})\)      &  \scriptsize\((\bm{1.1648\times 10^{-22}})\)   &  \scriptsize\((\bm{7.0541\times 10^{-11}})\)   \\
          \midrule
\scriptsize \multirow{2}{*}{HP@5}      & \scriptsize {\bf 0.7392} %
& \scriptsize {\bf 0.8350}    & \scriptsize {\bf 0.9001}    &   \scriptsize{\bf 0.8955 }      & \scriptsize {\bf 0.7070}      \\
          &  \scriptsize\((\bm{2.0489\times 10^{-12}})\)  %
          & \scriptsize\((\bm{5.4193\times 10^{-18}})\)     &  \scriptsize\((\bm{2.0612\times 10^{-24}})\)        & \scriptsize\((\bm{7.8000\times 10^{-24}})\) &  \scriptsize\(( \bm{4.6412\times 10^{-11}})\)    \\
          \midrule
\scriptsize \multirow{2}{*}{HS@50}     & \scriptsize {\bf 0.7671} %
& \scriptsize{\bf  0.7962}    & \scriptsize {\bf 0.5634}    &  \scriptsize0.1878        & \scriptsize {\bf 0.5276 }     \\
          & \scriptsize\((\bm{9.2047\times 10^{-14}})\)
          & \scriptsize\((\bm{2.2284\times 10^{-15}})\)    &  \scriptsize\((\bm{1.0275\times 10^{-6}})\)      &  \scriptsize\((1.3410\times 10^{-1})\)  &  \scriptsize\((\bm{6.3074\times 10^{-6}})\)   \\
          \midrule
\scriptsize \multirow{2}{*}{HS@250}     & \scriptsize {\bf 0.7911} & \scriptsize{\bf 0.8167}    & \scriptsize {\bf 0.8860}    &  \scriptsize {\bf 0.5514}        & \scriptsize {\bf 0.7210 }     \\
          & \scriptsize\((\bm{4.4738\times 10^{-15}})\) 
          & \scriptsize\((\bm{1.1029\times 10^{-16}})\)    &  \scriptsize\((\bm{1.0514\times 10^{-22}})\)      &  \scriptsize \((\bm{1.9333\times 10^{-6}})\)  &  \scriptsize\((\bm{1.2570\times 10^{-11}})\)   \\
          \midrule
\scriptsize \multirow{2}{*}{AHS@250}   & \scriptsize {\bf 0.7950} %
& \scriptsize {\bf 0.8186}    & \scriptsize {\bf 0.8034 }   & \scriptsize{\bf 0.3335 }        &  \scriptsize {\bf 0.6862 }    \\
          &  \scriptsize\((\bm{2.6376\times 10^{-15}})\)%
          &  \scriptsize\((\bm{8.1611\times 10^{-17}})\)  &  \scriptsize\((\bm{8.1089\times 10^{-16}})\)   &  \scriptsize\((\bm{6.6355\times 10^{-3}})\)    &  \scriptsize\((\bm{2.8150\times 10^{-10}})\)   \\
          \bottomrule
\end{tabular}
\begin{flushleft} \(p\) values are written in parentheses. Significant results (\(p\) value \(<0.01\)) are written in bold text. 
\end{flushleft}
\label{tab:corr_and_measures}
\end{adjustwidth}
\end{table}

In Table \ref{tab:top1_and_measures}, we investigate whether the top-1 accuracy (Figs. \ref{fig:2pri2f_acc}, \ref{fig:4pri_acc}, \ref{fig:5pri2_acc}, \ref{fig:cifar_acc}, and \ref{fig:nabird_acc}) is correlated with other performance measures (Figs. \ref{fig:2pri2f_acc}, \ref{fig:4pri_acc}, \ref{fig:5pri2_acc}, \ref{fig:cifar_acc}, \ref{fig:nabird_acc}, \ref{fig:2pri2f_H_measures}, \ref{fig:4pri_H_measures}, \ref{fig:5pri2_H_measures}, \ref{fig:cifar_H_measures}, and \ref{fig:nabird_H_measures}). The table shows that the AHD (k=1), HS@50, and AHS@250 measures %
are correlated with the top-1 accuracy. Note that these are the measures that did not show noticeable changes on some datasets when predefined hierarchical information was used during training. %

\begin{table}[H]
\begin{adjustwidth}{-2.25in}{0in} %
\centering
\caption{
{\bf Rank (Spearman) correlation coefficients %
of different measures with the top-1 accuracy}}
\begin{tabular}{\Mc{1.5cm}|\Mc{2.25cm}|\Mc{2.25cm}|\Mc{2.25cm}|\Mc{2.25cm}|\Mc{2.25cm}}
 Measure & MicroS & MicroL & MesoZ & CIFAR100 & NABirds \\
          \toprule
\scriptsize \multirow{2}{*}{Top-5}     & \scriptsize 0.2199 & \scriptsize 0.0365 &  \scriptsize 0.1033 &    \scriptsize {\bf 0.4398} & \scriptsize -0.1043 \\
          &  \scriptsize\(( 7.8357\times 10^{-2 })\)   &  \scriptsize\(( 7.7302\times 10^{-1 })\)  &  \scriptsize \(( 4.1302\times 10^{-1 })\)     &  \scriptsize\((\bm{2.4659\times 10^{-4 }})\)     &   \scriptsize\(( 4.0815\times 10^{-1 })\)   \\
          \midrule
\scriptsize \multirow{2}{*}{AHD (k=1)} & \scriptsize {\bf -0.8005} & \scriptsize {\bf -0.8302} &  \scriptsize {\bf -0.9309} &    \scriptsize {\bf -0.9507} & \scriptsize {\bf -0.9509} \\
          &  \scriptsize\(( \bm{1.2168\times 10^{-15 }})\)   &  \scriptsize\(( \bm{1.2369\times 10^{-17 }})\)  &  \scriptsize \(( \bm{2.9921\times 10^{-29 }})\)     &  \scriptsize\(( \bm{9.4257\times 10^{-34 }})\)     &   \scriptsize\(( \bm{8.6632\times 10^{-34 }})\)   \\
          \midrule
\scriptsize \multirow{2}{*}{AHD (k=5)} & \scriptsize -0.1906 & \scriptsize {\bf -0.3259} &  \scriptsize -0.2908 &    \scriptsize -0.2877 & \scriptsize {\bf -0.6136} \\
          &  \scriptsize\(( 1.2829\times 10^{-1 })\)   &  \scriptsize\(( \bm{8.0758\times 10^{-3 }})\)  &  \scriptsize \(( 1.8752\times 10^{-2 })\)     &  \scriptsize\(( 2.0115\times 10^{-2 })\)     &   \scriptsize\(( \bm{5.4992\times 10^{-8 }})\)   \\
          \midrule
\scriptsize \multirow{2}{*}{HP@5}      & \scriptsize 0.2263 & \scriptsize 0.2956 &  \scriptsize 0.2886 &    \scriptsize 0.2439 & \scriptsize {\bf 0.6029} \\
          &  \scriptsize\(( 6.9923\times 10^{-2 })\)   &  \scriptsize\(( 1.6811\times 10^{-2 })\)  &  \scriptsize \(( 1.9712\times 10^{-2 })\)     &  \scriptsize\(( 5.0222\times 10^{-2 })\)     &   \scriptsize\(( \bm{1.0692\times 10^{-7 }})\)   \\
          \midrule
\scriptsize \multirow{2}{*}{HS@50}     & \scriptsize {\bf 0.4325} & \scriptsize {\bf 0.5085} &  \scriptsize {\bf 0.6979} &    \scriptsize {\bf 0.8967} & \scriptsize {\bf 0.6832} \\
          &  \scriptsize\(( \bm{3.2041\times 10^{-4 }})\)   &  \scriptsize\(( \bm{1.5301\times 10^{-5 }})\)  &  \scriptsize \(( \bm{1.0420\times 10^{-10 }})\)     &  \scriptsize\(( \bm{5.5302\times 10^{-24 }})\)     &   \scriptsize\(( \bm{3.6174\times 10^{-10 }})\)   \\
          \midrule
\scriptsize \multirow{2}{*}{HS@250}     & \scriptsize 0.1369 & \scriptsize 0.3057 &  \scriptsize 0.2945 &    \scriptsize {\bf 0.5535} & \scriptsize {\bf 0.4915} \\
          &  \scriptsize\(( 2.7674\times 10^{-1 })\)   &  \scriptsize\(( 1.3279\times 10^{-2 })\)  &  \scriptsize \(( 1.7269\times 10^{-2 })\)     &  \scriptsize\(( \bm{1.7312\times 10^{-6 }})\)     &   \scriptsize\(( \bm{3.2203\times 10^{-5 }})\)   \\
          \midrule
\scriptsize \multirow{2}{*}{AHS@250}  & \scriptsize {\bf 0.3394} & \scriptsize {\bf 0.4233} &  \scriptsize {\bf 0.5067} &    \scriptsize {\bf 0.8517} & \scriptsize {\bf 0.5733} \\
          &  \scriptsize\(( \bm{5.6730\times 10^{-3 }})\)   &  \scriptsize\(( \bm{4.4195\times 10^{-4 }})\)  &  \scriptsize \(( \bm{1.6583\times 10^{-5 }})\)     &  \scriptsize\(( \bm{2.4413\times 10^{-19 }})\)     &   \scriptsize\(( \bm{6.0057\times 10^{-7 }})\)   \\
          \midrule
\scriptsize \multirow{2}{*}{\shortstack{Mean\\ correlation\\ (proxy)}}     & \scriptsize 0.2039 & \scriptsize 0.2260 &  \scriptsize 0.2171 &    \scriptsize 0.1675 & \scriptsize 0.3112 \\
          &  \scriptsize\(( 1.0336\times 10^{-1 })\)   &  \scriptsize\(( 7.0284\times 10^{-2 })\)  &  \scriptsize \(( 8.2377\times 10^{-2 })\)     &  \scriptsize\(( 1.8239\times 10^{-1 })\)     &   \scriptsize\(( 1.1627\times 10^{-2 })\)   \\
          \midrule
\scriptsize \multirow{2}{*}{\shortstack{Mean\\ correlation\\ (prototype)}}     & \scriptsize 0.2150 & \scriptsize 0.2055 &  \scriptsize 0.1738 &    \scriptsize 0.1978 & \scriptsize 0.2664 \\
          &  \scriptsize\(( 8.5503\times 10^{-2 })\)   &  \scriptsize\(( 1.0051\times 10^{-1 })\)  &  \scriptsize \(( 1.6616\times 10^{-1 })\)     &  \scriptsize\(( 1.1423\times 10^{-1 })\)     &   \scriptsize\(( 3.1936\times 10^{-2 })\)   \\
          \bottomrule
\end{tabular}
\begin{flushleft} \(p\) values are written in parentheses. Significant results (\(p\) value \(<0.01\)) are written in bold text. 
\end{flushleft}
\label{tab:top1_and_measures}
\end{adjustwidth}
\end{table}

In Table \ref{tab:top5_and_measures}, we investigate whether the top-5 accuracy (Figs. \ref{fig:2pri2f_acc}, \ref{fig:4pri_acc}, \ref{fig:5pri2_acc}, \ref{fig:cifar_acc}, and \ref{fig:nabird_acc}) is correlated with other performance measures (Figs. \ref{fig:2pri2f_acc}, \ref{fig:4pri_acc}, \ref{fig:5pri2_acc}, \ref{fig:cifar_acc}, \ref{fig:nabird_acc}, \ref{fig:2pri2f_H_measures}, \ref{fig:4pri_H_measures}, \ref{fig:5pri2_H_measures}, \ref{fig:cifar_H_measures}, and \ref{fig:nabird_H_measures}). The table shows that no measures are correlated with the top-5 accuracy on all datasets. %

\begin{table}[H]
\begin{adjustwidth}{-2.25in}{0in} %
\centering
\caption{
{\bf Rank (Spearman) correlation coefficients %
of different measures with the top-5 accuracy}}
\begin{tabular}{\Mc{1.5cm}|\Mc{2.25cm}|\Mc{2.25cm}|\Mc{2.25cm}|\Mc{2.25cm}|\Mc{2.25cm}}
 Measure & MicroS & MicroL & MesoZ & CIFAR100 & NABirds \\
          \toprule
\scriptsize \multirow{2}{*}{Top-1}     & \scriptsize 0.2199 & \scriptsize 0.0365 &  \scriptsize 0.1033 &    \scriptsize {\bf 0.4398} & \scriptsize -0.1043 \\
          &  \scriptsize\(( 7.8357\times 10^{-2 })\)   &  \scriptsize\((7.7302 \times 10^{-1 })\)  &  \scriptsize \((4.1302 \times 10^{-1 })\)     &  \scriptsize\(( \bm{2.4659\times 10^{-4}})\)     &   \scriptsize\(( 4.0815\times 10^{-1 })\)   \\
          \midrule
\scriptsize \multirow{2}{*}{AHD (k=1)} & \scriptsize -0.0716 & \scriptsize 0.0033 &  \scriptsize -0.0406 &    \scriptsize {\bf -0.4245} & \scriptsize 0.1196 \\
          &  \scriptsize\(( 5.7106\times 10^{-1 })\)   &  \scriptsize\(( 9.7911\times 10^{-1 })\)  &  \scriptsize \(( 7.4809\times 10^{-1 })\)     &  \scriptsize\(( \bm{4.2428\times 10^{-4 }})\)     &   \scriptsize\(( 3.4268\times 10^{-1 })\)   \\
          \midrule
\scriptsize \multirow{2}{*}{AHD (k=5)} & \scriptsize 0.2417 & \scriptsize {\bf 0.4842} &  \scriptsize 0.2732 &    \scriptsize {\bf 0.4561} & \scriptsize {\bf 0.4949} \\
          &  \scriptsize\(( 5.2436\times 10^{-2 })\)   &  \scriptsize\(( \bm{4.3828\times 10^{-5 }})\)  &  \scriptsize \(( 2.7654\times 10^{-2 })\)     &  \scriptsize\(( \bm{1.3422\times 10^{-4 }})\)     &   \scriptsize\(( \bm{2.7791\times 10^{-5 }})\)   \\
          \midrule
\scriptsize \multirow{2}{*}{HP@5}   & \scriptsize -0.2765 & \scriptsize {\bf -0.5318} &  \scriptsize -0.3113 &    \scriptsize {\bf -0.4902} & \scriptsize {\bf -0.5733} \\
          &  \scriptsize\(( 2.5757\times 10^{-2 })\)   &  \scriptsize\(( \bm{5.1600\times 10^{-6 }})\)  &  \scriptsize \(( 1.1590\times 10^{-2 })\)     &  \scriptsize\(( \bm{3.3964\times 10^{-5 }})\)     &   \scriptsize\(( \bm{5.9953\times 10^{-7 }})\)   \\
          \midrule
\scriptsize \multirow{2}{*}{HS@50}  & \scriptsize 0.0115 & \scriptsize -0.3164 &  \scriptsize -0.2331 &    \scriptsize 0.2515 & \scriptsize -0.0071 \\
          &  \scriptsize\(( 9.2784\times 10^{-1 })\)   &  \scriptsize\(( 1.0231\times 10^{-2 })\)  &  \scriptsize \(( 6.1677\times 10^{-2 })\)     &  \scriptsize\(( 4.3295\times 10^{-2 })\)     &   \scriptsize\(( 9.5522\times 10^{-1})\)   \\
          \midrule
\scriptsize \multirow{2}{*}{HS@250}  & \scriptsize -0.0677 & \scriptsize {\bf -0.4343} &  \scriptsize -0.1965 &    \scriptsize 0.0955 & \scriptsize {\bf -0.3192} \\
          &  \scriptsize\(( 5.9231\times 10^{-1 })\)   &  \scriptsize\(( \bm{3.0034\times 10^{-4 }})\)  &  \scriptsize \(( 1.1677\times 10^{-1 })\)     &  \scriptsize\(( 4.4900\times 10^{-1 })\)     &   \scriptsize\(( \bm{9.5513\times 10^{-3 }})\)   \\
          \midrule
\scriptsize \multirow{2}{*}{AHS@250}  & \scriptsize 0.0249 & \scriptsize {\bf -0.3603} &  \scriptsize -0.1446 &    \scriptsize 0.3157 & \scriptsize -0.2203 \\
          &  \scriptsize\(( 8.4393\times 10^{-1 })\)   &  \scriptsize\(( \bm{3.1949\times 10^{-3 }})\)  &  \scriptsize \(( 2.5059\times 10^{-1 })\)     &  \scriptsize\(( 1.0404\times 10^{-2 })\)     &   \scriptsize\(( 7.7867\times 10^{-2 })\)   \\
          \midrule
\scriptsize \multirow{2}{*}{\shortstack{Mean\\ correlation\\ (proxy)}}  & \scriptsize -0.1350 & \scriptsize {\bf -0.3716} &  \scriptsize -0.2120 &    \scriptsize {\bf -0.3364} & \scriptsize {\bf -0.5669} \\
          &  \scriptsize\(( 2.8361\times 10^{-1 })\)   &  \scriptsize\(( \bm{2.3043\times 10^{-3 }})\)  &  \scriptsize \((9.0045 \times 10^{-2 })\)     &  \scriptsize\(( \bm{6.1476\times 10^{-3 }})\)     &   \scriptsize\(( \bm{8.4964\times 10^{-7 }})\)   \\
          \midrule
\scriptsize \multirow{2}{*}{\shortstack{Mean\\ correlation\\ (prototype)}}   & \scriptsize -0.1040 & \scriptsize {\bf -0.4784} &  \scriptsize -0.2910 &    \scriptsize {\bf -0.3675} & \scriptsize {\bf -0.5457} \\
          &  \scriptsize\(( 4.0985\times 10^{-1 })\)   &  \scriptsize\(( \bm{5.5679\times 10^{-5 }})\)  &  \scriptsize \(( 1.8677\times 10^{-2 })\)     &  \scriptsize\(( \bm{2.5967\times 10^{-3 }})\)     &   \scriptsize\(( \bm{2.5927\times 10^{-6 }})\)   \\
          \bottomrule
\end{tabular}
\begin{flushleft} \(p\) values are written in parentheses. Significant results (\(p\) value \(<0.01\)) are written in bold text. 
\end{flushleft}
\label{tab:top5_and_measures}
\end{adjustwidth}
\end{table}

\end{document}